\documentclass{article}

\usepackage{PRIMEarxiv}

\usepackage[utf8]{inputenc} 
\usepackage[T1]{fontenc}    
\usepackage{hyperref}       
\usepackage{url}            
\usepackage{booktabs}       
\usepackage{amsfonts, amsmath, amssymb, bm}       
\usepackage{nicefrac}       
\usepackage{microtype}      
\usepackage{lipsum}
\usepackage{graphicx}
\usepackage{xcolor, soul}
\graphicspath{ {./images/} }
\usepackage{overpic}
\usepackage{color}
\usepackage[caption=false]{subfig}
\usepackage{adjustbox}
\usepackage{rotating}
\usepackage{float}

\usepackage{algorithm}
\usepackage{algpseudocode}
\usepackage{mathtools}

\usepackage{caption}
\usepackage{subcaption}
\usepackage{authblk}

\newtheorem{definition}{Definition}[section]

\title{HypeMARL: Multi-Agent Reinforcement Learning For High-Dimensional, Parametric, and Distributed Systems}

\author[1,*]{Nicolò Botteghi}
\author[2,*]{Matteo Tomasetto}
\author[3]{Urban Fasel}
\author[2]{Francesco Braghin}
\author[1]{Andrea Manzoni}
\affil[1]{MOX -- Department of Mathematics, Politecnico di Milano, Milano, Italy}
\affil[2]{Department of Mechanical Engineering, Politecnico di Milano, Milano, Italy}
\affil[3]{Department of Aeronautics, Imperial College London, London, United Kingdom}
\affil[*]{Equal contribution}

\begin{document}
\maketitle

\begin{abstract}
Deep reinforcement learning has recently emerged as a promising feedback control strategy for complex dynamical systems governed by partial differential equations (PDEs). When dealing with distributed, high-dimensional problems in state and control variables, multi-agent reinforcement learning (MARL) has been proposed as a scalable approach for breaking the curse of dimensionality. In particular, through decentralized training and execution, multiple agents cooperate to steer the system towards a target configuration, relying solely on local state and reward information. However, the principle of locality may become a limiting factor whenever a collective, nonlocal behavior of the agents is crucial to maximize the reward function, as typically happens in PDE-constrained optimal control problems. In this work, we propose \emph{HypeMARL}: a decentralized MARL algorithm tailored to the control of high-dimensional, parametric, and distributed systems. HypeMARL employs hypernetworks to effectively parametrize the agents' policies and value functions with respect to the system parameters and the agents' relative positions, encoded by sinusoidal positional encoding. Through the application on challenging control problems, such as density and flow control, we show that HypeMARL {\em (i)} can effectively control systems through a collective behavior of the agents, outperforming state-of-the-art decentralized MARL, {\em (ii)} can efficiently deal with parametric dependencies, {\em (iii)} requires minimal hyperparameter tuning and {\em (iv)} can reduce the amount of expensive environment interactions by a factor of $\sim 10$ thanks to its model-based extension, \emph{MB-HypeMARL}, which relies on computationally efficient deep learning-based surrogate models approximating the dynamics locally, with minimal deterioration of the policy performance.

\end{abstract}


\section{Introduction}

Many modern and large-scale engineering systems are composed of many smaller subsystems, such as in robotics, sensor networks, transportation systems, smart grids, and biological networks. We refer to these systems as \emph{distributed systems}, consisting of the interconnection of two or more subsystems that work together to perform complex tasks \cite{van2011control}. The control of distributed systems relies on two or more controllers, each receiving an observation stream from a local sensor and providing a control input to the local subsystem. Unlike \emph{centralized} systems, where a single controller controls the behavior of the entire system, distributed systems rely on \emph{decentralized} decision-making. Each controller typically operates autonomously, often with only partial knowledge of the global state, and interacts with its neighbors. The central challenge lies in designing local yet coordinated control strategies that leverage only limited and local information, ensuring optimal behavior across the entire system. The a priori design of an effective control strategy may be very difficult or even impossible due to the complexity inherent in distributed systems. Indeed, the controllers might need to reach agreement on a shared quantity, maintain a collective formation, synchronize behaviors, or optimize a common cost function. On top of that, the controllers may be heterogeneous, have different dynamics, sensing capabilities, or objectives. Additionally, the controller's behavior often needs to be updated online in order to reflect, for instance, changes in the environment over time, while gradually improving the performance of the whole distributed system \cite{bucsoniu2010multi}.

\emph{Reinforcement learning} (RL) is a branch of machine learning concerned with how \emph{agents}, namely the controllers of the distributed system, should take \emph{actions}, namely select control inputs, in an environment in order to maximize their cumulative reward \cite{sutton2018reinforcement}. 
Since the agent does not initially know which actions will yield the highest reward, it must explore different strategies to improve its behavior, balancing the trade-off between exploiting known high-reward actions and exploring new, potentially even higher-reward actions. This framework is suitable for a wide range of sequential decision-making problems, such as game playing, robotic control, autonomous driving, and resource management, to mention a few. The agent's behavior is defined by its \emph{policy} $\pi$, namely the set of rules that the agent exploits to select a specific action in a given state. The policy may be either stochastic $\pi:\mathbb{R}^{N_y} \times \mathbb{R}^{N_u}\to [0,1]$ or deterministic $\pi:\mathbb{R}^{N_y}\to \mathbb{R}^{N_u}$, where $N_y$ and $N_u$ denote the dimensions of the state and action space, respectively. The agent's goal is to find the optimal policy, i.e., the policy that maximizes the total expected cumulative reward from every state.
RL algorithms can be classified according to the strategy used to find the optimal policy: {\em (i)} \emph{value-based methods}, {\em (ii)} \emph{policy-based methods}, and {\em (iii)} \emph{actor-critic methods}. Value-based methods look for an estimate of the state-action value function $Q:\mathbb{R}^{N_y}\times \mathbb{R}^{N_u} \to \mathbb{R}$ associated with a policy $\pi$. The state-action value function gives the expected return obtained when following the policy $\pi$, starting from a given state. Policy-based methods, instead, consider a parametrization of the policy $\pi_{\boldsymbol{\theta}}$ with respect to a set of parameters $\boldsymbol{\theta}$ -- which may be, for instance, neural network (NN) weights and biases -- and aim at optimizing those parameters through the policy gradient \cite{sutton1999policy, sutton2018reinforcement} Eventually, actor-critic strategies combine the estimation of the value function with the direct optimization of the policy parameters. Another important distinction regarding RL algorithms is that between {\em (i)} \emph{model-free} and {\em (ii)} \emph{model-based} algorithms. Model-free algorithms utilize data collected from the interaction with the environment to directly learn the optimal policy and value function. Conversely, model-based RL exploits the data stream to build a \emph{surrogate} model of the environment dynamics that serves as a computationally efficient proxy of the real environment to optimize the control strategy. 


\emph{Multi-agent reinforcement learning} (MARL) is the generalization of RL to multi-agent systems where optimal policies have to be determined for a set of agents. Typical application domains of MARL are distributed control \cite{bucsoniu2010multi}, multi-robot systems \cite{orr2023multi, huttenrauch2019deep}, resource management \cite{peng2020multi}, games and competitive environments \cite{littman1994markov, tampuu2017multiagent}, autonomous driving and traffic control \cite{bazzan2009opportunities, wiering2000multi, van2016deep}, and only recently flow control \cite{vignon2023effective, suarez2025active}. An important distinction among MARL algorithms lies in the assumptions made when training and deploying the agents. In particular, we can identify three classes of methods, namely {\em (i)} \emph{centralized training and execution}, {\em (ii)} \emph{decentralized training and execution}, and {\em (iii)} \emph{centralized training and decentralized execution}. In the centralized training and execution approach, all agent share all the state observations with each other during the learning and deployment phase. In this case, the value function is defined using the state and joint actions space, i.e. $Q:\mathbb{R}^{N_y}\times \mathbb{R}^{N_u} \to \mathbb{R}$, while the stochastic policy of the $i^{th}$ agent is $\pi_i:\mathbb{R}^{N_y}\times \mathbb{R}^{N_{u_i}} \to [0, 1]$, where $\mathbb{R}^{N_{u_i}}$ denotes the local action space of the $i^{th}$ agent with dimension $N_{u_i}$. Conversely, decentralized training and execution approaches assume no shared knowledge among the agents, with every agent relying solely on local information. This results in local value functions and policies based only on local state information, i.e. $Q_i:\mathbb{R}^{N_{y_i}}\times \mathbb{R}^{N_{u_i}} \to \mathbb{R}$ and $\pi_i:\mathbb{R}^{N_{y_i}}\times \mathbb{R}^{N_{u_i}} \to [0, 1]$, where $N_{y_i}$ denotes the dimension of the local state space. Eventually, centralized training and decentralized execution methods combine the two aforementioned strategies, with observations shared among the agents during training, while only local state information is taken into account during execution. This third case is a hybrid case of the first two, in which a global value function is estimated, namely $Q:\mathbb{R}^{N_y}\times \mathbb{R}^{N_u} \to \mathbb{R}$, and a local policy is optimized, namely $\pi_i:\mathbb{R}^{N_{y_i}}\times \mathbb{R}^{N_{u_i}} \to [0, 1]$. Recent research in MARL has focused on centralized training and decentralized execution approaches. Examples in this direction are multi-agent DDPG (MADDPG) \cite{lowe2017multi}, multi-agent PPO \cite{yu2022surprising}, multi-agent trust region policy optimization \cite{kuba2021trust}, and multi-agent attention actor-critic \cite{iqbal2019actor}. However, when dealing with a large number of agents, learning a joint value function is extremely difficult \cite{iqbal2019actor}. To mitigate this issue, it is possible to consider value factorization methods, such as value decomposition network \cite{sunehag2017value}, QMIX \cite{rashid2020monotonic}, FacMADDPG \cite{de2020deep}, QTRAN \cite{son2019qtran}, and mean-field MARL \cite{yang2018mean}, which learn a factorization of the joint value function. However, these methods suffer from relative overgeneralization, where policies converge to suboptimal joint actions \cite{wei2016lenient, de2020independent}. While independent Q-learning has been first proposed in \cite{tan1993multi}, decentralized approaches -- such as, e.g., independent PPO \cite{de2020independent} and independent actor-critic COMA \cite{foerster2018counterfactual} -- have recently regained popularity as they outperform centralized approaches in very complex and high-dimensional problems \cite{de2020independent}. It is important to notice that, while independent learning improves algorithm complexity, it often does not work well when the task requires coordination among the agents \cite{matignon2012independent, gupta2017cooperative}. Many other factors can be used to classify MARL algorithms, such as the number of agents, the assumptions on the rewards, or the type of communication among the agents: for the sake of conciseness, we refer the reader to \cite{albrecht2024multi} for additional details. 

\begin{figure}[h!]
\centering
\includegraphics[width=\linewidth]{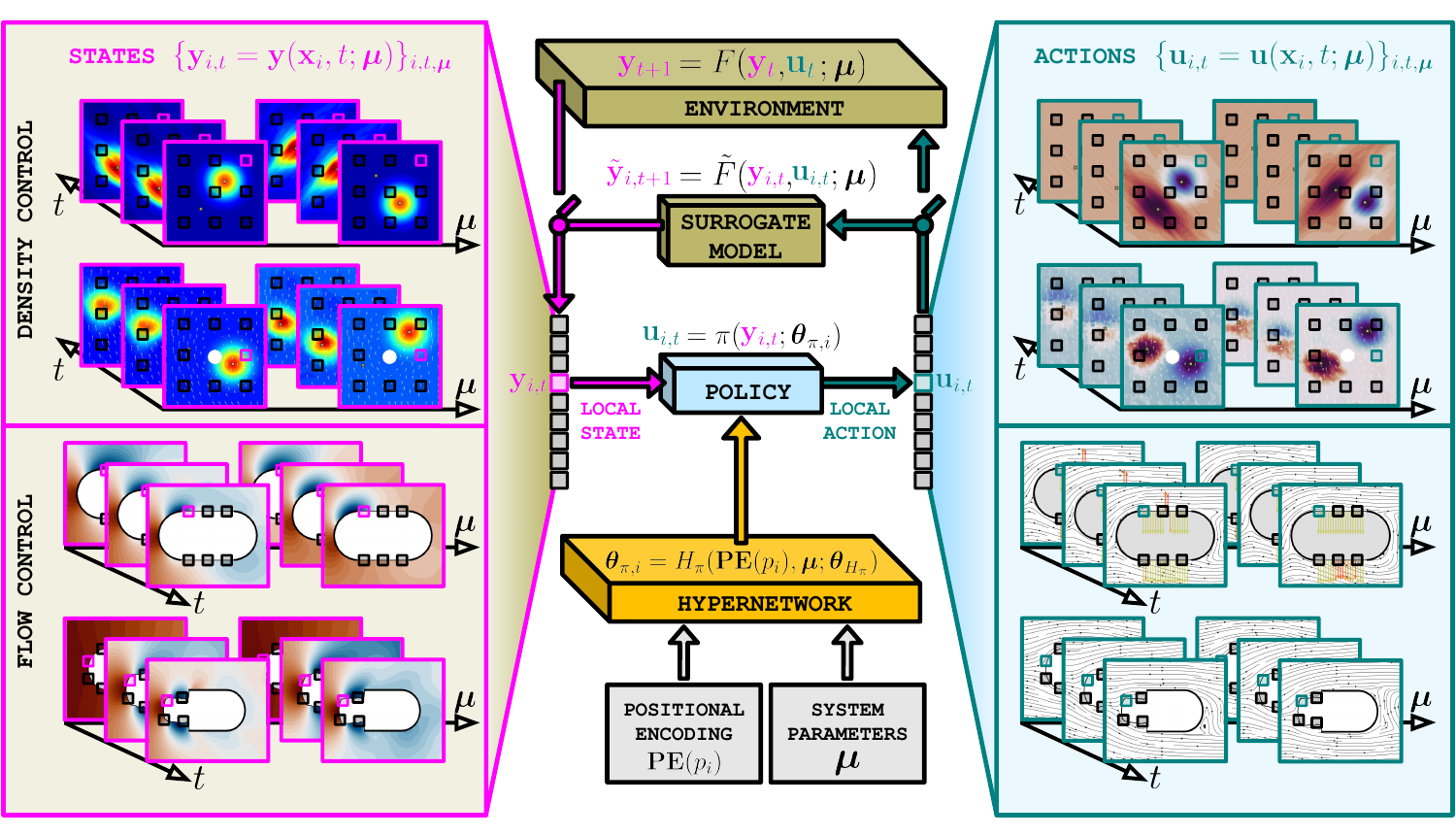}
\caption{HypeMARL for control of high-dimensional, parametric and distributed systems. Each agent relies on local rewards $r_{i,t}$ to learn the optimal local policy. After training, the local policy provides the optimal action for every agent over time $\mathbf{u}_{i,t} = \pi(\mathbf{y}_{i,t}; \boldsymbol{\theta}_{\pi,i})$, starting from the local state $\mathbf{y}_{i,t}$. To specialize the policy with respect to different agents, the policy parameters $\boldsymbol{\theta}_{\pi,i}$ are determined through a hypernetwork with respect to positional encoding of the agent's relative position $\mathbf{PE}(p_i)$ and the system parameters $\bm{\mu}$, that is $\boldsymbol{\theta}_{\pi,i} = H_\pi(\mathbf{PE}(p_i), \bm{\mu}; \boldsymbol{\theta}_H)$. While model-free HypeMARL interacts with the real environment $\mathbf{y}_{t+1} = F(\mathbf{y}_t, \mathbf{u}_t ; \bm{\mu})$, where $\mathbf{y}_t$ and $\mathbf{u}_t$ collect the local state and control values, respectively, the model-based version MB-HypeMARL learns a surrogate model approximating the local dynamics $\tilde{\mathbf{y}}_{i,t+1} = \tilde{F}(\mathbf{y}_{i,t}, \mathbf{u}_{i,t}; \bm{\mu})$ in order to enhance sample efficiency and speed up training.}
\label{fig:1}
\end{figure} 

In this research, we propose \emph{HypeMARL}: a novel decentralized MARL algorithm tailored to the control of high-dimensional, parametric, and distributed systems described in terms of nonlinear, time-dependent, partial differential equations (PDEs). Specifically, we employ \emph{hypernetworks} \cite{ha2016hypernetworks} to learn an effective parametrization of local policies and value functions with respect to the system parameters and the agents' relative positions. Hypernetworks are a class of NNs that provide the parameters, i.e. the weights and biases, of other NNs, often referred to as main or primary networks. Hypernetworks improve the expressivity and performance of NN-based architectures, with promising results in a wide variety of deep learning problems, including continual learning, causal inference, transfer learning, weight pruning, uncertainty quantification, zero-shot learning, natural language processing, and recently DRL \cite{chauhan2023brief}. Hypernetworks in DRL were first exploited in \cite{sarafian2021recomposing} to learn the parameters of the NNs modeling the value function or the policy. In the context of meta RL, zero-shot RL, continual RL, and control of parametric PDEs, hypernetworks have been employed by \cite{beck2023hypernetworks, rezaei2023hypernetworks,huang2021continual, botteghi2025hyperlparameterinformedreinforcementlearning} to improve the training and generalization performance of RL agents. In the context of decentralized MARL, instead, hypernetworks are an effective way to achieve parameter sharing, which is an important strategy for improving the agents' performance and scaling to large numbers of agents by significantly reducing the number of learnable parameters and consequently training times \cite{albrecht2024multi, tan1993multi, christianos2021scaling, terry2020revisiting}.

A summary of the HypeMARL strategy is provided in Figure~\ref{fig:1}. In particular, we utilize the knowledge of the system parameter $\boldsymbol{\mu}$ and the relative position of each agent with respect to the others $p_i$ as input to two hypernetworks aiming at learning the parameters of each agent's policy and value function, respectively. The relative position of each agent is provided using \emph{sinusoidal position encoding} \cite{vaswani2017attention}, as taken into account in natural language processing to encode the position of each word in a sentence. 
The encoding via hypernetworks of the physical system information and of the inter-agents relations enables effective coordination and enhances generalization capabilities. Besides the model-free HypeMARL strategy, we also propose a model-based extension, named \emph{model-based HypeMARL (MB-HypeMARL)}, where we learn a computationally efficient deep learning-based surrogate model approximating the local environment dynamics, resulting in faster training procedures with minimal deterioration of the policy performance. To assess the performance of the proposed algorithms, we consider different high-dimensional, parametric, and distributed control problems, namely \emph{density control} in a vacuum and in a fluid, and \emph{flow control} with different actuation strategies, such as jets and shape-morphing obstacles.

\section{Hypernetwork-based positional encoding for multi-agent reinforcement learning}
\label{sec:methods}

Let us consider a potentially high-dimensional, nonlinear, time-dependent, and parametric dynamical system discretized through suitable time and space discretization schemes
\begin{equation}
    \mathbf{y}_{t+1} = F(\mathbf{y}_t, \mathbf{u}_t; \bm{\mu})
\end{equation}
where $\mathbf{y}_t$ and $\mathbf{y}_{t+1} \in \mathbb{R}^{N_y}$ represent, respectively, the system state at time step $t$ and $t+1$, with $N_y$ indicating the dimension of the state and $t=1,...,T_{\max}$ the time steps, and $\mathbf{u}_t \in\mathbb{R}^{N_u}$ denotes the $N_u$-dimensional control input at time step $t$, $F:\mathbb{R}^{N_y}\times\mathbb{R}^{N_u}\times\mathbb{R}^{N_\mu}\to\mathbb{R}^{N_y}$ is the state transition map, and $\boldsymbol{\mu}\in \mathbb{R}^{N_\mu}$ is a $N_\mu$-dimensional parameter vector belonging to a suitable parameter space $\mathbb{R}^{N_\mu}$. The dynamical system may be high-dimensional in state and control dimensions, i.e. $N_y \gg 1$ and $N_u \gg 1$, making the use of single-agent RL unfeasible. Therefore, we cast the control problem of such a system as a MARL problem in which each agent relies solely on local state measurements and on the knowledge of the system parameters $\boldsymbol{\mu}$ to determine its actions. Similarly to \cite{botteghi2024parametric}, the policy and value function are explicitly dependent on the parameter vector $\boldsymbol{\mu}$ to enhance the agent's performance when dealing with systems showing parametric dependencies.

Instead of learning a global policy $\mathbf{u}_t = \pi(\mathbf{y}_t; \bm{\mu})$, we learn a local policy $\mathbf{u}_{i,t}= \pi(\mathbf{y}_{i,t}; \boldsymbol{\theta_{\pi,i}})$, obtained as specializations of the policy $\pi$ with respect to the local measurements of the state $\mathbf{y}_i$ and with parameters $\boldsymbol{\theta_{\pi_i}}$ dependent on the different relative agent positions $p_i$ and system parameters $\boldsymbol{\mu}$. While many positional encoding strategies have been proposed \cite{gehring2017convolutional}, in this work we focus on sinusoidal positional encoding \cite{vaswani2017attention}. Doing so, the relative position of an agent $p_i$ is embedded into a higher-dimensional vector $\mathbf{PE}$ using sine and cosine functions at different frequencies
\begin{equation}
    \mathbf{PE}(p_i) = \left[\sin\left(\frac{p_i}{\omega_1}\right), \cos\left(\frac{p_i}{\omega_1}\right), \sin\left(\frac{p_i}{\omega_2}\right), \dots, \sin\left(\frac{p_i}{\omega_d}\right), \cos\left(\frac{p_i}{\omega_{d}}\right)\right]\, ,
    \label{eq:sin_positional_encoding}
\end{equation}
where $\omega_j=n^{2j/d}$ for $1 \leq j \leq d$, with $d$ being the dimension of the positional embedding vector and $n$ a constant value. In this work, we set $d=2048$ and $n=1000$. 
The knowledge of the relative position is crucial to specialize the independent agents even when observing the same state values and to achieve coordinated strategies even though relying solely on local information. The positional encoding vector $\mathbf{PE}(p_i)$ may be concatenated to the local state $\mathbf{y}_{i}^{t}$ and the system parameters $\boldsymbol{\mu}$, and provided as input to the agent's policy and value function. However, following the work in \cite{botteghi2025hyperlparameterinformedreinforcementlearning}, we propose to split the state and parameter dependencies through the use of two hypernetworks, namely $H_\pi$ and $H_Q$, which parametrize the weights and biases of the policy and value function with respect to positional encoding $\mathbf{PE}(p_i)$ and system parameters $\boldsymbol{\mu}$. Specifically, the hypernetworks $H_{\pi}:
\mathbb{R}^{d + N_{\bm{\mu}}}\to \mathbb{R}^{N_{\boldsymbol{\theta}_{\pi}}}$ and $H_Q:
\mathbb{R}^{d + N_{\bm{\mu}}}\to \mathbb{R}^{N_{\boldsymbol{\theta}_{Q}}}$ learn the parameters $\boldsymbol{\theta}_{\pi,i}$ and $\boldsymbol{\theta}_{Q, i}$ of the NNs modeling the $i^{\text{th}}$-local policy and value function, respectively. The parameters of the local policies and the hypernetworks, namely $\boldsymbol{\theta}_{H_{\pi}}$ and $\boldsymbol{\theta}_{H_{Q}}$ are jointly updated by minimizing a prescribed loss function. 

HypeMARL relies on the twin-delayed deep deterministic policy gradient algorithm (TD3) \cite{fujimoto2018addressing} as the core RL algorithm. TD3 is an actor-critic method that learns a deterministic policy while estimating the value function. See Appendix~\ref{sec:TD3} for further details on the TD3 algorithm. 
 Differently from its model-free counterpart, MB-HypeMARL exploits a local surrogate model of the dynamics $\mathbf{y}_{i,t+1} \approx \tilde{\mathbf{y}}_{i,t+1} = \tilde{F}(\mathbf{y}_{i,t}, \mathbf{u}_{i,t}; \boldsymbol{\mu})$ to generate additional samples while learning policy and value function, drastically reducing the interactions with the real environment, as shown in Section~\ref{sec:results}.

Model-free HypeMARL relies only on local information to learn the policy and the value function through the TD3 algorithm. In particular, combining hypernetworks and positional encoding, we can write the local policy of the $i^{\text{th}}$-agent as
\begin{equation}
    \begin{split}
        \boldsymbol{\theta}_{\pi.i} &= H_\pi(\mathbf{PE}(p_i),\boldsymbol{\mu}; \boldsymbol{\theta}_{H_\pi}) \\
        \mathbf{u}_{i,t} &= \pi(\mathbf{y}_{i,t}; \boldsymbol{\theta}_{\pi,i})\, ,
    \end{split}
    \label{eq:hyperpolicy}
\end{equation}
and the local value function as
\begin{equation}
    \begin{split}
        \boldsymbol{\theta}_{Q,i} &= H_Q(\mathbf{PE}(p_i),\boldsymbol{\mu}; \boldsymbol{\theta}_{H_Q}) \\
        q_{i,t} &= Q(\mathbf{y}_{i,t}, \mathbf{u}_{i,t}; \boldsymbol{\theta}_{Q,i}).
    \end{split}
    \label{eq:hypervalue}
\end{equation}
In the context of MARL, hypernetworks can be viewed as an effective form of parameter sharing. Indeed, the two hypernetworks in Equations~\eqref{eq:hyperpolicy} and \eqref{eq:hypervalue} are exploited to specialize the weights and biases of the policy and value function for every agent. It is worth mentioning that the hypernetwork-based parametrization of policy and value function can be used in combination with any RL algorithm, without changing its core working principles. 


In general, model-free MARL can be quite sample inefficient. Sample efficiency may be a limiting factor for very complex systems where only limited data can be collected and their simulation is extremely computationally expensive. 
To mitigate this issue, we propose MB-HypeMARL, aiming to reduce the number of environment interactions. MB-HypeMARL learns a surrogate model of the local dynamics $\tilde{F}$ that can be autoregressively queried to predict the local evolution of the system
\begin{equation}
        \tilde{\mathbf{y}}_{i,t+1} = \tilde{F}(\tilde{\mathbf{y}}_{i,t}, \mathbf{u}_{i,t}; \boldsymbol{\mu})\, ,
    \label{eq:model_forward}
\end{equation}
where we indicate with $\tilde{\mathbf{y}}_{i,t+1}$ and $\tilde{\mathbf{y}}_{i,t}$ the states predicted by the surrogate model $\tilde{F}$ to distinguish them from the real-environment states $\mathbf{y}_{i,t+1}$ and $\mathbf{y}_{i,t}$. Specifically, we approximate $\tilde{F}(\tilde{\mathbf{y}}_{i,t}, \mathbf{u}_{i,t}; \boldsymbol{\mu})=\tilde{F}(\tilde{\mathbf{y}}_{i,t}, \mathbf{u}_{i,t}, \boldsymbol{\mu}; \boldsymbol{\theta}_{\tilde{F}})$ with a shallow neural network of parameters $\boldsymbol{\theta}_{\tilde{F}}$.  Using the data collected from the real-environment interaction, the parameters of the local forward model $\boldsymbol{\theta}_{\tilde{F}}$ are optimized by minimizing the prediction error between the samples from the environment $\mathbf{y}_{t+1}$ and the surrogate model predictions $\tilde{\mathbf{y}}_{t+1}$, that is the loss function
\begin{equation}
\begin{split}
        \mathcal{L}(\boldsymbol{\theta}_{\tilde{F}})
        &= \mathbb{E}_{\mathbf{y}_t, \mathbf{u}_t, \mathbf{y}_{t+1}, \boldsymbol{\mu}}\left[\sum_{i=1}^N ||\mathbf{y}_{i,t+1} - \tilde{\mathbf{y}}_{i,t+1} ||_2^2\right] \\
        &= \mathbb{E}_{\mathbf{y}_t, \mathbf{u}_t, \mathbf{y}_{t+1}, \boldsymbol{\mu}}\left[\sum_{i=1}^N ||\mathbf{y}_{i,t+1} - \tilde{F}(\mathbf{y}_{i,t}, \mathbf{u}_{i,t}, \boldsymbol{\mu}; \boldsymbol{\theta}_{\tilde{F}}) ||_2^2\right]. \\
\end{split}
\end{equation}


The pseudo-code of the two variants of HypeMARL, namely model-free HypeMARL and model-based HypeMARL (MB-HypeMARL), are summarized in Algorithm~\ref{alg:model_free_model_based_MARL_pseudo_code}.

\begin{center}
\begin{minipage}{0.7\linewidth}
\begin{algorithm}[H]
    \centering
    \caption{HypeMARL and {\color{magenta} MB-HypeMARL}}\label{alg:model_free_model_based_MARL_pseudo_code}
    \begin{algorithmic}[1]
        \State Initialize models' parameters and noise variance $\boldsymbol{\sigma}^2$
        \State Set maximum number of episodes $E_{\max}$ and time steps $T_{\max}$
        \For{$e=1:E_{\max}$}
            \State Sample $\boldsymbol{\mu}$ 
            \State Get initial measurement $\mathbf{y}_{i,1}$ for $i=1:N$
            \State {\color{magenta} Set $\tilde{\mathbf{y}}_{i,1}=\mathbf{y}_{i,1}$ for $i=1:N$}
            \For{$t=1:T_{\max}$}
                    \State Sample $\boldsymbol{\theta}_{\pi,i} = H_\pi(\mathbf{PE}(p_i),\boldsymbol{\mu}; \boldsymbol{\theta}_{H_\pi})$ for $i = 1:N$
                \If{interact with real environment} 
                    \State Sample $\mathbf{u}_{i,t} \sim \pi(\mathbf{y}_{i,t}; \boldsymbol{\theta}_{\pi,i}) + \boldsymbol{\epsilon}$, where $ \boldsymbol{\epsilon} \sim \mathcal{N}(\bm{0}, \boldsymbol{\sigma}^2)$ for $i = 1:N$
                    \State {\color{black}Apply joint action $\mathbf{u}_t=[\mathbf{u}_{1,t},\dots,\mathbf{u}_{N,t}]^{\top}$} 
                    \State {\color{black}Observe $\mathbf{y}_{i,t+1}=F(\mathbf{y}_t, \mathbf{u}_t,\boldsymbol{\mu})$ for $i=1:N$} 
                    \State {\color{black}Compute local rewards $r_{i,t}$ for $i=1:N$}
                    \If{train models}
                        \State {\color{black} Update $\pi$ and $Q$ using the tuple $(\mathbf{y}_{i,t}, \mathbf{u}_{i,t}, r_{i,t}, \mathbf{y}_{i,t+1},   \boldsymbol{\mu})_{i=1:N}$}
                        \State {\color{magenta} Update $\tilde{F}$ using the tuple $(\mathbf{y}_{i,t}, \mathbf{u}_{i,t}, r_{i,t}, \mathbf{y}_{i,t+1},   \boldsymbol{\mu})_{i=1:N}$}
                    \EndIf
                \ElsIf{interact with surrogate environment }
                    \State {\color{magenta}Sample $\mathbf{u}_{i,t} \sim \pi(\tilde{\mathbf{y}}_{i,t}; \boldsymbol{\theta}_{\pi,i}) + \boldsymbol{\epsilon}$, where $ \boldsymbol{\epsilon} \sim \mathcal{N}(\bm{0}, \boldsymbol{\sigma}^2)$ for $i = 1:N$}
                    \State {\color{magenta}Apply joint action $\mathbf{u}_t=[\mathbf{u}_{1,t},\dots,\mathbf{u}_{N,t}]$}
                    \State {\color{magenta}Observe $\tilde{\mathbf{y}}_{i,t+1}=\tilde{F}(\tilde{\mathbf{y}}_{i,t}, \mathbf{u}_{i,t},\boldsymbol{\mu}; \boldsymbol{\theta}_{\tilde{F}})$ for $i=1:N$}
                    \State {\color{magenta}Compute rewards $\tilde{r}_{i,t}$ for $i=1,...,N$}
                    \If{train models}
                        \State {\color{magenta} Update $\pi$ and $Q$ using the tuple $(\tilde{\mathbf{y}}_{i,t}, \mathbf{u}_{i,t}, \tilde{r}_{i,t}, \tilde{\mathbf{y}}_{i,t+1},   \boldsymbol{\mu})_{i=1:N}$}
                    \EndIf
                \EndIf
            \EndFor
        \EndFor
    \end{algorithmic}
\end{algorithm}
\end{minipage}
\end{center}

\section{Numerical results}\label{sec:results}
In this section, we present the numerical results obtained when applying HypeMARL to four high-dimensional, parametric, and distributed control problems. Section~\ref{subsec:densitycontrol} presents two applications dealing with \emph{density control}, while Section~\ref{subsec:flowcontrol} presents two applications on \emph{flow control} problems with different distributed actuation strategies. It is worth highlighting that, due to the high-dimensionality of these problems in terms of states and, especially, controls, a single-agent RL algorithm is not a viable option, as shown in the Appendix~\ref{sec:RLcomparison}. In recent years, decentralized MARL has shown promising results in controlling distributed systems governed by partial differential equations \cite{vignon2023effective, suarez2025active, peitz2023distributed}, and it will be used as a baseline in the following sections to compare with HypeMARL and MB-HypeMARL. Differently from HypeMARL, decentralized MARL does not take into account positional encoding and hypernetworks, with local policies that predict local control actions relying solely on local state values and system parameters. It is important to note that HypeMARL requires minimal hyperparameter tuning. Indeed, all the test cases presented take into account the same architectures for NNs and positional encoding, as described in Section~\ref{sec:methods} and Appendix~\ref{sec:hyperparam}, and the same training setting.


\subsection{Density control}
\label{subsec:densitycontrol}
This section presents two challenging test cases in which the aim is to move a Gaussian density from a random initial mean position to a desired target position by optimally steering the source of a diffusion-advection PDE. In particular, inspired by \cite{tomasetto2024latent}, we consider two settings: density control in a vacuum and in a parametric fluid. While the former case does not consider an underlying transport effect, the latter deals with a fluid moving the density upwards. Density control may be of interest in several scientific fields. Indeed, density functions serve as macroscopic models of different quantities, such as, e.g., drone or wildlife swarms, traffic, and concentrations of pollutants or chemicals, to mention a few.

\subsubsection{Density control in a vacuum}

In this section, we present the first test case dealing with density control in a vacuum, i.e. we do not take into account an underlying transport effect, such as a wind or a water current, moving the density. Let $y(\mathbf{x},t) \in \mathbb{R}$ be the state variable representing a density function in the space-time domain $\Omega \times (0,T]$, with final time $T>0$. The state dynamics is described by the Fokker-Planck equation
\begin{equation}
\label{eq:density_vacuum}
\begin{cases}
      \dfrac{\partial y}{\partial t} - \kappa \Delta y  = u \quad
      & \text{in} \ \Omega \times (0,T]
      \\
      \nabla y \cdot \mathbf{n} = 0
      \quad
      & \text{on} \ \partial \Omega  \times (0,T]
      \\
       y(\mathbf{x}, 0) = y_0(\mathbf{x}; \bm{\mu}_0)
       \quad
       & \text{in} \ \Omega\, ,
\end{cases}
\end{equation}
where $\Omega=(-1, 1)^2$ is the spatial domain, $  \kappa=0.001$ is the diffusion coefficient, $\mathbf{n}$ is the outward normal versor to the boundary $\partial \Omega$, and $u(\mathbf{x},t) \in [-5,5] \subset \mathbb{R}$ is the external source term. Starting from a Gaussian density with fixed variance equal to $0.05$ and random mean $\boldsymbol{\mu}_0 =(\mu_{0,1}, \mu_{0,2})$, that is
\begin{equation}
    y_0(\mathbf{x}; \bm{\mu}_0) = \frac{10}{\pi}\exp(-10(x_1 - \mu_{0,1})^2 - 10(x_2 -\mu_{0,2})^2)\, ,
\end{equation}
where $\mathbf{x} = (x_1, x_2) \in \Omega$, the density slightly diffuses over time. The goal in this setting is to move the density from a random initial position $\boldsymbol{\mu}_0 \in [-0.75,0]\times[-0.75,0.75]$ to a desired final destination $\boldsymbol{\mu}_T = (\mu_{T,1}, \mu_{T,2}) \in [0,0.75]\times[-0.75,0.75]$. To do so, we control the term $u$ in Equation~\eqref{eq:density_vacuum} over the whole domain $\Omega$, which can be interpreted as a source or a sink effect. Notice that only the target destination selected by the controller is regarded as a system parameter, i.e. $\bm{\mu} = \bm{\mu}_T$, while the initial position $\bm{\mu}_0$ is implicitly captured by the local states observed in the algorithm. Notice also that, after discretizing the spatial domain with a regular mesh, the dimensionality of the state and the control is equal to $N_y = N_u = 1089$, thus resulting in a high-dimensional and distributed control problem. In addition, to solve the Fokker-Planck equation with the finite element method, we utilize a uniform time discretization of the interval $[0, T]$ with time step $\Delta t= 0.1 \text{s}$ and final time $T=1\text{s}$.

\begin{figure*}
    \centering

    \subfloat{\includegraphics[height=0.26\textwidth]{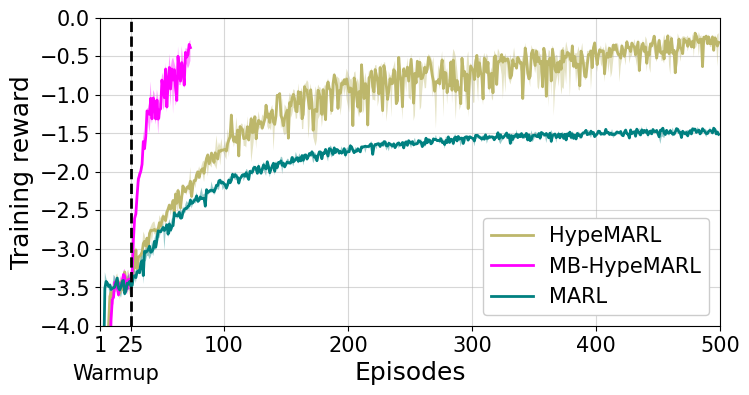}}
    \subfloat{\includegraphics[height=0.26\textwidth]{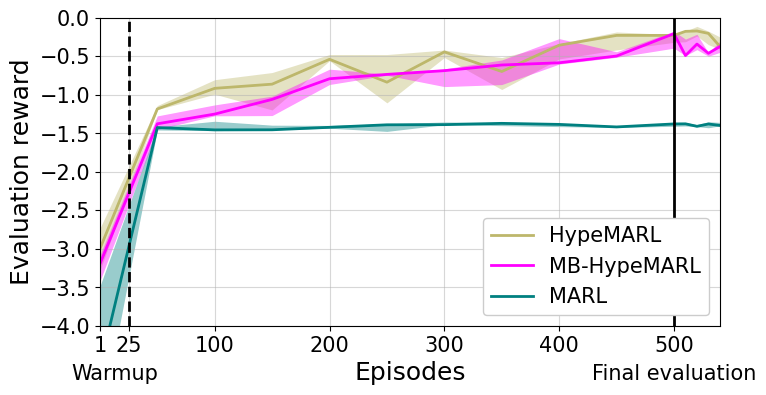}}
    
    \caption{{\em Density control in a vacuum}. Median and interquartile range of HypeMARL, MB-HypeMARL and MARL training end evaluation rewards.}
    \label{fig:FP_Vacuum1}
\end{figure*}

We employ MARL, HypeMARL and MB-HypeMARL to learn a parametric policy capable of controlling this challenging high-dimensional, parametric, and distributed setting. The local rewards $r_{i,t}$ maximized while training the local agents are defined as
\begin{equation}
\label{eq:rewards_densitycontrol}
    r_{i,t} = r(y_{i,t}, u_{i,t}; \boldsymbol{\mu}) = -(y_{i,t} - y_{i,T}(\boldsymbol{\mu}))^2 ,
\end{equation}
where $y_{i,T}(\boldsymbol{\mu}) = y_T(\mathbf{x};\bm{\mu})$ corresponds to the local values of the target Gaussian density
\begin{equation}
    y_T(\mathbf{x};\bm{\mu}_T) = \frac{10}{\pi}\exp(-10(x_1 - \mu_{T,1})^2 - 10(x_2 -\mu_{T,2})^2)\, ,
\end{equation}
centered at the final target destination. Figure~\ref{fig:FP_Vacuum1} reports the average local rewards during training and evaluation of the different MARL algorithms with respect to the number of training episodes. In particular, we display the median and interquartile range of $5$ independent runs with $5$ different seeds for a fair and robust comparison. We train the algorithms for a total of $500$ episodes, exploring the action space with additive Gaussian noise, which linearly decays after the first $25$ warm-up interactions, on the actions predicted by the policy. Regarding the evaluations, we test the policy during training every $50$ episodes, and we evaluate the optimal policy after training on $5$ different combinations of initial positions $\bm{\mu}_0$ and final destination $\bm{\mu}=\bm{\mu}_T$. When considering MB-HypeMARL, to highlight the data efficiency with respect to model-free algorithms, we only report the training rewards collected when interacting with the environment, leaving out the interactions with the surrogate environment. In particular, after the first $25$ episodes of warm-up, MB-HypeMARL performs $10$ episodes of interaction with the surrogate environment for every episode of interaction with the real environment. Both HypeMARL and MB-HypeMARL outperform MARL in terms of expected cumulative reward during training and evaluation. In addition, while $500$ environment interactions are required to optimally steer the system with HypeMARL, MB-HypeMARL is capable of learning effective policies with only $73$ environment interactions.

\begin{figure*}
    \centering    
    \begin{sideways} \makebox[0pt][l]{\hspace{-1.1cm} \shortstack[c]{{\bf \footnotesize No control} \\ {\footnotesize State}}} \end{sideways}
    \begin{minipage}{0.33\linewidth}
    \centering
    \hspace{-0.17cm}
    \subfloat[$t=0.0\text{s}$]{\begin{overpic}[width=0.4\textwidth]{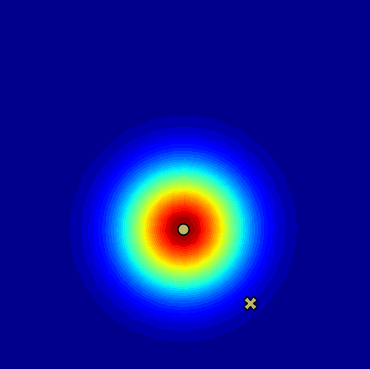}\put(40,130){\shortstack[c]{$\bm{\mu}_0 = (-0.01,-0.24)$ \\ $\bm{\mu}_T = (0.35,-0.64)$}}\end{overpic}}
    \hspace{0.005cm}
    \subfloat[$t=0.5\text{s}$]{\includegraphics[width=0.4\linewidth]{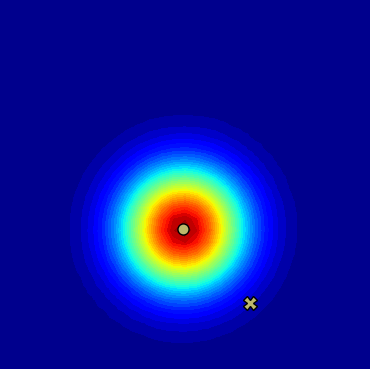}}
    \subfloat{\includegraphics[height=0.4\linewidth]{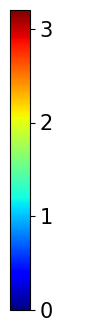}}
    \end{minipage}
    \hspace{-0.5cm}
    \begin{minipage}{0.33\linewidth}
    \centering
    \hspace{-0.17cm}
    \subfloat[$t=0.0\text{s}$]{\begin{overpic}[width=0.4\textwidth]{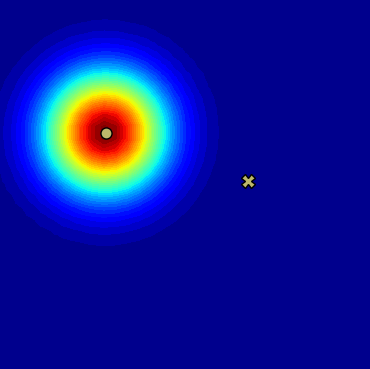}\put(40,130){\shortstack[c]{$\bm{\mu}_0 = (-0.43,0.28)$ \\ $\bm{\mu}_T = (0.34,0.02)$}}\end{overpic}}
    \hspace{0.005cm}
    \subfloat[$t=0.5\text{s}$]{\includegraphics[width=0.4\linewidth]{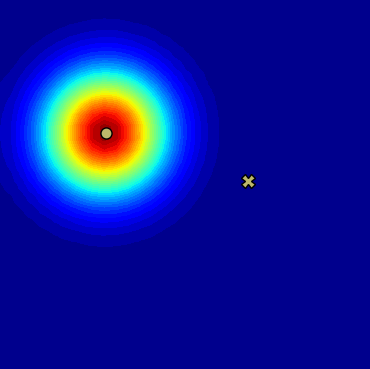}}
    \subfloat{\includegraphics[height=0.4\linewidth]{Figures/FP_Vacuum/Colorbar_State_Uncontrolled.png}}
    \end{minipage}
    \hspace{-0.5cm}
    \begin{minipage}{0.33\linewidth}
    \centering
    \hspace{-0.17cm}
    \subfloat[$t=0.0\text{s}$]{\begin{overpic}[width=0.4\textwidth]{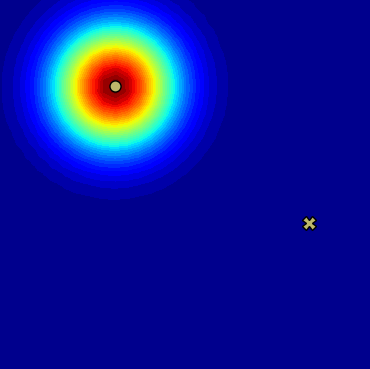}\put(40,130){\shortstack[c]{$\bm{\mu}_0 = (-0.38,0.53)$ \\ $\bm{\mu}_T = (0.67,-0.21)$}}\end{overpic}}
    \hspace{0.005cm}
    \subfloat[$t=0.5\text{s}$]{\includegraphics[width=0.4\linewidth]{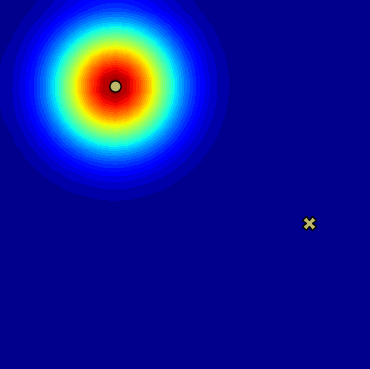}}
    \subfloat{\includegraphics[height=0.4\linewidth]{Figures/FP_Vacuum/Colorbar_State_Uncontrolled.png}}
    \end{minipage}
    \vspace{-0.3cm}

    \begin{sideways} \makebox[0pt][l]{\hspace{-3cm} \shortstack[c]{{\bf \footnotesize HypeMARL} \\ {\footnotesize Action \hspace{1.3cm} State}}} \end{sideways}
    \begin{minipage}{0.33\linewidth}
    \centering
    \subfloat{\includegraphics[width=0.4\textwidth]{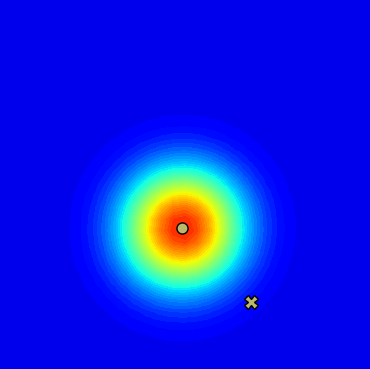}}
    \hspace{0.005cm}
    \subfloat{\includegraphics[width=0.4\linewidth]{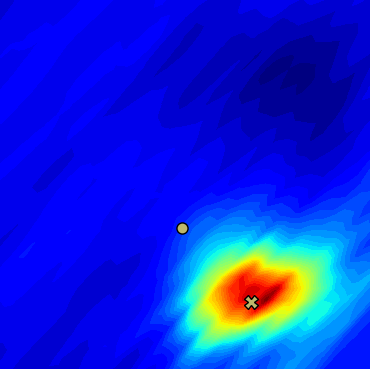}}
    \subfloat{\includegraphics[height=0.4\linewidth]{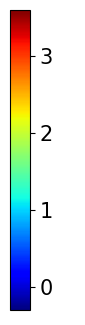}}
    \end{minipage}
    \hspace{-0.5cm}
    \begin{minipage}{0.33\linewidth}
    \centering
    \subfloat{\includegraphics[width=0.4\textwidth]{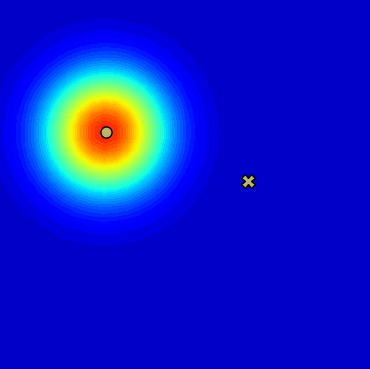}}
    \hspace{0.005cm}
    \subfloat{\includegraphics[width=0.4\linewidth]{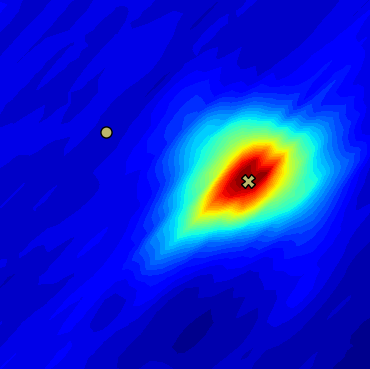}}
    \subfloat{\includegraphics[height=0.4\linewidth]{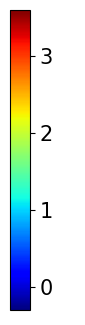}}
    \end{minipage}
    \hspace{-0.5cm}
    \begin{minipage}{0.33\linewidth}
    \centering
    \subfloat{\includegraphics[width=0.4\textwidth]{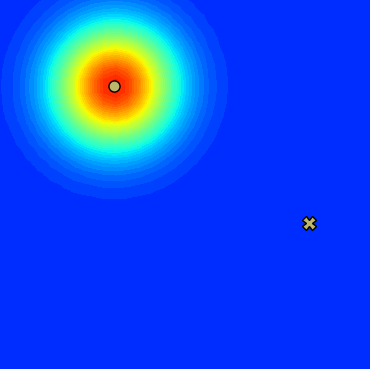}}
    \hspace{0.005cm}
    \subfloat{\includegraphics[width=0.4\linewidth]{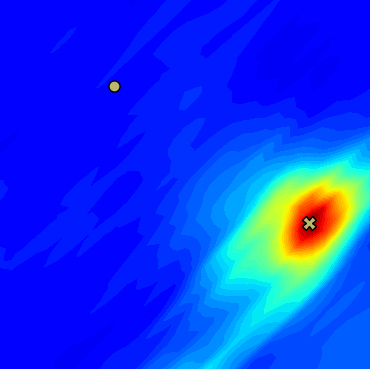}}
    \subfloat{\includegraphics[height=0.4\linewidth]{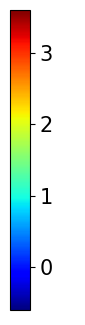}}
    \end{minipage}
    \vspace{-0.3cm}

    \begin{sideways} \makebox[0pt][l]{\hspace{-1cm} \shortstack[c]{{\bf \footnotesize \phantom{HypeMARL}} \\ {\footnotesize \phantom{Action}}}} \end{sideways}
    \begin{minipage}{0.33\linewidth}
    \centering
    \subfloat{\includegraphics[width=0.4\linewidth]{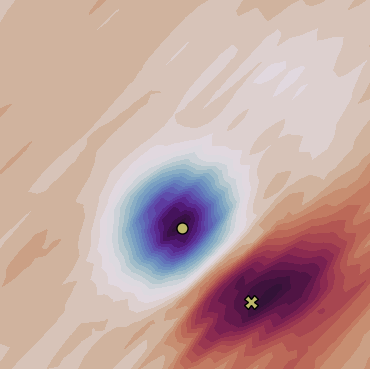}}
    \hspace{0.005cm}
    \subfloat{\includegraphics[width=0.4\linewidth]{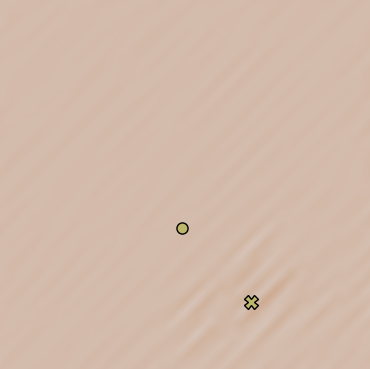}}
    \subfloat{\includegraphics[height=0.4\linewidth]{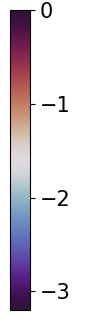}}
    \end{minipage}
    \hspace{-0.5cm}
    \begin{minipage}{0.33\linewidth}
    \centering
    \subfloat{\includegraphics[width=0.4\linewidth]{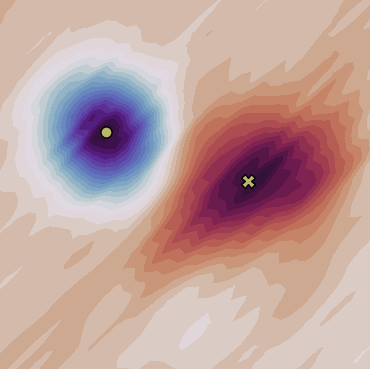}}
    \hspace{0.005cm}
    \subfloat{\includegraphics[width=0.4\linewidth]{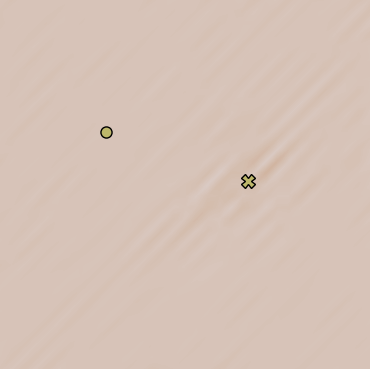}}
    \subfloat{\includegraphics[height=0.4\linewidth]{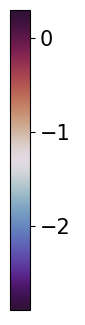}}
    \end{minipage}
    \hspace{-0.5cm}
    \begin{minipage}{0.33\linewidth}
    \centering
    \subfloat{\includegraphics[width=0.4\linewidth]{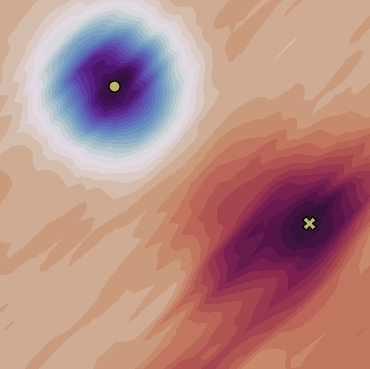}}
    \hspace{0.005cm}
    \subfloat{\includegraphics[width=0.4\linewidth]{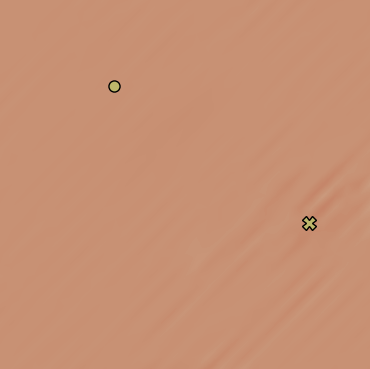}}
    \subfloat{\includegraphics[height=0.4\linewidth]{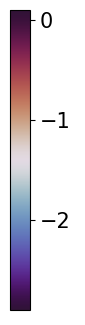}}
    \end{minipage}
    \vspace{-0.3cm}

    \begin{sideways} \makebox[0pt][l]{\hspace{-3cm} \shortstack[c]{{\bf \footnotesize MB-HypeMARL} \\ {\footnotesize Action \hspace{1.5cm} State}}} \end{sideways}
    \begin{minipage}{0.33\linewidth}
    \centering
    \subfloat{\includegraphics[width=0.4\linewidth]{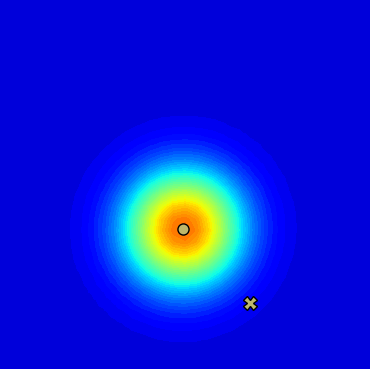}}
    \hspace{0.005cm}
    \subfloat{\includegraphics[width=0.4\linewidth]{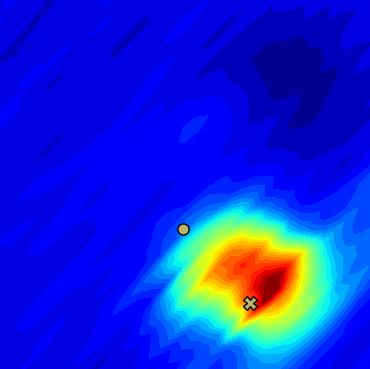}}
    \subfloat{\includegraphics[height=0.4\linewidth]{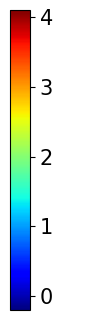}}
    \end{minipage}
    \hspace{-0.5cm}
    \begin{minipage}{0.33\linewidth}
    \centering
    \subfloat{\includegraphics[width=0.4\linewidth]{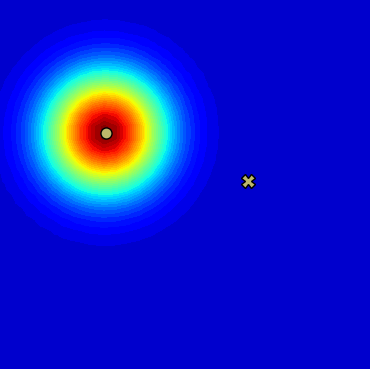}}
    \hspace{0.005cm}
    \subfloat{\includegraphics[width=0.4\linewidth]{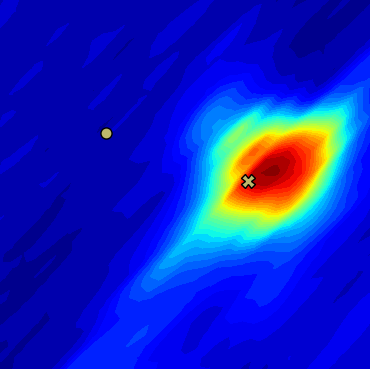}}
    \subfloat{\includegraphics[height=0.4\linewidth]{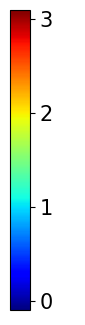}}
    \end{minipage}
    \hspace{-0.5cm}
    \begin{minipage}{0.33\linewidth}
    \centering
    \subfloat{\includegraphics[width=0.4\linewidth]{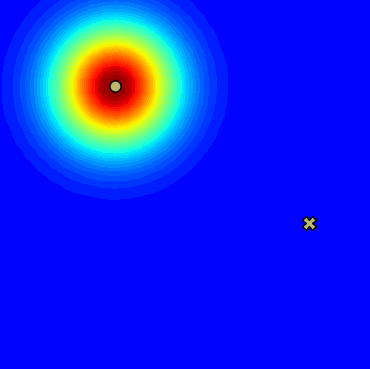}}
    \hspace{0.005cm}
    \subfloat{\includegraphics[width=0.4\linewidth]{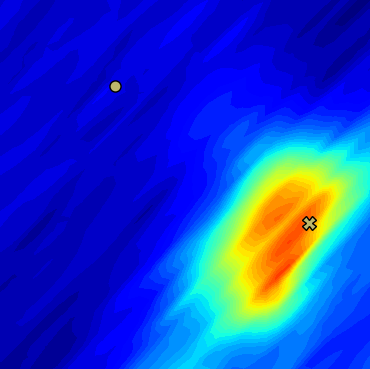}}
    \subfloat{\includegraphics[height=0.4\linewidth]{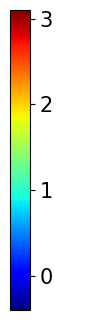}}
    \end{minipage}
    \vspace{-0.3cm}

    \begin{sideways} \makebox[0pt][l]{\hspace{-1.3cm} \shortstack[c]{{\bf \footnotesize \phantom{MB-HypeMARL}} \\ {\footnotesize \phantom{Action}}}} \end{sideways}
    \begin{minipage}{0.33\linewidth}
    \centering
    \subfloat{\includegraphics[width=0.4\linewidth]{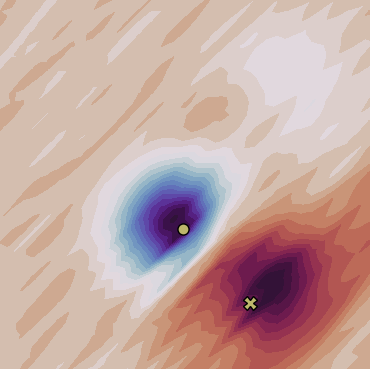}}
    \hspace{0.005cm}
    \subfloat{\includegraphics[width=0.4\linewidth]{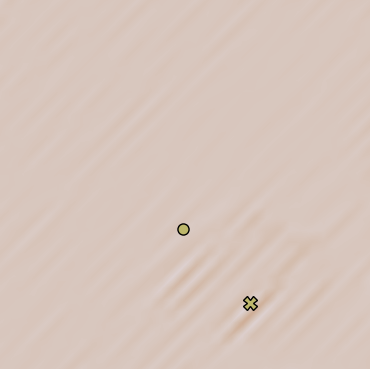}}
    \subfloat{\includegraphics[height=0.4\linewidth]{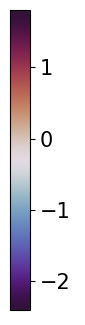}}
    \end{minipage}
    \hspace{-0.5cm}
    \begin{minipage}{0.33\linewidth}
    \centering
    \subfloat{\includegraphics[width=0.4\linewidth]{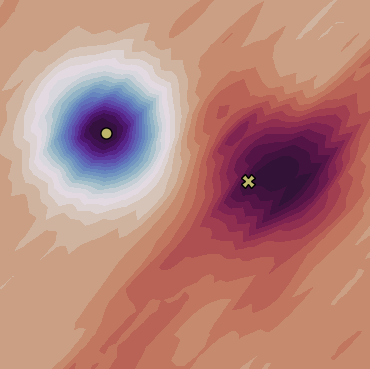}}
    \hspace{0.005cm}
    \subfloat{\includegraphics[width=0.4\linewidth]{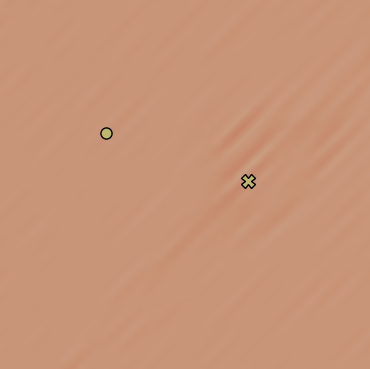}}
    \subfloat{\includegraphics[height=0.4\linewidth]{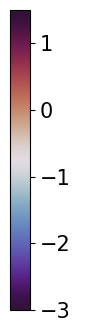}}
    \end{minipage}
    \hspace{-0.5cm}
    \begin{minipage}{0.33\linewidth}
    \centering
    \subfloat{\includegraphics[width=0.4\linewidth]{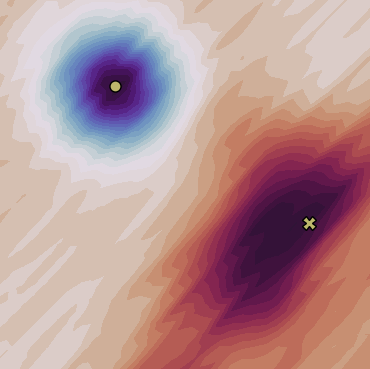}}
    \hspace{0.005cm}
    \subfloat{\includegraphics[width=0.4\linewidth]{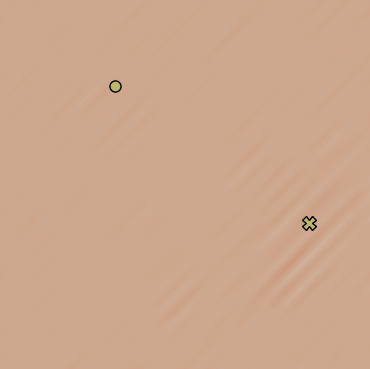}}
    \subfloat{\includegraphics[height=0.4\linewidth]{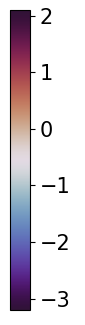}}
    \end{minipage}
    \vspace{-0.3cm}

    \begin{sideways} \makebox[0pt][l]{\hspace{-3cm} \shortstack[c]{{\bf \footnotesize \phantom{Hyper}MARL\phantom{Hyper}} \\ {\footnotesize Action \hspace{1.5cm} State}}} \end{sideways}
    \begin{minipage}{0.33\linewidth}
    \centering
    \subfloat{\includegraphics[width=0.4\linewidth]{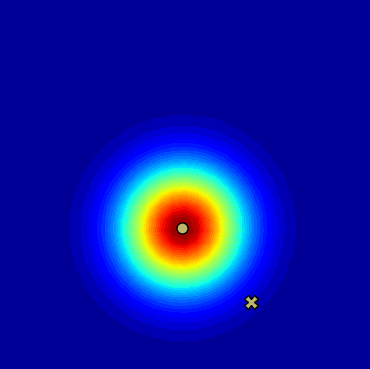}}
    \hspace{0.005cm}
    \subfloat{\includegraphics[width=0.4\linewidth]{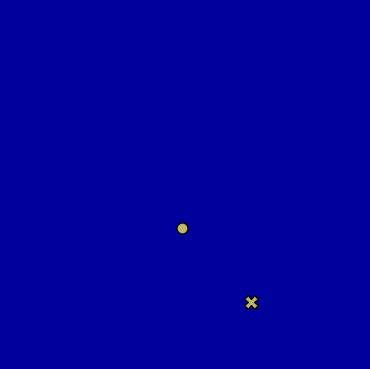}}
    \subfloat{\includegraphics[height=0.4\linewidth]{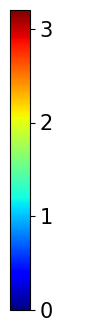}}
    \end{minipage}
    \hspace{-0.5cm}
    \begin{minipage}{0.33\linewidth}
    \centering
    \subfloat{\includegraphics[width=0.4\linewidth]{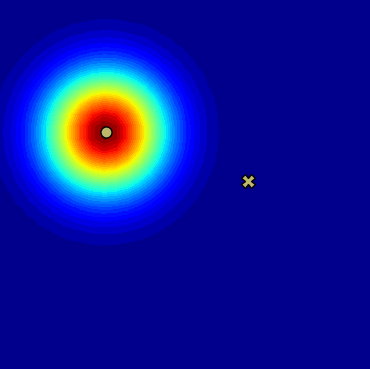}}
    \hspace{0.005cm}
    \subfloat{\includegraphics[width=0.4\linewidth]{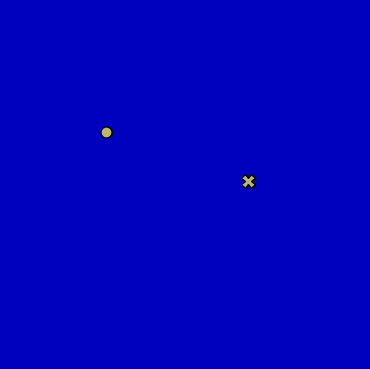}}
    \subfloat{\includegraphics[height=0.4\linewidth]{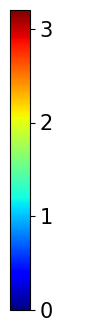}}
    \end{minipage}
    \hspace{-0.5cm}
    \begin{minipage}{0.33\linewidth}
    \centering
    \subfloat{\includegraphics[width=0.4\linewidth]{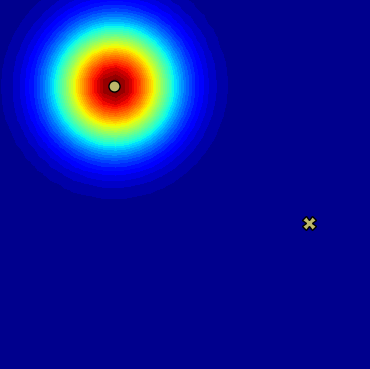}}
    \hspace{0.005cm}
    \subfloat{\includegraphics[width=0.4\linewidth]{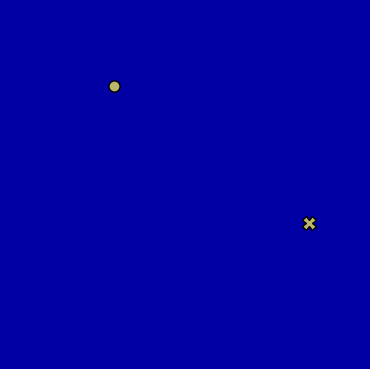}}
    \subfloat{\includegraphics[height=0.4\linewidth]{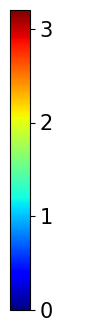}}
    \end{minipage}
    \vspace{-0.3cm}

    \begin{sideways} \makebox[0pt][l]{\hspace{-1.5cm} \shortstack[c]{{\bf \footnotesize \phantom{HyperMARLHyper}} \\ {\footnotesize \phantom{Action}}}} \end{sideways}
    \begin{minipage}{0.33\linewidth}
    \centering
    \hspace{-0.18cm}
    \subfloat{\includegraphics[width=0.4\linewidth]{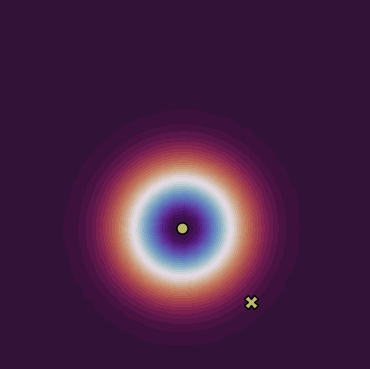}}
    \hspace{0.005cm}
    \subfloat{\includegraphics[width=0.4\linewidth]{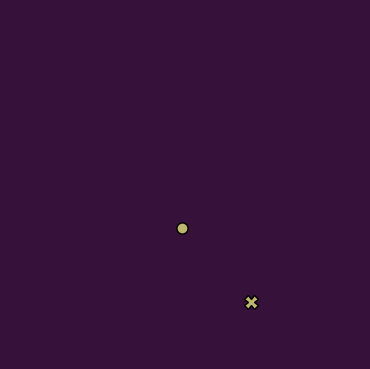}}
    \subfloat{\includegraphics[height=0.4\linewidth]{Figures/FP_Vacuum/Colorbar_Action_LocalMA_td3_seed_2_episode_500.png}}
    \end{minipage}
    \hspace{-0.5cm}
    \begin{minipage}{0.33\linewidth}
    \centering
    \hspace{-0.18cm}
    \subfloat{\includegraphics[width=0.4\linewidth]{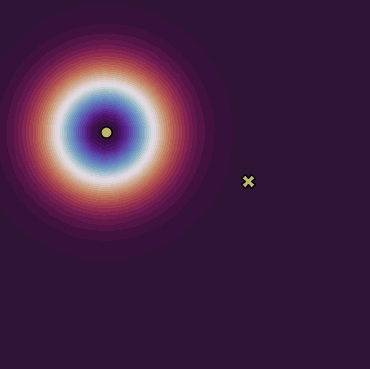}}
    \hspace{0.005cm}
    \subfloat{\includegraphics[width=0.4\linewidth]{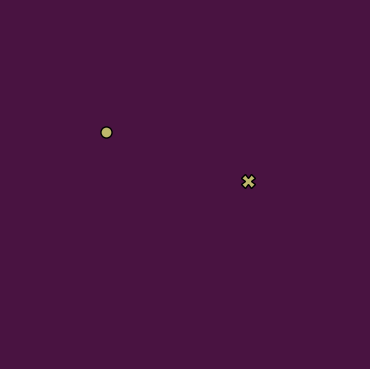}}
    \subfloat{\includegraphics[height=0.4\linewidth]{Figures/FP_Vacuum/Colorbar_Action_LocalMA_td3_seed_3_episode_503.png}}
    \end{minipage}
    \hspace{-0.5cm}
    \begin{minipage}{0.33\linewidth}
    \centering
    \hspace{-0.18cm}
    \subfloat{\includegraphics[width=0.4\linewidth]{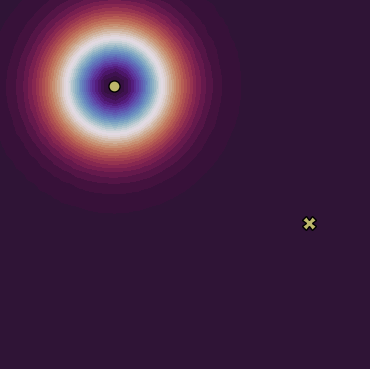}}
    \hspace{0.005cm}
    \subfloat{\includegraphics[width=0.4\linewidth]{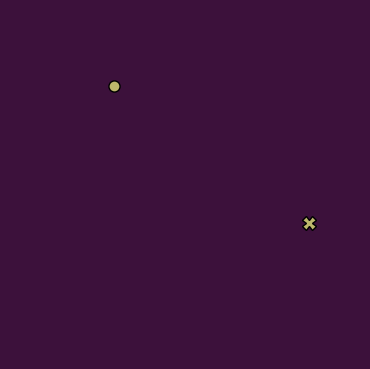}}
    \subfloat{\includegraphics[height=0.4\linewidth]{Figures/FP_Vacuum/Colorbar_Action_LocalMA_td3_seed_5_episode_502.png}}
    \end{minipage}
    
    \caption{{\em Density control in a vacuum}. Uncontrolled states (first row), controlled states and optimal actions obtained with HypeMARL (second and third row), MB-HypeMARL (fourth and fifth row) and MARL (sixth and seventh row) over time in three different random parametric settings.}
    \label{fig:FP_Vacuum2}
\end{figure*}

Figure~\ref{fig:FP_Vacuum2}, instead, displays three examples of controlled solutions at two different time instants, namely $t=0.0\text{s}$ and $t=0.5\text{s}$, obtained after training HypeMARL, MB-HypeMARL, and MARL when dealing with three different random initial positions $\bm{\mu}_0$ and target destinations $\bm{\mu}=\bm{\mu}_T$. Although all algorithms are capable of suppressing the initial density around $\bm{\mu}_0$, only HypeMARL and MB-HypeMARL are able to make it move around $\bm{\mu}_T$. The difficulties of MARL are intrinsic to the fully decentralized nature of the algorithm. Indeed, MARL policies have no way of differentiating states with similar values but located in different positions. As soon as the density is suppressed, all the local states show similar values ($\approx 0$) and, without any knowledge of their own position, the agents are not capable of making the distribution appear in the target location. Thanks to positional encoding, the local agents of HypeMARL are aware of their (relative) position with respect to the others and can effectively solve this challenging task. These results show the importance of developing coordination among agents even in decentralized settings.


\subsubsection{Density control in a fluid}

In this second test case, differently from the setting presented in the previous section, we consider a circular obstacle with radius $0.15$ in the middle of the square $(-1,1)^2$, that is $\Omega = (-1,1)^2 \setminus \mathbb{B}_{0.15}(0,0)$. Moreover, we include a parametric transport term in Equation~\eqref{eq:density_vacuum}, that is, we deal with the following system
\begin{equation}
\label{eq:density_fluid}
\begin{cases}
      \dfrac{\partial y}{\partial t} + \nabla \cdot ( -\kappa \nabla y + \mathbf{v}(\alpha)y) = u \quad
      & \text{in} \ \Omega \times (0,T]
      \\
      ( -\kappa \nabla y + \mathbf{v}(\alpha)y) \cdot \mathbf{n} = 0
      \quad
      & \text{on} \ \partial \Omega  \times (0,T]
      \\
       y(\mathbf{x}, 0) = y_0(\mathbf{x}; \bm{\mu}_0)
       \quad
       & \text{in} \ \Omega
\end{cases}
\end{equation}
where $\kappa = 0.001$ is the diffusion coefficient, $\mathbf{n}$ is the outward normal versor to
the domain boundary, and $u(\mathbf{x},t) \in [-5, 5] \subset \mathbb{R}$ is the distributed control. The advection $\mathbf{v}(\mathbf{x},t;\alpha)$ models an underlying fluid flow representing, for instance, a wind or a water current. The fluid enters the domain from the bottom boundary $\Gamma_{\text{in}} = \partial \Omega \cap \{x_2 = -1\}$, and moves the density upwards according to the steady Navier-Stokes equations
\begin{equation}
\label{eq:steadyNS}
\begin{cases} 
-\nu \Delta\mathbf{v} + (\mathbf{v}\cdot\nabla)\mathbf{v} + \nabla p = \bm{0} \quad & \text{in} \ \Omega \\ 
\nabla \cdot \mathbf{v} = 0 \quad & \text{in} \ \Omega \\ 
\mathbf{v} = \bm{0} \quad & \text{on} \ \Gamma_{\text{obs}} \\
\mathbf{v} = \mathbf{v}_{\text{in}}(\alpha) \quad & \text{on} \ \Gamma_{\text{in}} \\
\mathbf{v}\cdot\mathbf{n} = 0 \quad & \text{on} \ \Gamma_{\text{walls}} \\ 
(\nu \nabla \mathbf{v}-p)\mathbf{n} \cdot \mathbf{t} = 0 \quad & \text{on} \ \Gamma_{\text{walls}} \\
 (\nu \nabla \mathbf{v}-p)\mathbf{n} = \bm{0} \quad & \text{on} \ \Gamma_{\text{out}} , \\
\end{cases}
\end{equation}
where $\nu=0.1$ is the kinematic viscosity, $p(\mathbf{x},t) \in \mathbb{R}$ is the pressure field, $\mathbf{t}$ is the tangential versor to the boundary $\partial \Omega$, $\Gamma_{\text{out}}=\partial \Omega \cap\{x_2=1\}$, $\Gamma_{\text{wall}}=\partial \Omega \cap\{x_1=\pm 1\}$, and $\Gamma_{\text{obs}}$ denotes the obstacle boundary. The inflow Dirichlet boundary datum is defined as
\[
\mathbf{v}_{\text{in}} = \left[(x_1+1)(x_1-1)\sin(\alpha), \cos(\alpha)\right]^{\top}\, , 
\]
where the angle of attack $\alpha \in [-1,1]\text{rad}$ is regarded as an additional system parameter, that is $\bm{\mu}=[\mu_{T,1}, \mu_{T,2}, \alpha]$, while the parabolic profile $(x_1+1)(1-x_1)$ avoids singularities at the domain corners.

To solve the full-order model of Equations~\eqref{eq:density_fluid} and \eqref{eq:steadyNS} through the finite element method, we utilize a spatial discretization of the domain, resulting in state and control dimensions $N_y=N_u=1638$, and a uniform time discretization of the interval $[0, T]$ with $\Delta t= 0.1\text{s}$ and final time $T=1\text{s}$. Similarly to the test case detailed in the previous section, the goal is to move the density from a random initial position $\bm{\mu}_0=[\mu_{0,1}, \mu_{0,2}] \in [-0.75, -0.25]\times[-0.75,0.75]$ towards the parametric final destination $\bm{\mu}_T=[\mu_{T,1}, \mu_{T,2}] \in [0.25, 0.75]\times[-0.75,0.75]$, maximizing the local rewards defined in Equation~\eqref{eq:rewards_densitycontrol}.

\begin{figure*}
    \centering
    \subfloat{\includegraphics[height=0.26\textwidth]{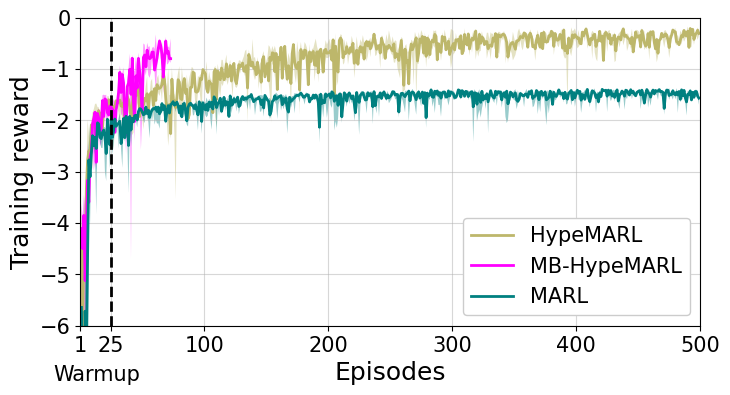}}
    \subfloat{\includegraphics[height=0.26\textwidth]{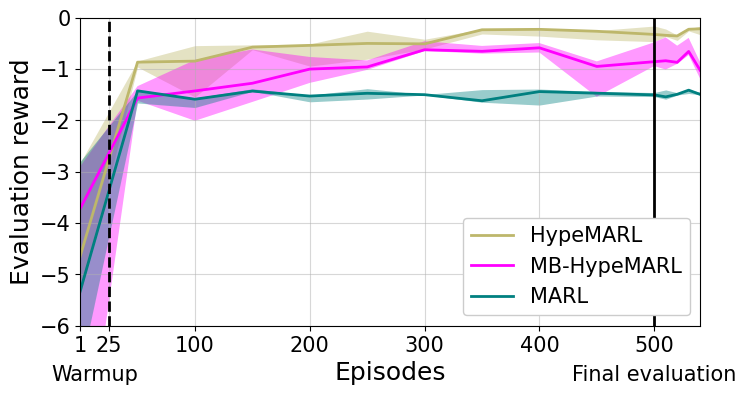}}

    \caption{{\em Density control in a fluid}. Median and interquartile range of HypeMARL, MB-HypeMARL and MARL training end evaluation rewards.}
    \label{fig:FP_Fluid1}
\end{figure*}

\begin{figure*}
    \centering
    \begin{sideways} \makebox[0pt][l]{\hspace{-1.1cm} \shortstack[c]{{\bf \footnotesize No control} \\ {\footnotesize State}}} \end{sideways}
    \begin{minipage}{0.33\linewidth}
    \centering
    \hspace{-0.12cm}
    \subfloat[$t=0.0\text{s}$]{\begin{overpic}[width=0.375\textwidth]{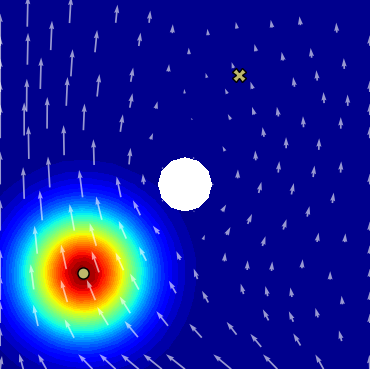}\put(40,130){\shortstack[c]{$\bm{\mu}_0 = (-0.55,-0.48)$ \\ $\bm{\mu}_T = (0.29,0.59)$ \\ $\alpha = -0.91\text{rad}$}}\end{overpic}}
    \subfloat{\includegraphics[height=0.375\linewidth]{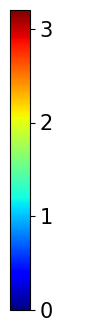}}
    \hspace{-0.24cm}
    \subfloat[$t=0.5\text{s}$]{\includegraphics[width=0.375\linewidth]{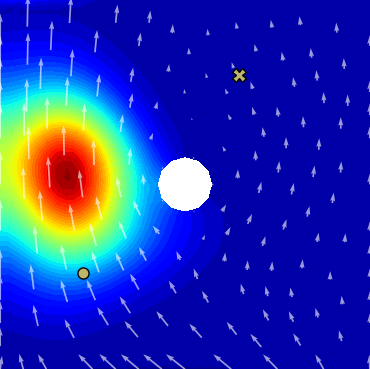}}
    \subfloat{\includegraphics[height=0.375\linewidth]{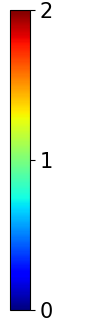}}
    \end{minipage}
    \hspace{-0.5cm}
    \begin{minipage}{0.33\linewidth}
    \centering
    \hspace{-0.2cm}
    \subfloat[$t=0.0\text{s}$]{\begin{overpic}[width=0.375\textwidth]{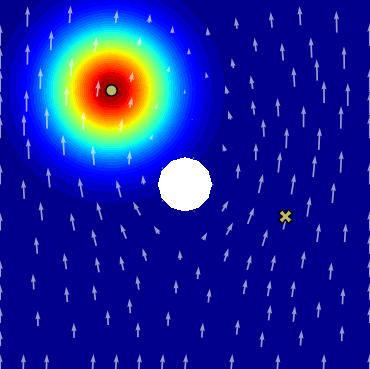}\put(40,130){\shortstack[c]{$\bm{\mu}_0 = (-0.40,0.51)$ \\ $\bm{\mu}_T = (0.54,-0.17)$ \\ $\alpha = 0.08\text{rad}$}}\end{overpic}}
    \subfloat{\includegraphics[height=0.375\linewidth]{Figures/FP_Fluid/Colorbar_State_Uncontrolled.png}}
    \hspace{-0.24cm}
    \subfloat[$t=0.5\text{s}$]{\includegraphics[width=0.375\linewidth]{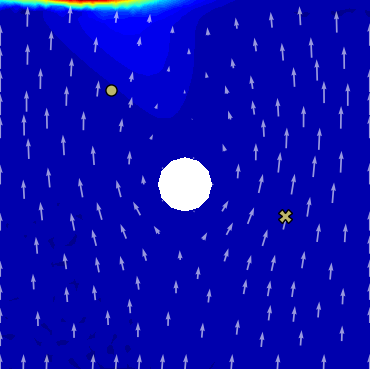}}
    \subfloat{\includegraphics[height=0.375\linewidth]{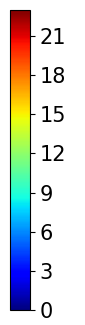}}
    \end{minipage}
    \hspace{-0.5cm}
    \begin{minipage}{0.33\linewidth}
    \centering
    \hspace{-0.2cm}
    \subfloat[$t=0.0\text{s}$]{\begin{overpic}[width=0.375\textwidth]{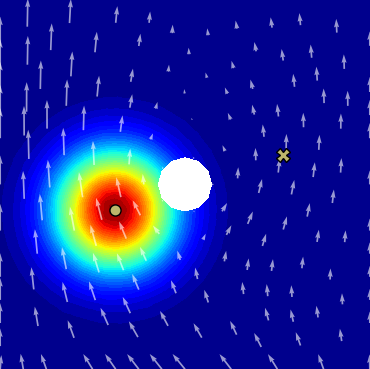}\put(40,130){\shortstack[c]{$\bm{\mu}_0 = (-0.38,-0.14)$ \\ $\bm{\mu}_T = (0.53,0.16)$ \\ $\alpha = -0.69\text{rad}$}}\end{overpic}}
    \subfloat{\includegraphics[height=0.375\linewidth]{Figures/FP_Fluid/Colorbar_State_Uncontrolled.png}}
    \hspace{-0.24cm}
    \subfloat[$t=0.5\text{s}$]{\includegraphics[width=0.375\linewidth]{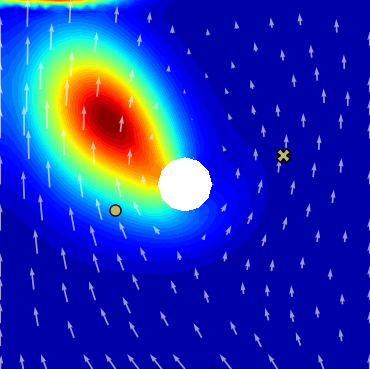}}
    \subfloat{\includegraphics[height=0.375\linewidth]{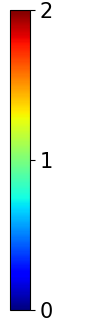}}
    \end{minipage}
    \vspace{-0.3cm}

    \begin{sideways} \makebox[0pt][l]{\hspace{-3cm} \shortstack[c]{{\bf \footnotesize HypeMARL} \\ {\footnotesize Action \hspace{1.3cm} State}}} \end{sideways}
    \begin{minipage}{0.33\linewidth}
    \centering
    \subfloat{\includegraphics[width=0.4\textwidth]{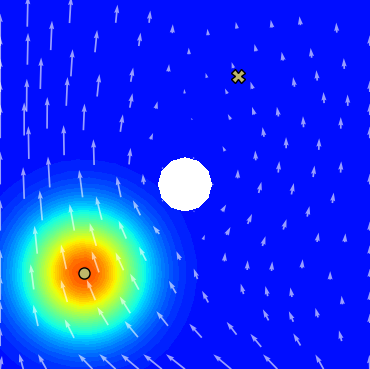}}
    \hspace{0.005cm}
    \subfloat{\includegraphics[width=0.4\linewidth]{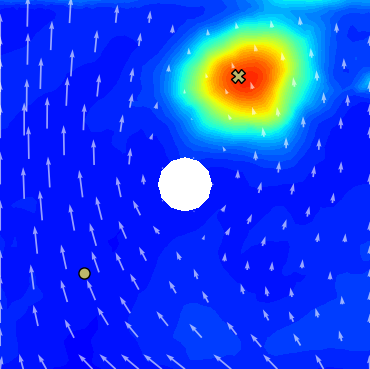}}
    \subfloat{\includegraphics[height=0.4\linewidth]{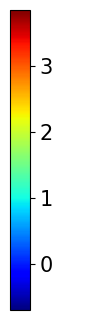}}
    \end{minipage}
    \hspace{-0.5cm}
    \begin{minipage}{0.33\linewidth}
    \centering
    \subfloat{\includegraphics[width=0.4\textwidth]{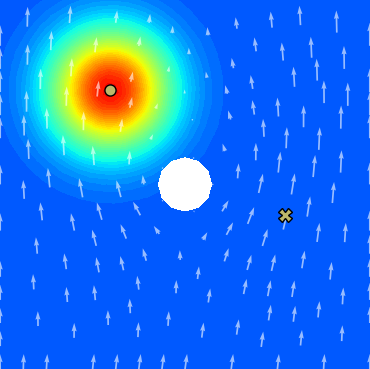}}
    \hspace{0.005cm}
    \subfloat{\includegraphics[width=0.4\linewidth]{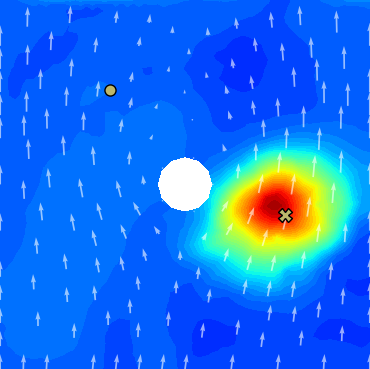}}
    \subfloat{\includegraphics[height=0.4\linewidth]{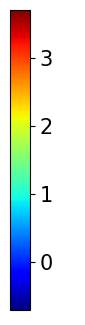}}
    \end{minipage}
    \hspace{-0.5cm}
    \begin{minipage}{0.33\linewidth}
    \centering
    \subfloat{\includegraphics[width=0.4\textwidth]{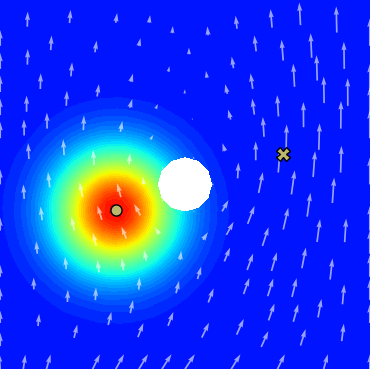}}
    \hspace{0.005cm}
    \subfloat{\includegraphics[width=0.4\linewidth]{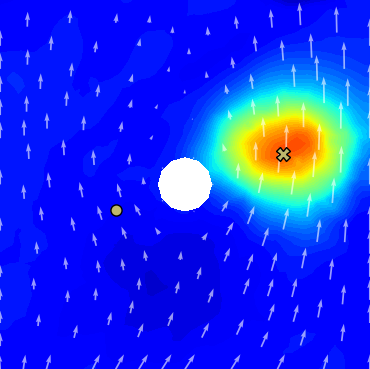}}
    \subfloat{\includegraphics[height=0.4\linewidth]{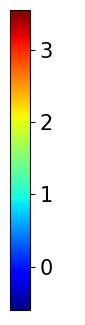}}
    \end{minipage}
    \vspace{-0.3cm}

    \begin{sideways} \makebox[0pt][l]{\hspace{-1cm} \shortstack[c]{{\bf \footnotesize \phantom{HypeMARL}} \\ {\footnotesize \phantom{Action}}}} \end{sideways}
    \begin{minipage}{0.33\linewidth}
    \centering
    \subfloat{\includegraphics[width=0.4\linewidth]{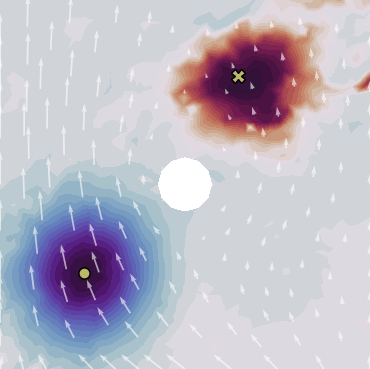}}
    \hspace{0.005cm}
    \subfloat{\includegraphics[width=0.4\linewidth]{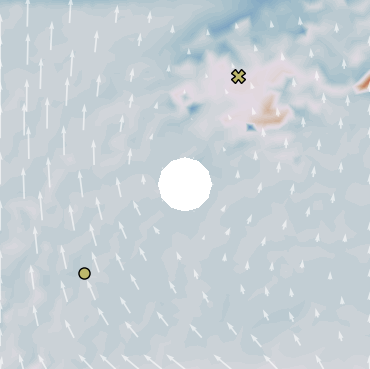}}
    \subfloat{\includegraphics[height=0.4\linewidth]{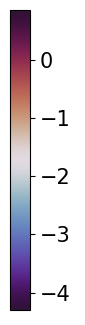}}
    \end{minipage}
    \hspace{-0.5cm}
    \begin{minipage}{0.33\linewidth}
    \centering
    \subfloat{\includegraphics[width=0.4\linewidth]{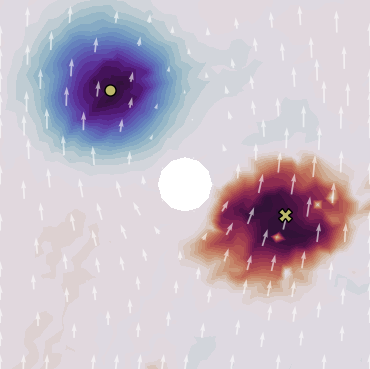}}
    \hspace{0.005cm}
    \subfloat{\includegraphics[width=0.4\linewidth]{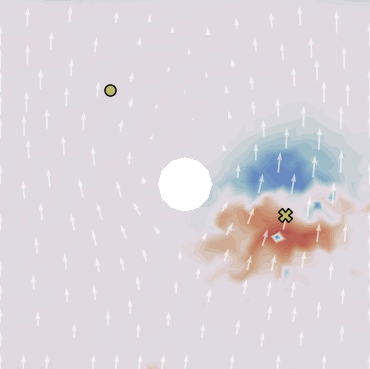}}
    \subfloat{\includegraphics[height=0.4\linewidth]{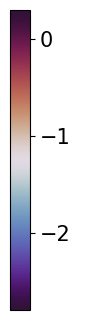}}
    \end{minipage}
    \hspace{-0.5cm}
    \begin{minipage}{0.33\linewidth}
    \centering
    \subfloat{\includegraphics[width=0.4\linewidth]{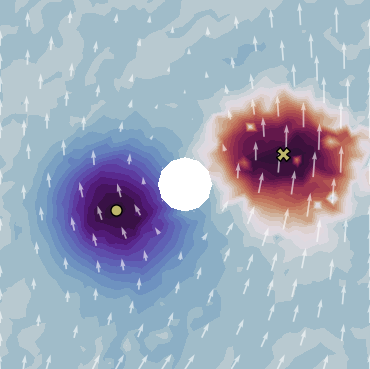}}
    \hspace{0.005cm}
    \subfloat{\includegraphics[width=0.4\linewidth]{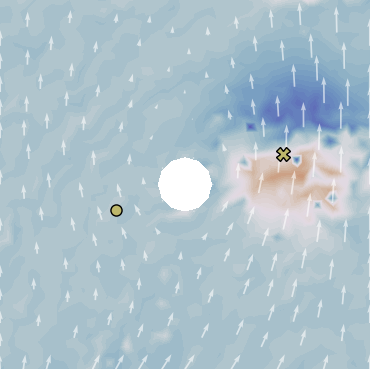}}
    \subfloat{\includegraphics[height=0.4\linewidth]{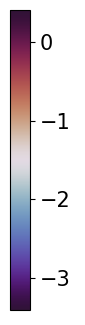}}
    \end{minipage}
    \vspace{-0.3cm}

    \begin{sideways} \makebox[0pt][l]{\hspace{-3cm} \shortstack[c]{{\bf \footnotesize MB-HypeMARL} \\ {\footnotesize Action \hspace{1.5cm} State}}} \end{sideways}
    \begin{minipage}{0.33\linewidth}
    \centering
    \subfloat{\includegraphics[width=0.4\linewidth]{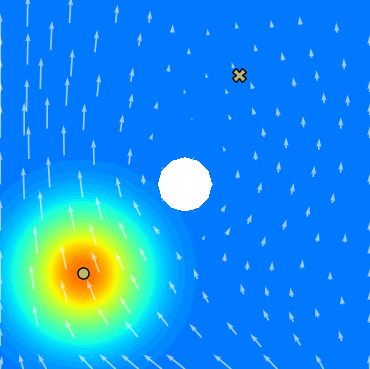}}
    \hspace{0.005cm}
    \subfloat{\includegraphics[width=0.4\linewidth]{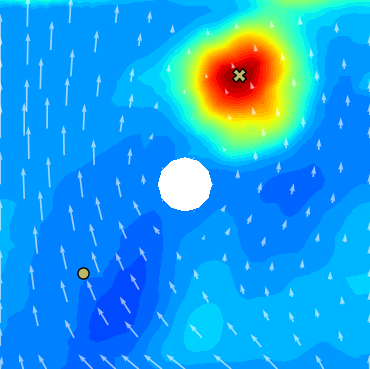}}
    \subfloat{\includegraphics[height=0.4\linewidth]{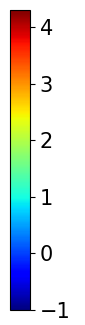}}
    \end{minipage}
    \hspace{-0.5cm}
    \begin{minipage}{0.33\linewidth}
    \centering
    \subfloat{\includegraphics[width=0.4\linewidth]{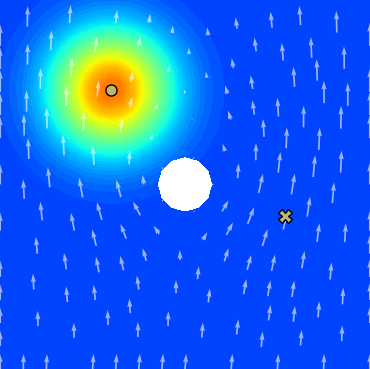}}
    \hspace{0.005cm}
    \subfloat{\includegraphics[width=0.4\linewidth]{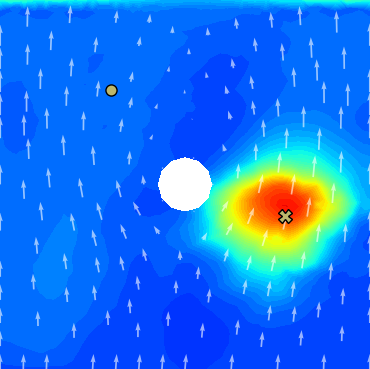}}
    \subfloat{\includegraphics[height=0.4\linewidth]{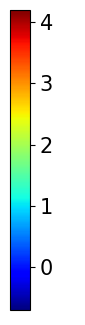}}
    \end{minipage}
    \hspace{-0.5cm}
    \begin{minipage}{0.33\linewidth}
    \centering
    \subfloat{\includegraphics[width=0.4\linewidth]{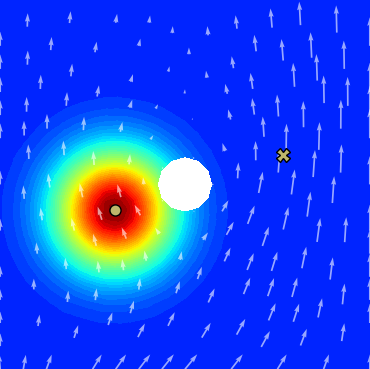}}
    \hspace{0.005cm}
    \subfloat{\includegraphics[width=0.4\linewidth]{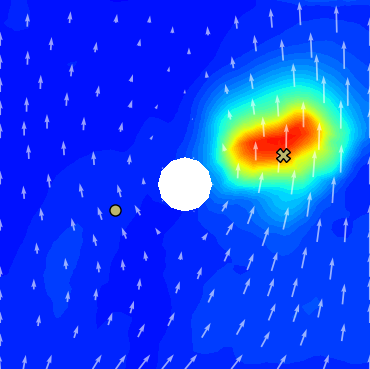}}
    \subfloat{\includegraphics[height=0.4\linewidth]{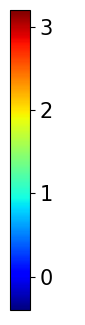}}
    \end{minipage}
    \vspace{-0.3cm}

    \begin{sideways} \makebox[0pt][l]{\hspace{-1.3cm} \shortstack[c]{{\bf \footnotesize \phantom{MB-HypeMARL}} \\ {\footnotesize \phantom{Action}}}} \end{sideways}
    \begin{minipage}{0.33\linewidth}
    \centering
    \subfloat{\includegraphics[width=0.4\linewidth]{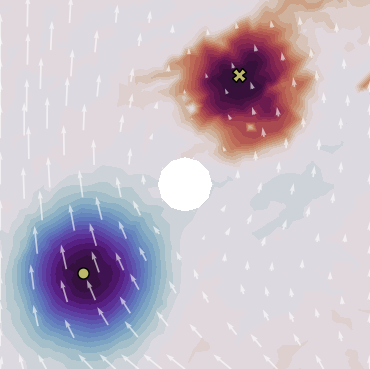}}
    \hspace{0.005cm}
    \subfloat{\includegraphics[width=0.4\linewidth]{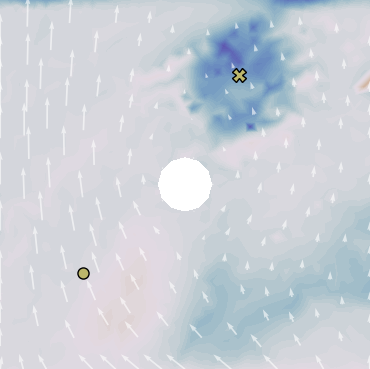}}
    \subfloat{\includegraphics[height=0.4\linewidth]{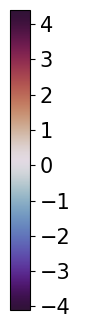}}
    \end{minipage}
    \hspace{-0.5cm}
    \begin{minipage}{0.33\linewidth}
    \centering
    \subfloat{\includegraphics[width=0.4\linewidth]{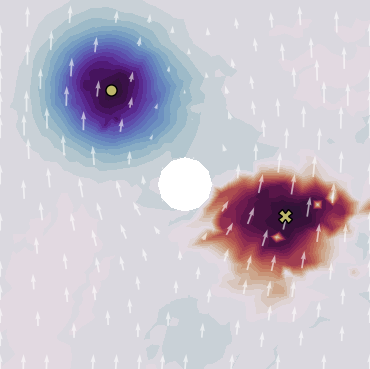}}
    \hspace{0.005cm}
    \subfloat{\includegraphics[width=0.4\linewidth]{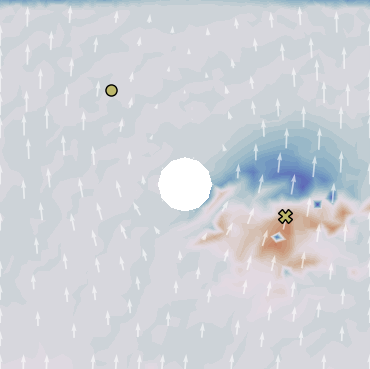}}
    \subfloat{\includegraphics[height=0.4\linewidth]{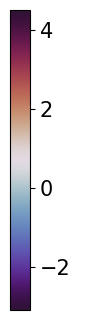}}
    \end{minipage}
    \hspace{-0.5cm}
    \begin{minipage}{0.33\linewidth}
    \centering
    \subfloat{\includegraphics[width=0.4\linewidth]{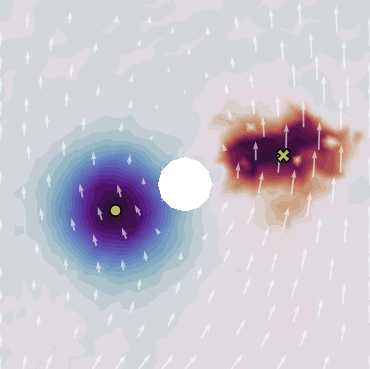}}
    \hspace{0.005cm}
    \subfloat{\includegraphics[width=0.4\linewidth]{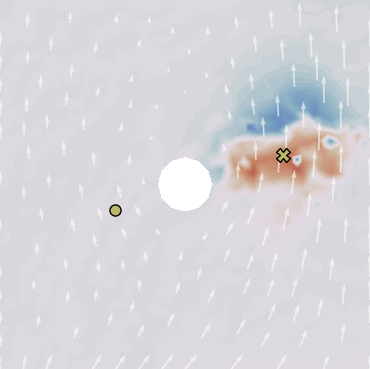}}
    \subfloat{\includegraphics[height=0.4\linewidth]{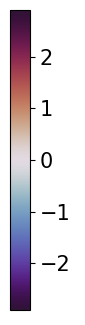}}
    \end{minipage}
    \vspace{-0.3cm}

    \begin{sideways} \makebox[0pt][l]{\hspace{-3cm} \shortstack[c]{{\bf \footnotesize \phantom{Hyper}MARL\phantom{Hyper}} \\ {\footnotesize Action \hspace{1.5cm} State}}} \end{sideways}
    \begin{minipage}{0.33\linewidth}
    \centering
    \subfloat{\includegraphics[width=0.4\linewidth]{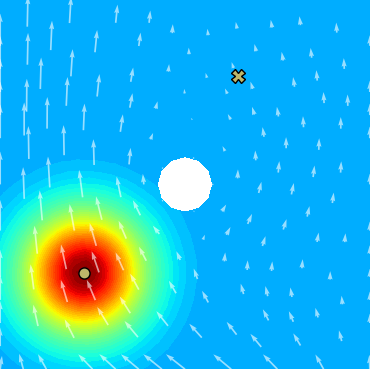}}
    \hspace{0.005cm}
    \subfloat{\includegraphics[width=0.4\linewidth]{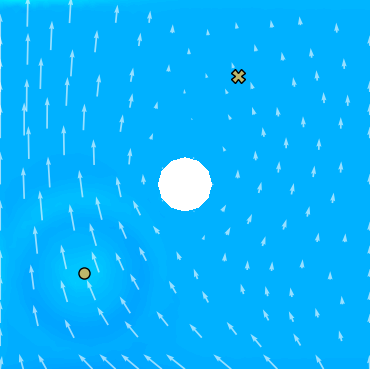}}
    \subfloat{\includegraphics[height=0.4\linewidth]{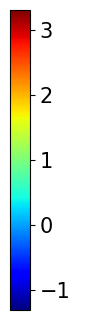}}
    \end{minipage}
    \hspace{-0.5cm}
    \begin{minipage}{0.33\linewidth}
    \centering
    \subfloat{\includegraphics[width=0.4\linewidth]{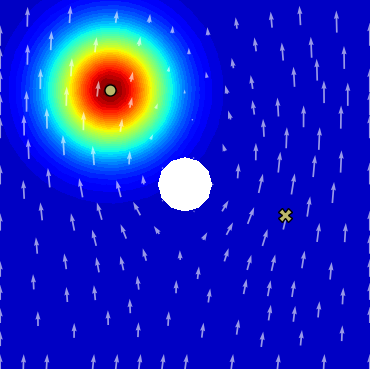}}
    \hspace{0.005cm}
    \subfloat{\includegraphics[width=0.4\linewidth]{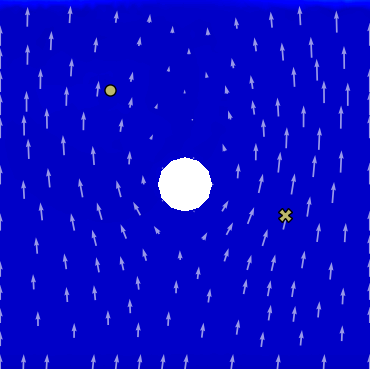}}
    \subfloat{\includegraphics[height=0.4\linewidth]{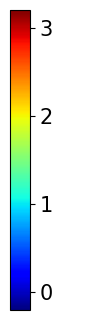}}
    \end{minipage}
    \hspace{-0.5cm}
    \begin{minipage}{0.33\linewidth}
    \centering
    \subfloat{\includegraphics[width=0.4\linewidth]{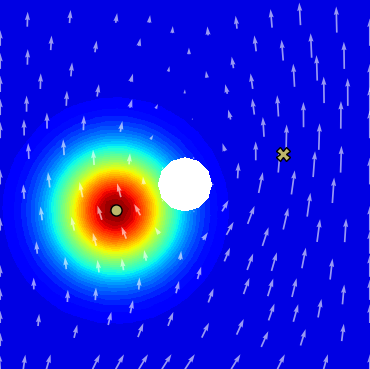}}
    \hspace{0.005cm}
    \subfloat{\includegraphics[width=0.4\linewidth]{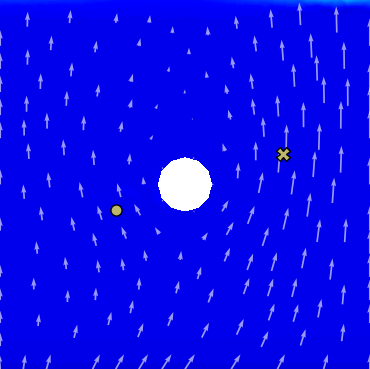}}
    \subfloat{\includegraphics[height=0.4\linewidth]{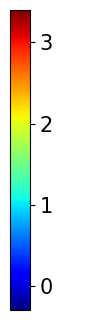}}
    \end{minipage}
    \vspace{-0.3cm}

    \begin{sideways} \makebox[0pt][l]{\hspace{-1.5cm} \shortstack[c]{{\bf \footnotesize \phantom{HyperMARLHyper}} \\ {\footnotesize \phantom{Action}}}} \end{sideways}
    \begin{minipage}{0.33\linewidth}
    \centering
    \hspace{-0.18cm}
    \subfloat{\includegraphics[width=0.4\linewidth]{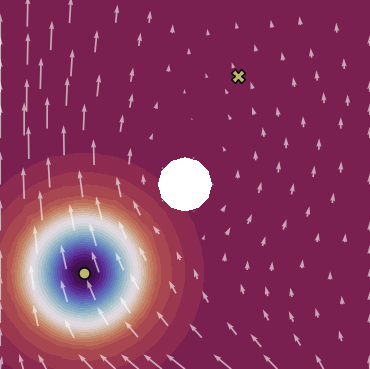}}
    \hspace{0.005cm}
    \subfloat{\includegraphics[width=0.4\linewidth]{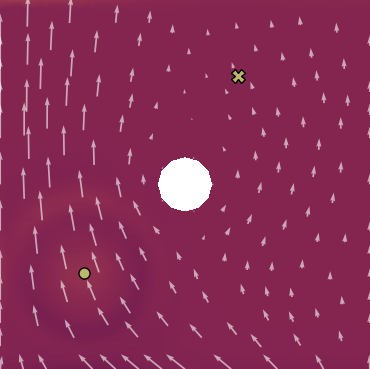}}
    \subfloat{\includegraphics[height=0.4\linewidth]{Figures/FP_Fluid/Colorbar_Action_LocalMA_td3_seed_1_episode_500.png}}
    \end{minipage}
    \hspace{-0.5cm}
    \begin{minipage}{0.33\linewidth}
    \centering
    \hspace{-0.18cm}
    \subfloat{\includegraphics[width=0.4\linewidth]{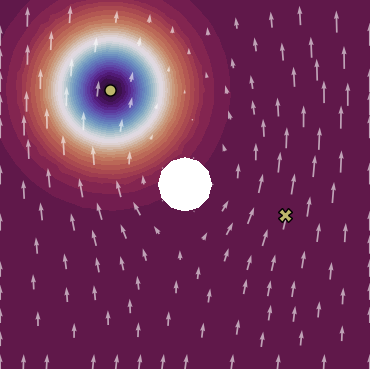}}
    \hspace{0.005cm}
    \subfloat{\includegraphics[width=0.4\linewidth]{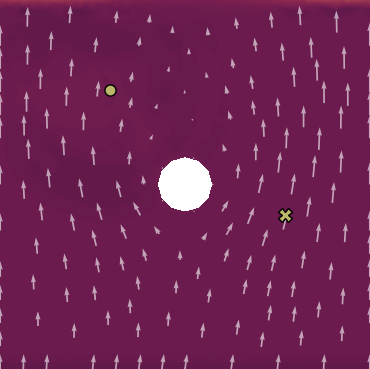}}
    \subfloat{\includegraphics[height=0.4\linewidth]{Figures/FP_Fluid/Colorbar_Action_LocalMA_td3_seed_5_episode_504.png}}
    \end{minipage}
    \hspace{-0.5cm}
    \begin{minipage}{0.33\linewidth}
    \centering
    \hspace{-0.18cm}
    \subfloat{\includegraphics[width=0.4\linewidth]{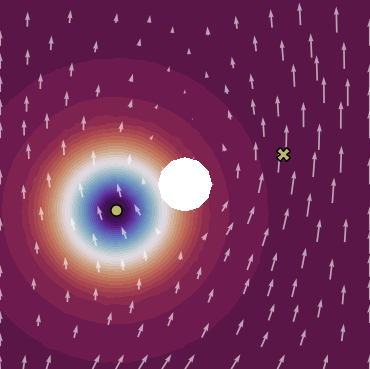}}
    \hspace{0.005cm}
    \subfloat{\includegraphics[width=0.4\linewidth]{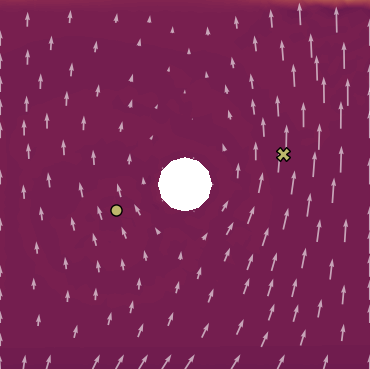}}
    \subfloat{\includegraphics[height=0.4\linewidth]{Figures/FP_Fluid/Colorbar_Action_LocalMA_td3_seed_4_episode_500.png}}
    \end{minipage}

    \caption{{\em Density control in a fluid}. Uncontrolled states (first row), controlled states and optimal actions obtained with HypeMARL (second and third row), MB-HypeMARL (fourth and fifth row) and MARL (sixth and seventh row) over time in three different random parametric settings.}
    \label{fig:FP_Fluid2}
\end{figure*}

Figure~\ref{fig:FP_Fluid1} shows the median and interquartile range of training and evaluation rewards obtained with $5$ independent runs with $5$ different seeds. Similarly to the previous test case, HypeMARL and MB-HypeMARL outperform MARL with remarkably higher rewards. Indeed, as visible in Figure~\ref{fig:FP_Fluid2} showing three examples of controlled solutions over time, all control methods are able to suppress the initial density. However, only HypeMARL and MB-HypeMARL are capable of controlling the source term to make the density appear at the final desired position. Notice that, due to the presence of the underlying fluid, after moving the density to the target location, the agents have to maintain the density in place, counterbalancing the parametric transport effect. Thanks to an accurate deep learning-based surrogate model approximating the local dynamics, which represents a very limited additional complexity to the algorithm, it is possible to limit the real environment interactions with the model-based algorithm MB-HypeMARL, thus resulting in a faster training procedure, with minimal deterioration of the control performance.

\subsection{Flow control}
\label{subsec:flowcontrol}

The second class of applications considered deals with the control of a flow around an object exploiting different actuation strategies. In particular, the first setting considers vertical jets, which blow and inject fluid on the top and bottom sides of the object to maximize (minimize) the pressure below (above) the obstacle. In the second scenario, the controller actively changes the obstacle's shape to minimize the pressure at the front of the obstacle.

\subsubsection{Flow control with vertical jets}

In this section we test the MARL algorithms to steer an incompressible fluid flow around an object. We consider a rectangular channel $\Omega = (0,10) \times (0,2) \setminus \Omega_{\text{obs}}$ with an inner obstacle consisting of the union of a rectangle and two semicircles, that is,  $\Omega_{\text{obs}} = \Omega_{\text{l}} \cup \Omega_{\text{c}} \cup \Omega_{\text{r}}$ where $\Omega_{\text{l}} = \mathbb{B}_{0.25}(0.75,1.00)$, $\Omega_{\text{c}} = [0.75, 1.25] \times[0.75, 1.25]$ and $\Omega_{\text{r}} = \mathbb{B}_{0.25}(1.25,1.00)$. The fluid dynamics in terms of velocity $\mathbf{v}(\mathbf{x},t) \in \mathbb{R}^2$ and pressure $p(\mathbf{x},t) \in \mathbb{R}$ is described by the unsteady Navier-Stokes equations 
\begin{equation}
\label{eq:unsteadyNS}
\begin{cases} 
\dfrac{\partial\mathbf{v}}{\partial t}-\nu \Delta\mathbf{v} + (\mathbf{v}\cdot\nabla)\mathbf{v} + \nabla p = \bm{0} \quad & \text{in} \ \Omega \times (0,T] \\ 
\nabla \cdot \mathbf{v} = 0 \quad & \text{in} \ \Omega \times (0,T] \\ 
\mathbf{v} = \bm{0} \quad & \text{on} \ \Gamma_{\text{l}} \cup\Gamma_{\text{r}} \\
\mathbf{v} = u\mathbf{n} \quad & \text{on} \ \Gamma_{\text{c}} \\
\mathbf{v} = \mathbf{v}_{\text{in}}(\alpha) \quad & \text{on} \ \Gamma_{\text{in}} \\
\mathbf{v}\cdot\mathbf{n} = 0 \quad & \text{on} \ \Gamma_{\text{walls}} \\ 
(\nu \nabla \mathbf{v}-p)\mathbf{n} \cdot \mathbf{t} = 0 \quad & \text{on} \ \Gamma_{\text{walls}} \\
 (\nu \nabla \mathbf{v}-p)\mathbf{n} = \bm{0} \quad & \text{on} \ \Gamma_{\text{out}} \\
 \mathbf{v}(\mathbf{x},0) = \mathbf{0} \quad & \text{in} \ \Omega\, ,
\end{cases}
\end{equation}
where $\nu = 0.01$ is the kinematic viscosity, $\mathbf{n}$ and $\mathbf{t}$ are the normal and tangential versors to the domain boundary, $\Gamma_{\text{in}} = \partial \Omega \cap \{x_1 = 0\}$ is the inflow boundary, $\Gamma_{\text{out}} = \partial \Omega \cap \{ x_1 = 10\}$ is the outflow boundary, $\Gamma_{\text{wall}} = \partial \Omega \cap (\{ x_2 = 0\} \cup \{ x_2 = 2\})$ is the wall, $\Gamma_{\text{c}} = \partial \Omega_{\text{obs}} \cap \{0.75 < x_1 < 1.25 \}$ is the control boundary, while $\Gamma_{\text{l}} = \partial \Omega_{\text{obs}} \cap \{ x_1 \leq 0.75 \}$ and $\Gamma_{\text{r}} = \partial \Omega_{\text{obs}} \cap \{ x_1 \geq 1.25 \}$ are the rounded obstacle boundaries. To steer the fluid flow, we consider vertical jets on $\Gamma_{\text{c}}$ with intensities $u(\mathbf{x},t) \in [-1,1] \subset \mathbb{R}$. Moreover, we take into account an inflow velocity defined as
\[
\mathbf{v}_{\text{in}}(\mathbf{x},t;\alpha) = [10x_2(2 - x_2  )\cos(\alpha), 10\sin(\alpha)]^{\top} ,
\]
where the angle of attack $\alpha \in [-1,1] \text{rad}$ is regarded as a system parameter, i.e. $\mu = \alpha$. The Reynolds number for this test case is equal to $1000$. To solve Equation~\eqref{eq:unsteadyNS} with the finite element method, we discretize the spatial domain, resulting in a velocity and pressure dimension equal to, respectively, $N_{\mathbf{v}}= 100600$ and $N_p = 12712$. Moreover, we consider a uniform time grid in the interval $[0,T]$ with time step $\Delta t = 0.1\text{s}$ and final time $T=5.5\text{s}$. 

We now consider MARL, HypeMARL and MB-HypeMARL to control this challenging parametric and high-dimensional problem. Differently from the test cases detailed in Section~\ref{subsec:densitycontrol}, where the state and the actions are observed and predicted over the whole domain, the vertical jets are defined on the obstacle boundary $\Gamma_{c}$. Therefore, starting from the $N_y = 74$ local pressure values $y_{i,t} = p_{i,t}=p(\mathbf{x}_i,t)$, with $\mathbf{x}_i = (x_{i,1}, x_{i,2}) \in \Gamma_{c}$, MARL algorithms have to predict the optimal intensities of the corresponding $N_u = 74$ jets $u_{i,t}$ in the same locations in order to maximize the local rewards
\[
r_{i,t} = r(p_{i,t}, u_{i,t}) = 
\begin{cases}
p_{i,t} \; & \text{if} \ x_{i,2} = 0.75
\\
- p_{i,t} \; & \text{if} \ x_{i,2} = 1.25
\end{cases}
\ .
\]
Specifically, we aim at maximizing the pressure values at the bottom boundary of the obstacle, while minimizing the local pressure on top. Notice that the boundary control is applied after letting the uncontrolled system evolve for $0.5 \text{s}$, and it is updated by the control methods every $0.5 \text{s}$. 

\begin{figure*}
    
    \begin{minipage}{0.49\linewidth}
    \centering \subfloat{\includegraphics[height=0.5\textwidth]{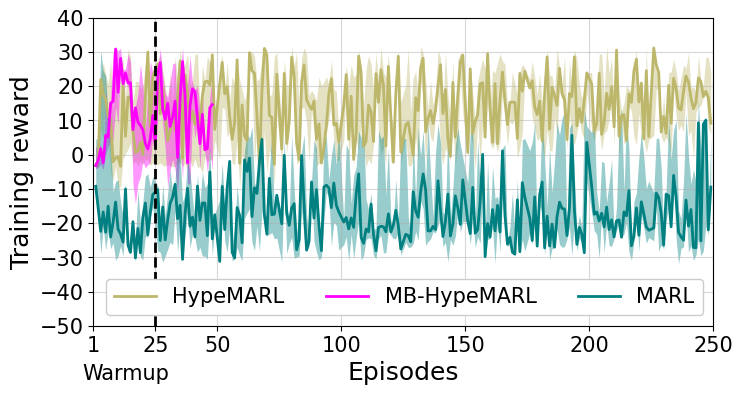}}
    \vspace{-0.5cm}
    
    \subfloat{\includegraphics[height=0.5\textwidth]{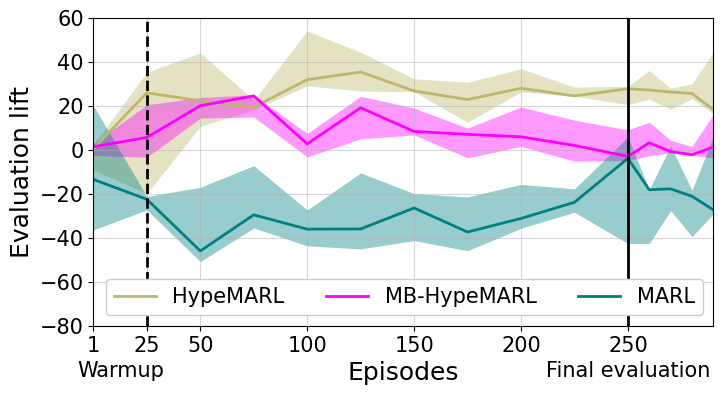}}
    \end{minipage}
    \begin{minipage}{0.49\linewidth}
    \centering \subfloat{\includegraphics[height=0.5\textwidth]{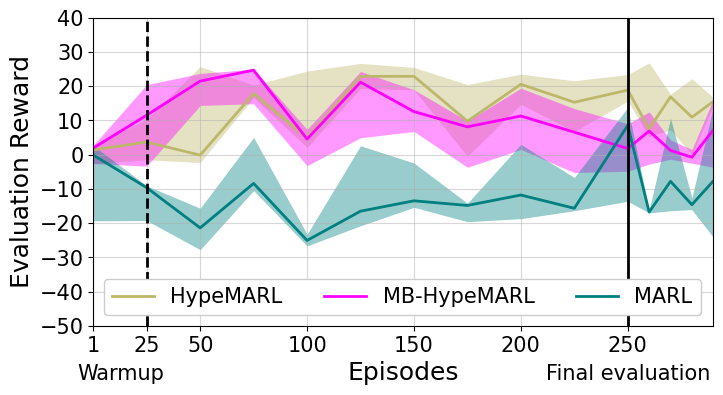}}
    \vspace{-0.5cm}
    
    \subfloat{\includegraphics[height=0.5\textwidth]{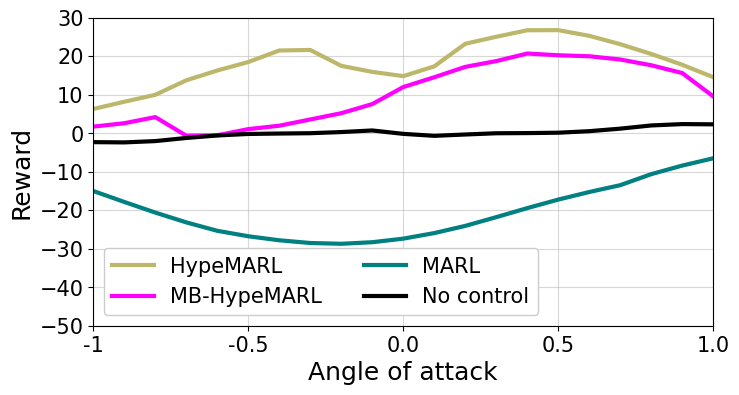}}
    \end{minipage}
    
    \caption{{\em Flow control with vertical jets}. First row: median and interquartile range of HypeMARL, MB-HypeMARL and MARL training end evaluation rewards. Second row: median and interquartile range of HypeMARL, MB-HypeMARL and MARL evaluation lifts (left) and reward obtained with HypeMARL, MB-HypeMARL, MARL and no control for different angles of attack (right).}
    \label{fig:NS_Jets1}
\end{figure*}

The first row of Figure~\ref{fig:NS_Jets1} reports the average local rewards in training and evaluation of the different competing algorithms with respect to the number of training episodes. The results show the median and interquartile range of $5$ independent runs with $5$ different seeds for a fair and robust comparison. We train the algorithms for a total of $250$ episodes, exploring the action space with additive Gaussian random noise, which linearly decays after the first $25$ episodes of warm-up, on the actions predicted by the policy. To highlight the data efficiency of MB-HypeMARL, we only report the training rewards collected when interacting with the environment, while we leave out the interactions with the surrogate one. After the first $25$ warm-up interactions with the real environment, MB-HypeMARL performs $10$ episodes of interaction with the surrogate environment for every episode of interaction with the real one. The agents are evaluated once every $25$ training episodes. HypeMARL and MB-HypeMARL outperform classical decentralized MARL with remarkably higher rewards. Moreover, the model-based version is extremely data efficient and is capable of achieving effective control strategies with only $48$ real environment interactions. Besides the evaluation rewards, the left panel in the second row of Figure~\ref{fig:NS_Jets1} provides the evaluation lift of the different control strategies. It is important to note that, due to the high correlation of the selected reward with the lift force, HypeMARL and MB-HypeMARL provide higher lift values compared to the classical decentralized MARL. Moreover, the right panel in the second row of Figure~\ref{fig:NS_Jets1} compares the reward of the different competing methods across different angles of attack in the range $[-1,1]\text{rad}$ with respect to the uncontrolled setting. In particular, it is possible to further assess the effectiveness of the proposed algorithms in all the parametric scenarios considered and the unsatisfactory results of classical decentralized MARL.

\begin{figure*}
    
    \begin{sideways} \makebox[0pt][l]{\hspace{-0.7cm} {\bf \footnotesize No control}} \end{sideways}
    \begin{minipage}{0.33\linewidth}
    \hspace{-0.17cm}
    \subfloat{\begin{overpic}[height=0.32\linewidth]{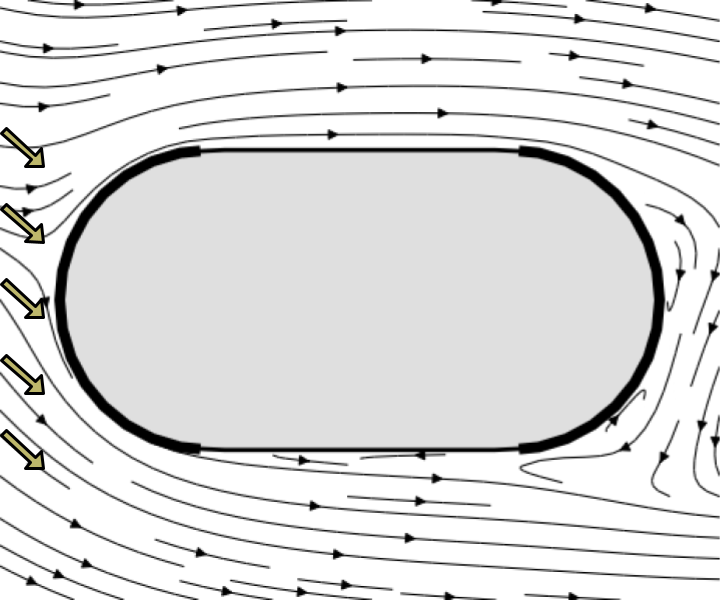}\put(65,90){\shortstack[c]{$\alpha = -0.95\text{rad}$ \\ $t=3.0\text{s}$}}\end{overpic}}
    \hspace{0.005cm}
    \subfloat{\includegraphics[height=0.32\linewidth]{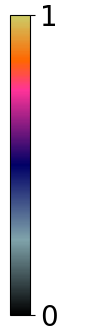}}
    \subfloat{\includegraphics[height=0.32\linewidth]{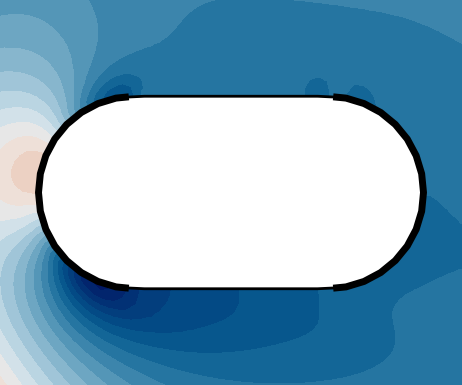}}
    \subfloat{\includegraphics[height=0.32\linewidth]{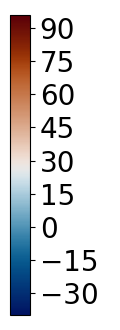}}
    \vspace{-0.3cm}
    
    \subfloat{\includegraphics[width=0.9\linewidth]{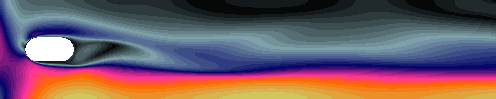}}
    \vspace{-0.4cm}

    \subfloat{\includegraphics[width=0.9\linewidth]{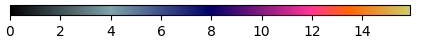}}
    \end{minipage}
    \hspace{-0.25cm}
    \begin{minipage}{0.33\linewidth}
    \hspace{-0.17cm}
    \subfloat{\begin{overpic}[height=0.32\linewidth]{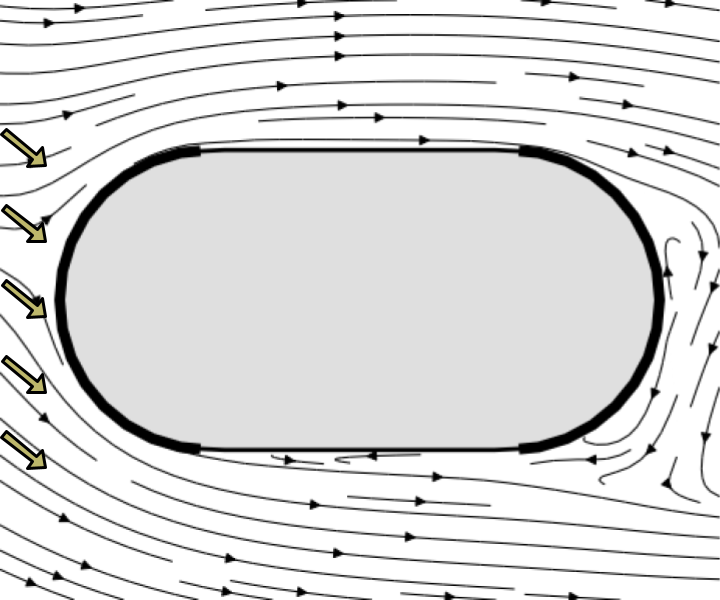}\put(65,90){\shortstack[c]{$\alpha = -0.81\text{rad}$ \\ $t=4.0\text{s}$}}\end{overpic}}
    \hspace{0.005cm}
    \subfloat{\includegraphics[height=0.32\linewidth]{Figures/NS_Jets/Colorbar_Action.png}}
    \subfloat{\includegraphics[height=0.32\linewidth]{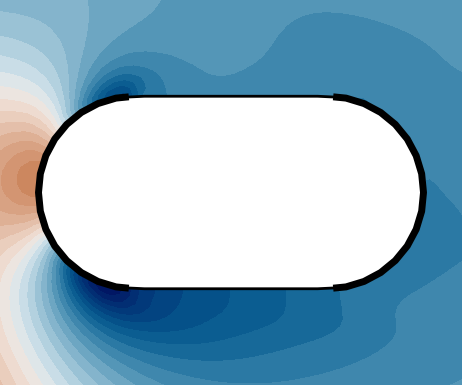}}
    \subfloat{\includegraphics[height=0.32\linewidth]{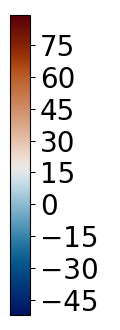}}
    \vspace{-0.3cm}
    
    \subfloat{\includegraphics[width=0.9\linewidth]{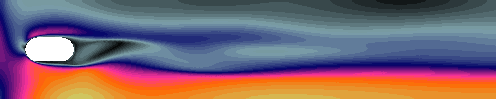}}
    \vspace{-0.4cm}

    \subfloat{\includegraphics[width=0.9\linewidth]{Figures/NS_Jets/Colorbar_Velocity_Uncontrolled.png}}
    \end{minipage}
    \hspace{-0.25cm}
    \begin{minipage}{0.33\linewidth}
    \hspace{-0.17cm}
    \subfloat{\begin{overpic}[height=0.32\linewidth]{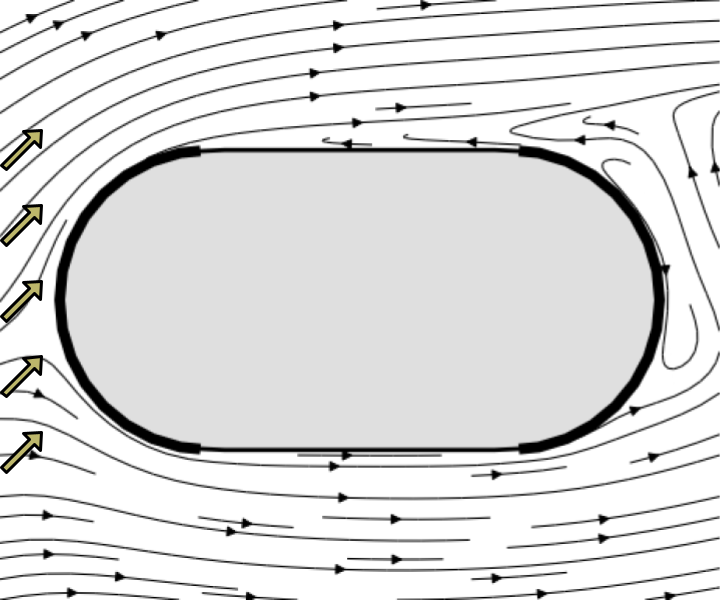}\put(65,90){\shortstack[c]{$\alpha = 0.90\text{rad}$ \\ $t=5.0\text{s}$}}\end{overpic}}
    \hspace{0.005cm}
    \subfloat{\includegraphics[height=0.32\linewidth]{Figures/NS_Jets/Colorbar_Action.png}}
    \subfloat{\includegraphics[height=0.32\linewidth]{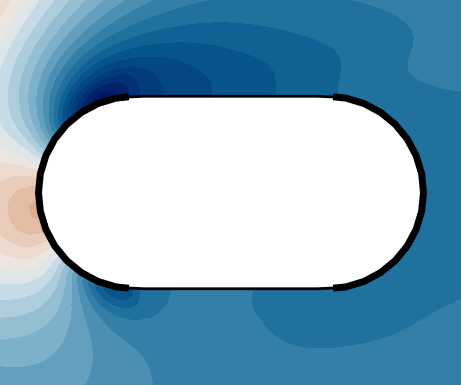}}
    \subfloat{\includegraphics[height=0.32\linewidth]{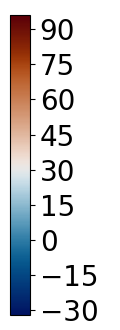}}
    \vspace{-0.3cm}
    
    \subfloat{\includegraphics[width=0.9\linewidth]{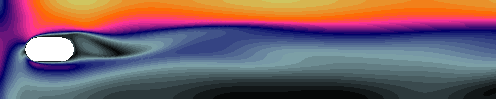}}
    \vspace{-0.4cm}

    \subfloat{\includegraphics[width=0.9\linewidth]{Figures/NS_Jets/Colorbar_Velocity_Uncontrolled.png}}
    \end{minipage}
    \vspace{-0.3cm}

    \begin{sideways} \makebox[0pt][l]{\hspace{-0.7cm} {\bf \footnotesize HypeMARL}} \end{sideways}
    \begin{minipage}{0.33\linewidth}
    \subfloat{\includegraphics[height=0.32\linewidth]{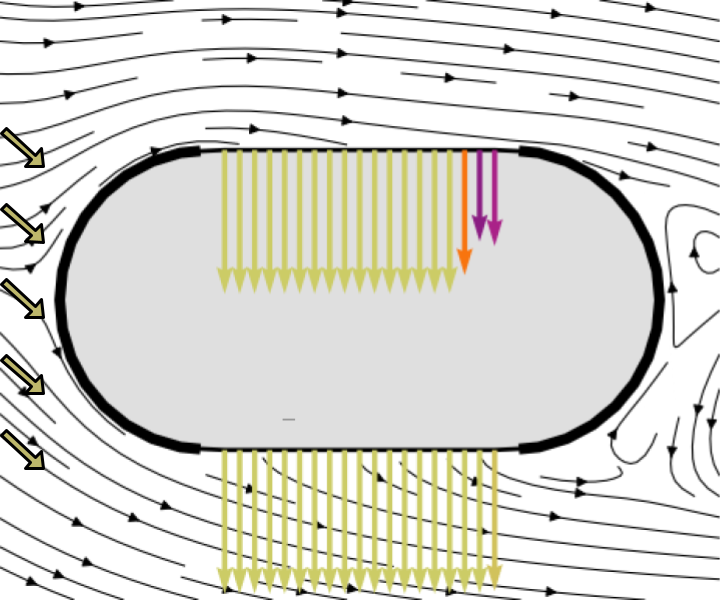}}
    \hspace{0.005cm}
    \subfloat{\includegraphics[height=0.32\linewidth]{Figures/NS_Jets/Colorbar_Action.png}}
    \subfloat{\includegraphics[height=0.32\linewidth]{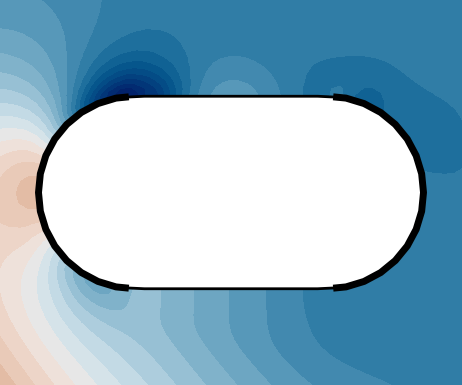}}
    \subfloat{\includegraphics[height=0.32\linewidth]{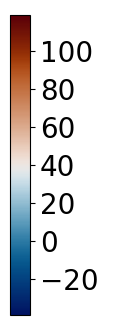}}
    \vspace{-0.3cm}
    
    \subfloat{\includegraphics[width=0.9\linewidth]{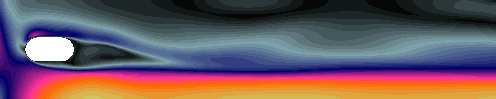}}
    \vspace{-0.4cm}

    \subfloat{\includegraphics[width=0.9\linewidth]{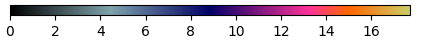}}
    \end{minipage}
    \hspace{-0.25cm}
    \begin{minipage}{0.33\linewidth}
    \subfloat{\includegraphics[height=0.32\linewidth]{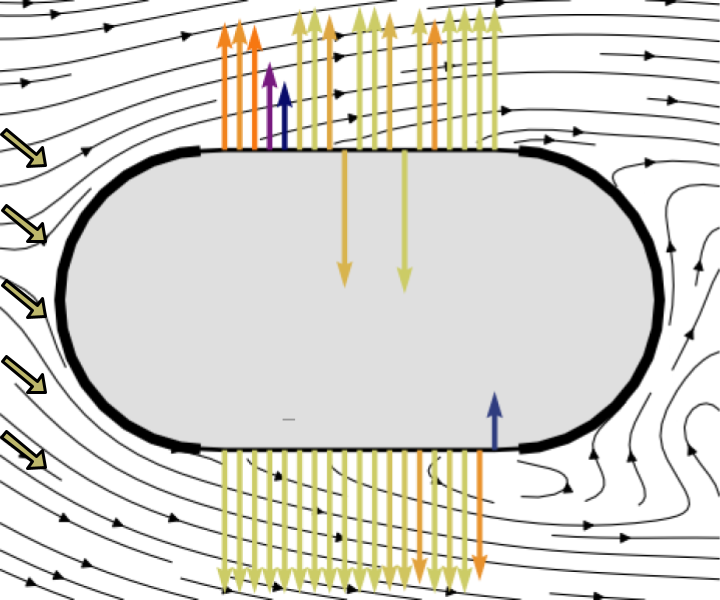}}
    \hspace{0.005cm}
    \subfloat{\includegraphics[height=0.32\linewidth]{Figures/NS_Jets/Colorbar_Action.png}}
    \subfloat{\includegraphics[height=0.32\linewidth]{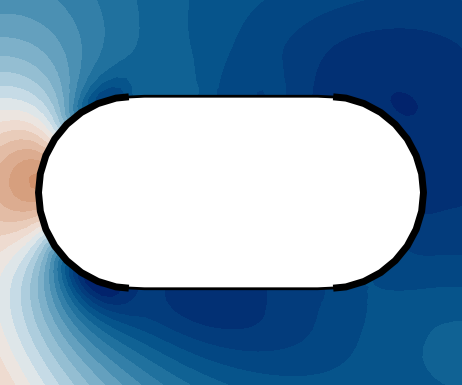}}
    \subfloat{\includegraphics[height=0.32\linewidth]{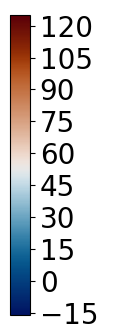}}
    \vspace{-0.3cm}
    
    \subfloat{\includegraphics[width=0.9\linewidth]{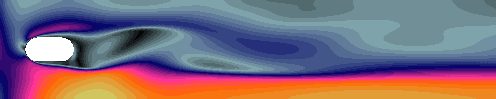}}
    \vspace{-0.4cm}

    \subfloat{\includegraphics[width=0.9\linewidth]{Figures/NS_Jets/Colorbar_Velocity.png}}
    \end{minipage}
    \hspace{-0.25cm}
    \begin{minipage}{0.33\linewidth}
    \subfloat{\includegraphics[height=0.32\linewidth]{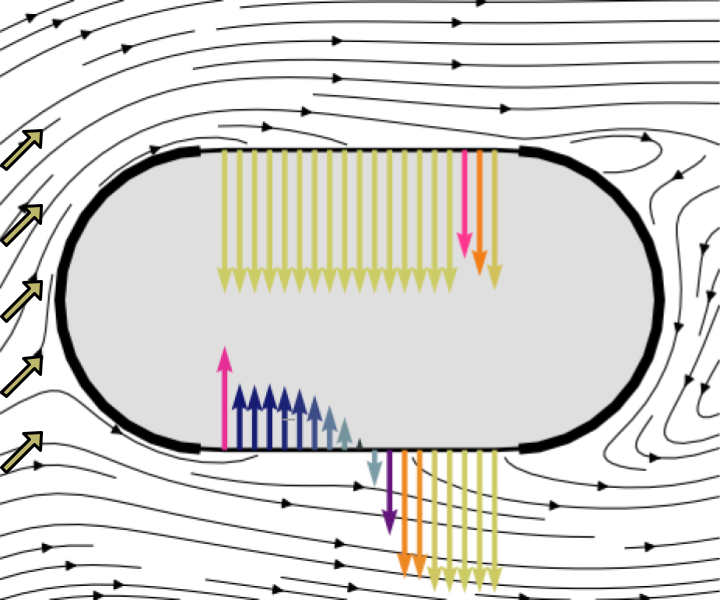}}
    \hspace{0.005cm}
    \subfloat{\includegraphics[height=0.32\linewidth]{Figures/NS_Jets/Colorbar_Action.png}}
    \subfloat{\includegraphics[height=0.32\linewidth]{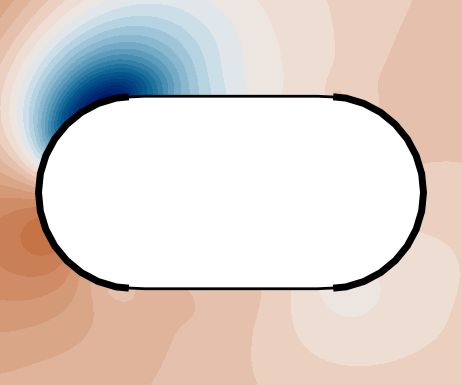}}
    \subfloat{\includegraphics[height=0.32\linewidth]{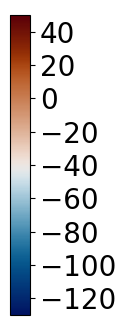}}
    \vspace{-0.3cm}
    
    \subfloat{\includegraphics[width=0.9\linewidth]{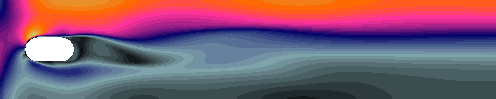}}
    \vspace{-0.4cm}

    \subfloat{\includegraphics[width=0.9\linewidth]{Figures/NS_Jets/Colorbar_Velocity.png}}
    \end{minipage}
    \vspace{-0.3cm}

    \begin{sideways} \makebox[0pt][l]{\hspace{-1.1cm} {\bf \footnotesize MB-HypeMARL}} \end{sideways}
    \begin{minipage}{0.33\linewidth}
    \subfloat{\includegraphics[height=0.32\linewidth]{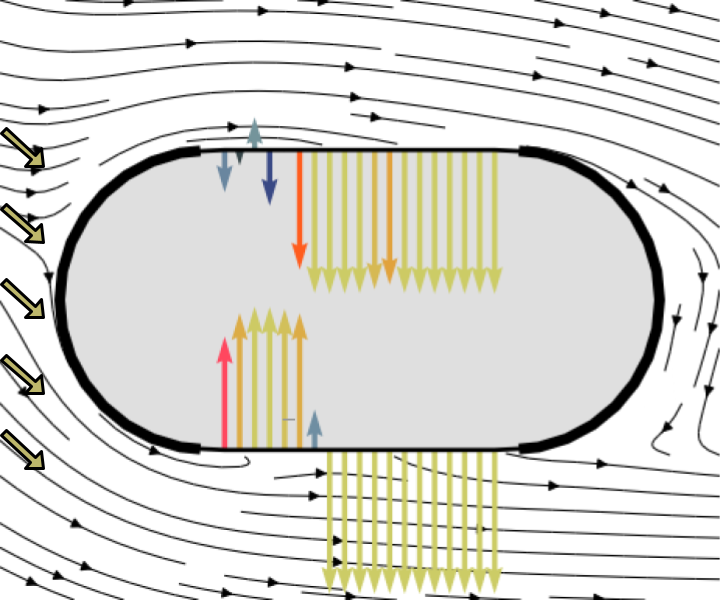}}
    \hspace{0.005cm}
    \subfloat{\includegraphics[height=0.32\linewidth]{Figures/NS_Jets/Colorbar_Action.png}}
    \subfloat{\includegraphics[height=0.32\linewidth]{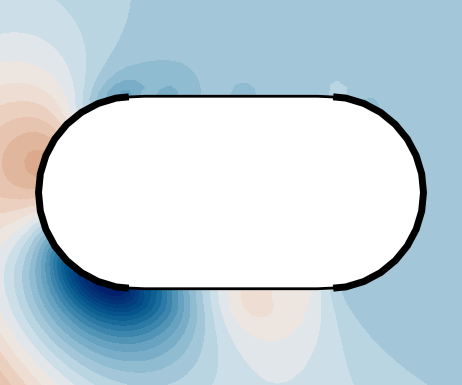}}
    \subfloat{\includegraphics[height=0.32\linewidth]{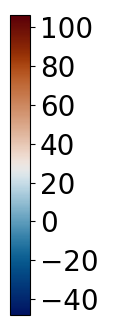}}
    \vspace{-0.3cm}
    
    \subfloat{\includegraphics[width=0.9\linewidth]{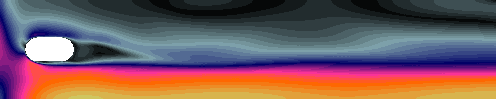}}
    \vspace{-0.4cm}

    \subfloat{\includegraphics[width=0.9\linewidth]{Figures/NS_Jets/Colorbar_Velocity.png}}
    \end{minipage}
    \hspace{-0.25cm}
    \begin{minipage}{0.33\linewidth}
    \subfloat{\includegraphics[height=0.32\linewidth]{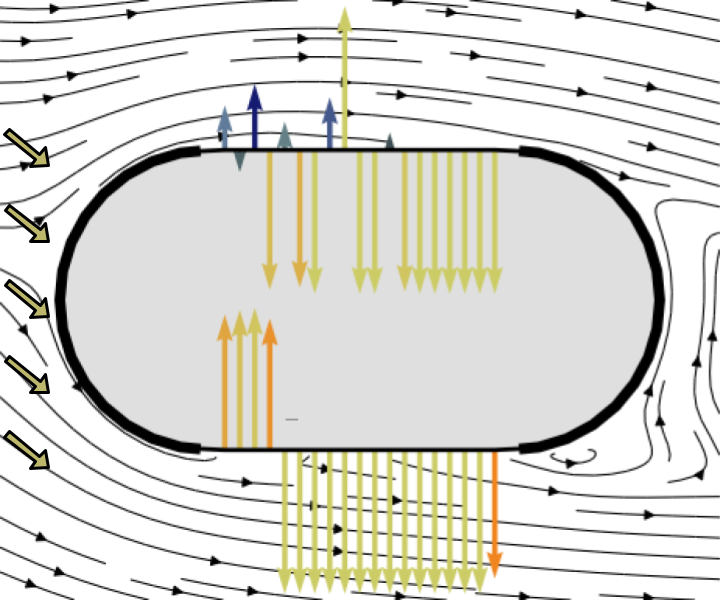}}
    \hspace{0.005cm}
    \subfloat{\includegraphics[height=0.32\linewidth]{Figures/NS_Jets/Colorbar_Action.png}}
    \subfloat{\includegraphics[height=0.32\linewidth]{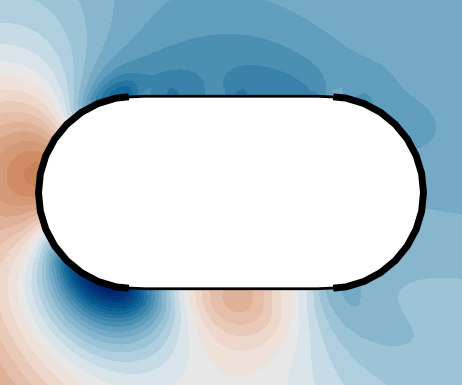}}
    \subfloat{\includegraphics[height=0.32\linewidth]{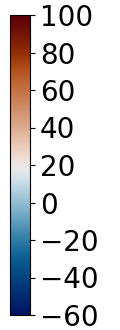}}
    \vspace{-0.3cm}
    
    \subfloat{\includegraphics[width=0.9\linewidth]{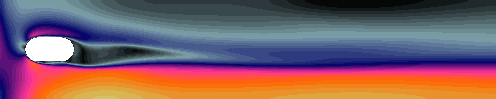}}
    \vspace{-0.4cm}

    \subfloat{\includegraphics[width=0.9\linewidth]{Figures/NS_Jets/Colorbar_Velocity.png}}
    \end{minipage}
    \hspace{-0.25cm}
    \begin{minipage}{0.33\linewidth}
    \subfloat{\includegraphics[height=0.32\linewidth]{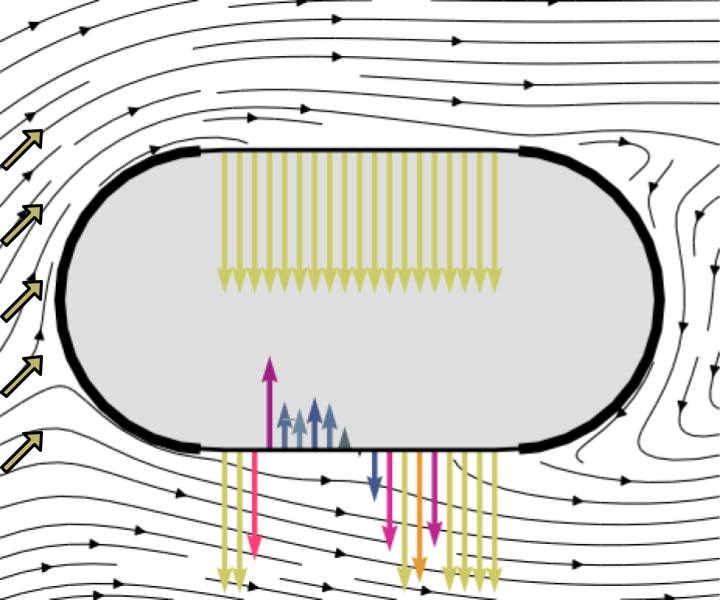}}
    \hspace{0.005cm}
    \subfloat{\includegraphics[height=0.32\linewidth]{Figures/NS_Jets/Colorbar_Action.png}}
    \subfloat{\includegraphics[height=0.32\linewidth]{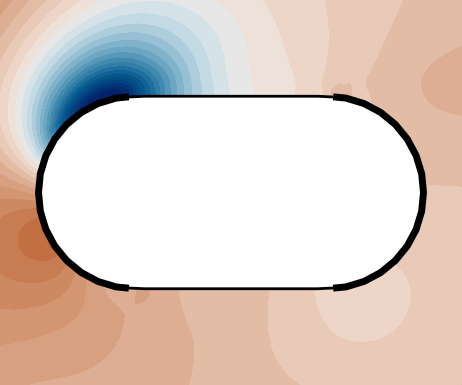}}
    \subfloat{\includegraphics[height=0.32\linewidth]{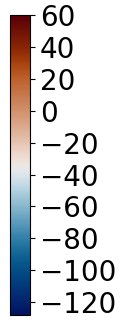}}
    \vspace{-0.3cm}
    
    \subfloat{\includegraphics[width=0.9\linewidth]{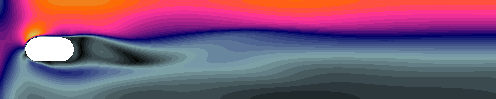}}
    \vspace{-0.4cm}

    \subfloat{\includegraphics[width=0.9\linewidth]{Figures/NS_Jets/Colorbar_Velocity.png}}
    \end{minipage}
    \vspace{-0.3cm}

    \begin{sideways} \makebox[0pt][l]{\hspace{-1.2cm} {\bf \footnotesize \phantom{Hype}MARL\phantom{Hype}}} \end{sideways}
    \begin{minipage}{0.33\linewidth}
    \subfloat{\includegraphics[height=0.32\linewidth]{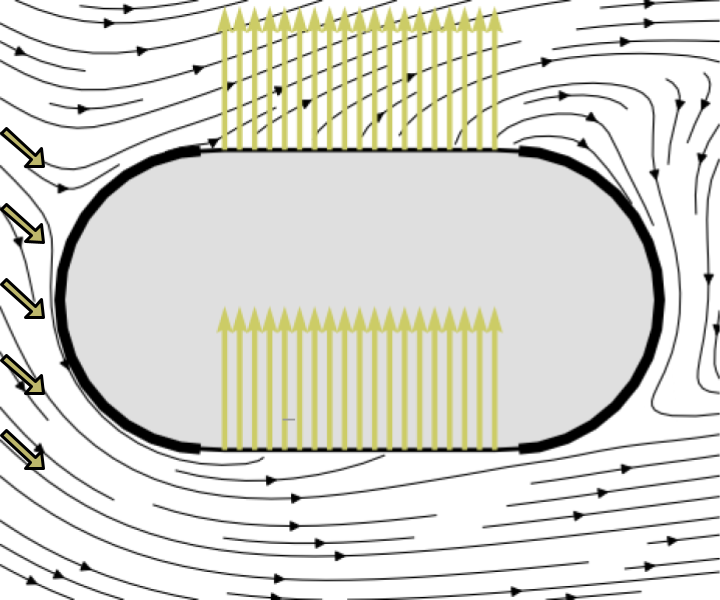}}
    \hspace{0.005cm}
    \subfloat{\includegraphics[height=0.32\linewidth]{Figures/NS_Jets/Colorbar_Action.png}}
    \subfloat{\includegraphics[height=0.32\linewidth]{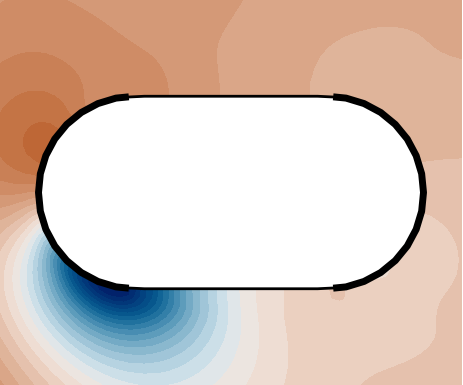}}
    \subfloat{\includegraphics[height=0.32\linewidth]{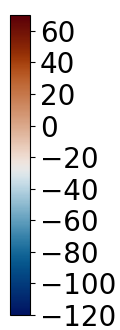}}
    \vspace{-0.3cm}
    
    \subfloat{\includegraphics[width=0.9\linewidth]{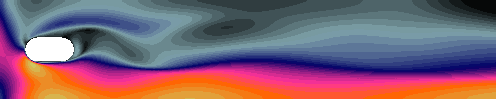}}
    \vspace{-0.4cm}

    \subfloat{\includegraphics[width=0.9\linewidth]{Figures/NS_Jets/Colorbar_Velocity.png}}
    \end{minipage}
    \hspace{-0.25cm}
    \begin{minipage}{0.33\linewidth}
    \subfloat{\includegraphics[height=0.32\linewidth]{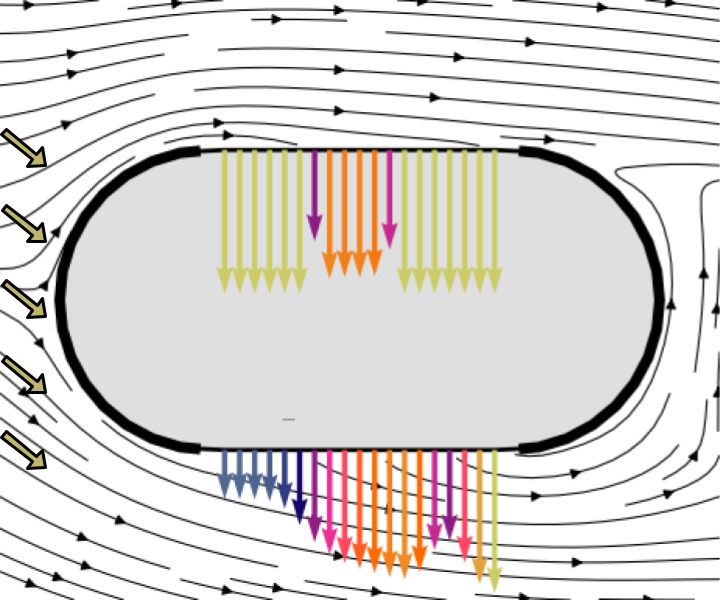}}
    \hspace{0.005cm}
    \subfloat{\includegraphics[height=0.32\linewidth]{Figures/NS_Jets/Colorbar_Action.png}}
    \subfloat{\includegraphics[height=0.32\linewidth]{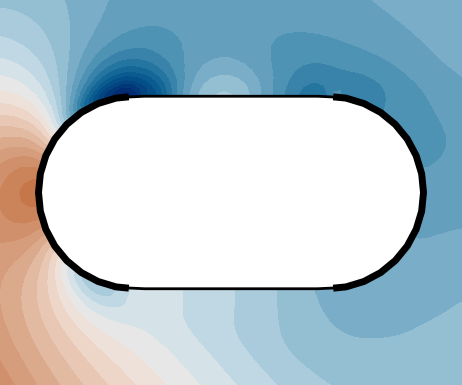}}
    \subfloat{\includegraphics[height=0.32\linewidth]{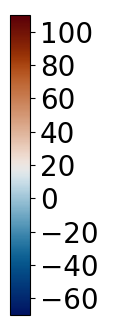}}
    \vspace{-0.3cm}
    
    \subfloat{\includegraphics[width=0.9\linewidth]{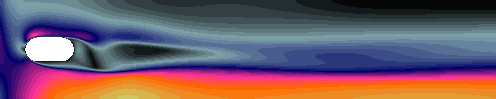}}
    \vspace{-0.4cm}

    \subfloat{\includegraphics[width=0.9\linewidth]{Figures/NS_Jets/Colorbar_Velocity.png}}
    \end{minipage}
    \hspace{-0.25cm}
    \begin{minipage}{0.33\linewidth}
    \subfloat{\includegraphics[height=0.32\linewidth]{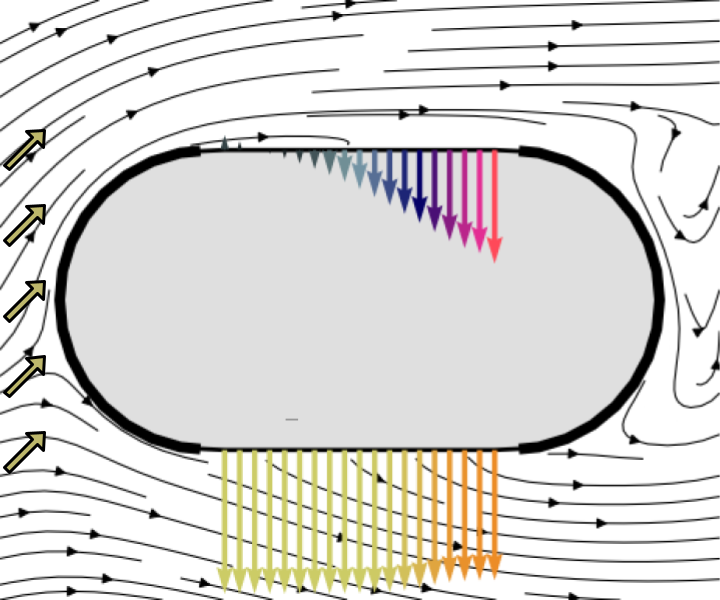}}
    \hspace{0.005cm}
    \subfloat{\includegraphics[height=0.32\linewidth]{Figures/NS_Jets/Colorbar_Action.png}}
    \subfloat{\includegraphics[height=0.32\linewidth]{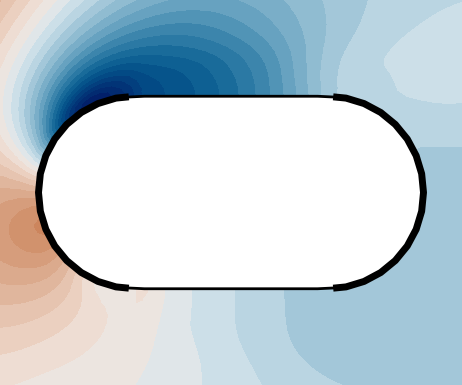}}
    \subfloat{\includegraphics[height=0.32\linewidth]{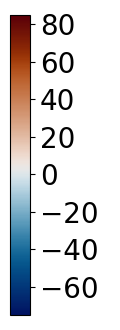}}
    \vspace{-0.3cm}
    
    \subfloat{\includegraphics[width=0.9\linewidth]{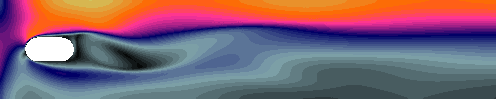}}
    \vspace{-0.4cm}

    \subfloat{\includegraphics[width=0.9\linewidth]{Figures/NS_Jets/Colorbar_Velocity.png}}
    \end{minipage}
    
    \caption{{\em Flow control with vertical jets}. Uncontrolled dynamics (first row), controlled dynamics obtained with HypeMARL (second row), MB-HypeMARL (third row) and MARL (fourth row) in three different random parametric settings. Every panel presents the vertical jets applied with the velocity streamlines around the obstacle (top left), the pressure around the obstacle (top right), and the flow velocity norm (bottom).}
    \label{fig:NS_Jets2}
\end{figure*}

Figure~\ref{fig:NS_Jets2} displays, instead, three examples of controlled solutions obtained with the different approaches for as many random angles of attack $\mu = \alpha$. Specifically, each subfigure shows the vertical jets with the velocity streamlines around the obstacle (top left), the pressure around the obstacle (top right), and the flow velocity (bottom).

\subsubsection{Flow control with shape morphing}

This last test case aims at controlling the flow around an object by morphing the front shape of the object itself. Similarly to the previous test case, we consider a fluid flow in the domain $\Omega(\mathbf{d}) = (0,10) \times (0,2) \setminus \Omega_{\text{obs}}(\mathbf{d})$. Starting from the shape introduced in the previous section, the obstacle $\Omega_{\text{obs}}(\mathbf{d}) = \Omega_{\text{l}}(\mathbf{d}) \cup \Omega_{\text{c}} \cup \Omega_{\text{r}}$ can now modify the round left boundary $\Gamma_{\text{l}}(\mathbf{d}) = \partial \Omega_{\text{obs}}(\mathbf{d}) \cap \{x_1 \leq 0.75\}$ according to a deformation field $\mathbf{d}(\mathbf{x},t) = d(\mathbf{x},t) \mathbf{n}_{\text{ref}}$, with intensity $d(\mathbf{x},t) \in [-0.1,0.1]$ and direction $\mathbf{n}_{\text{ref}}$, that is the normal versor to the initial boundary with no deformation. The flow dynamics is modeled by the unsteady Navier-Stokes equations
\begin{equation}
\label{eq:unsteadyNS1}
\begin{cases} 
\dfrac{\partial\mathbf{v}}{\partial t}-\nu \Delta\mathbf{v} + (\mathbf{v}\cdot\nabla)\mathbf{v} + \nabla p = \bm{0} \quad & \text{in} \ \Omega(\mathbf{d}) \times (0,T] \\ 
\nabla \cdot \mathbf{v} = 0 \quad & \text{in} \ \Omega(\mathbf{d}) \times (0,T] \\ 
\mathbf{v} = \bm{0} \quad & \text{on} \ \Gamma_{\text{c}} \cup \Gamma_{\text{r}} \\
\mathbf{v} = u \mathbf{n}_{\text{ref}} \quad & \text{on} \ \Gamma_{\text{l}}(\mathbf{d}) \\
\mathbf{v} = \mathbf{v}_{\text{in}}(\alpha) \quad & \text{on} \ \Gamma_{\text{in}} \\
\mathbf{v}\cdot\mathbf{n} = 0 \quad & \text{on} \ \Gamma_{\text{walls}} \\ 
(\nu \nabla \mathbf{v}-p)\mathbf{n} \cdot \mathbf{t} = 0 \quad & \text{on} \ \Gamma_{\text{walls}} \\
 (\nu \nabla \mathbf{v}-p)\mathbf{n} = \bm{0} \quad & \text{on} \ \Gamma_{\text{out}} \\
 \mathbf{v}(\mathbf{x},0) = \mathbf{0} \quad & \text{in} \ \Omega(\mathbf{d}) \, ,
\end{cases}
\end{equation}
with $\nu = 0.01$ the kinematic viscosity, and $u(\mathbf{x},t) \in [-0.01,0.01]$ the incremental deformation. Notice that, by discretizing the time interval $[0,T]$ with a uniform grid $t=1,...,T_{\text{max}}$, we have $d(\mathbf{x},\tau) = \sum_{t=1}^{\tau} u(\mathbf{x},t)$ for every $\tau = 1,...,T_{\text{max}}$. 
After observing the local pressure values $y_{i,t}=p_{i,t}$ at $N_y = 37$ points on $\Gamma_{l}(\mathbf{d})$, the controllers are allowed to deform the front part of the obstacle every $0.5\text{s}$ by applying a deformation $u_{i,t}$ in as many boundary points ($N_u = 37$). In this setting, the goal is to minimize the local pressure in front of the obstacle, that is
\[
r_{i,t} = r(p_{i,t}, u_{i,t}) = - p_{i,t} \ .
\]
See the previous section for further details on domain boundaries, space-time discretization and inflow velocity. Similarly to the previous test case, the angle of attack $\alpha \in [-1,1] \text{rad}$ is regarded as a system parameter, i.e. $\mu = \alpha$, and the Reynolds number ranges from $900$ to $1100$, according to the obstacle deformation.

\begin{figure*}
    
    \begin{minipage}{0.49\linewidth}
    \centering \subfloat{\includegraphics[height=0.5\textwidth]{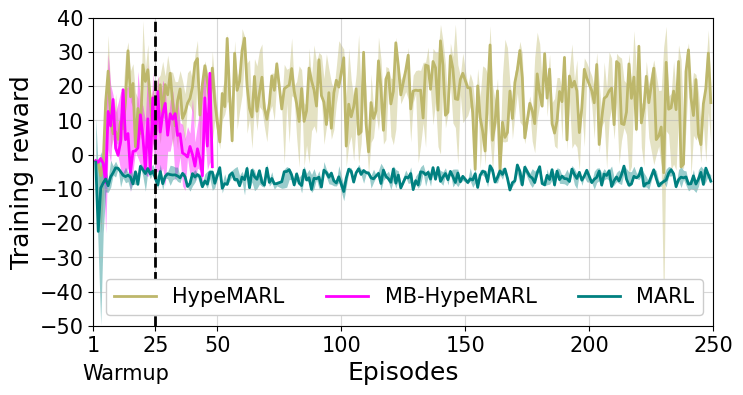}}
    \vspace{-0.5cm}
    
    \subfloat{\includegraphics[height=0.5\textwidth]{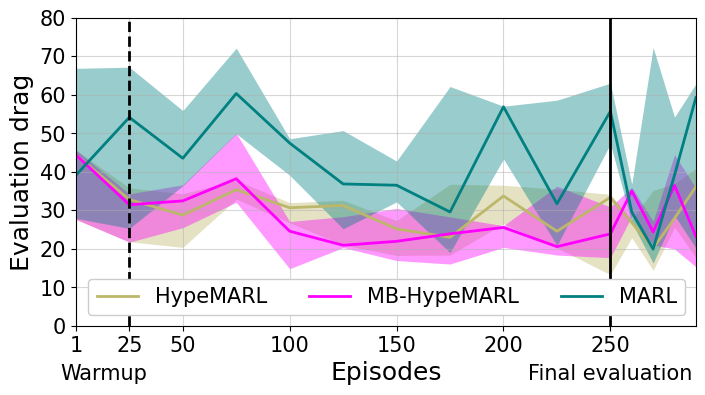}}
    \end{minipage}
    \begin{minipage}{0.49\linewidth}
    \centering \subfloat{\includegraphics[height=0.5\textwidth]{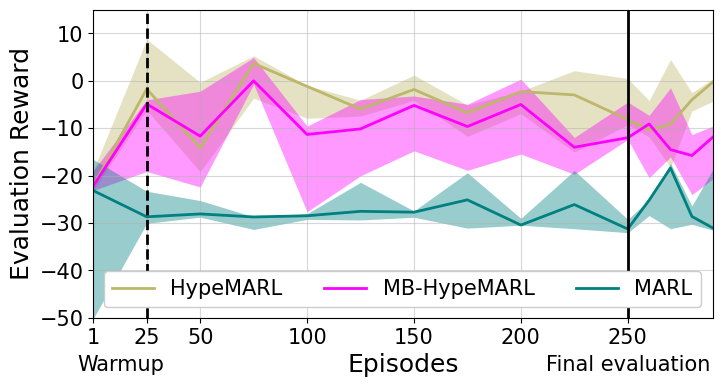}}
    \vspace{-0.5cm}
    
    \subfloat{\includegraphics[height=0.5\textwidth]{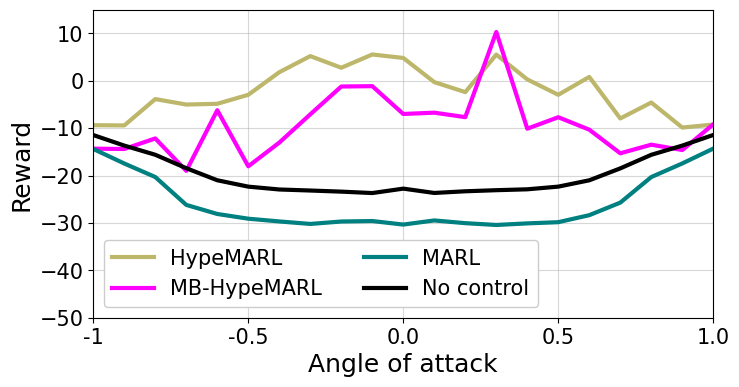}}
    \end{minipage}

    \caption{{\em Flow control with shape morphing}. First row: median and interquartile range of HypeMARL, MB-HypeMARL and MARL training end evaluation rewards. Second row: median and interquartile range of HypeMARL, MB-HypeMARL and MARL evaluation drags (left) and reward obtained with HypeMARL, MB-HypeMARL, MARL and no control for different angles of attack (right).}
    \label{fig:NS_Morphing1}
\end{figure*}

The first row of Figure~\ref{fig:NS_Morphing1} presents the median and interquartile range of training and evaluation rewards obtained with $5$ independent runs with $5$ different seeds. HypeMARL and MB-HypeMARL agents outperform MARL with higher training and evaluation rewards, and thus more effective morphing strategies. Moreover, as visible in the left panel in the second row of Figure~\ref{fig:NS_Morphing1}, despite the simplicity of the reward function, HypeMARL and MB-HypeMARL indirectly allow for smaller drag values across different parametric settings with respect to classical decentralized MARL. The right panel in the second row of Figure~\ref{fig:NS_Morphing1} shows the rewards of the different competing models for different angles of attack in the range $[-1,1]\text{rad}$, along with the ones in the uncontrolled setting.

\begin{figure*}
    
    \begin{sideways} \makebox[0pt][l]{\hspace{-0.7cm} {\bf \footnotesize No control}} \end{sideways}
    \begin{minipage}{0.33\linewidth}
    \hspace{-0.17cm}
    \subfloat{\begin{overpic}[height=0.345\linewidth]{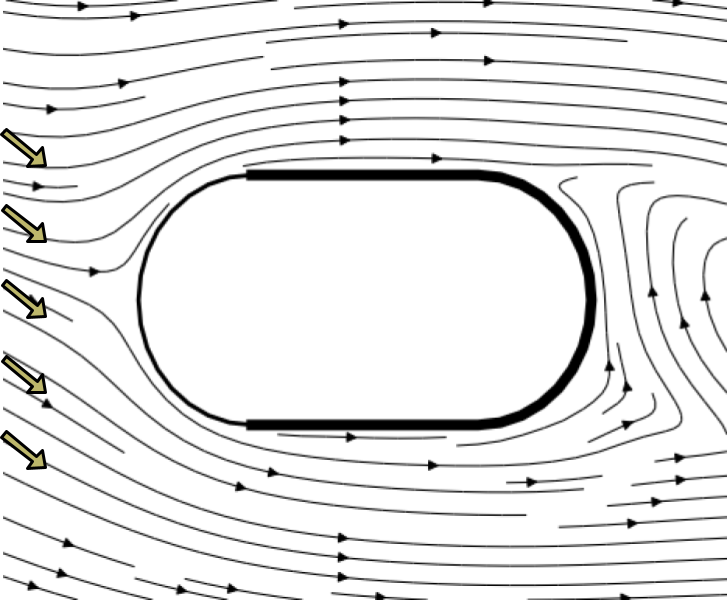}\put(65,90){\shortstack[c]{$\alpha = -0.5\text{rad}$ \\ $t=4.0\text{s}$}}\end{overpic}}
    \hspace{0.01cm}
    \subfloat{\includegraphics[height=0.345\linewidth]{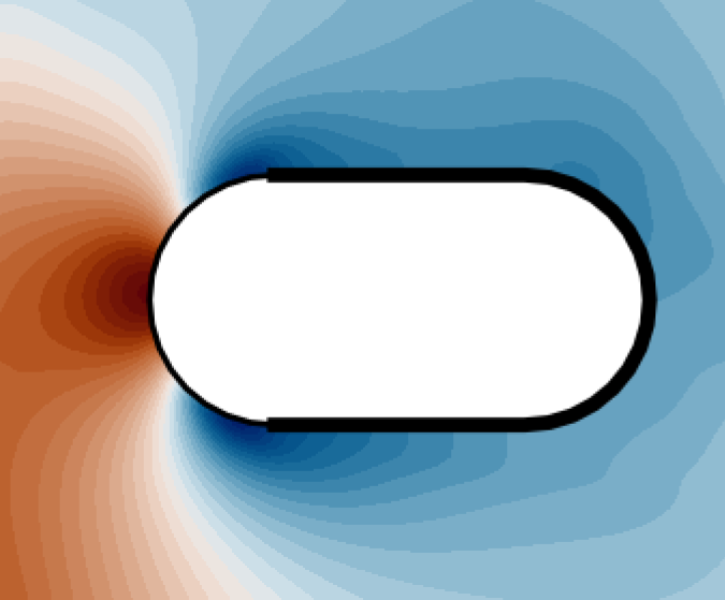}}
    \subfloat{\includegraphics[height=0.345\linewidth]{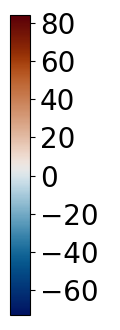}}
    \vspace{-0.3cm}
    
    \subfloat{\includegraphics[width=0.9\linewidth]{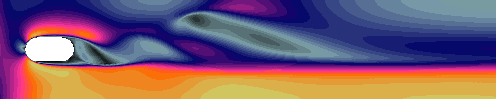}}
    \vspace{-0.4cm}

    \subfloat{\includegraphics[width=0.9\linewidth]{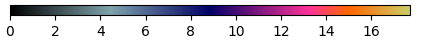}}
    \end{minipage}
    \hspace{-0.25cm}
    \begin{minipage}{0.33\linewidth}
    \hspace{-0.17cm}
    \subfloat{\begin{overpic}[height=0.345\linewidth]{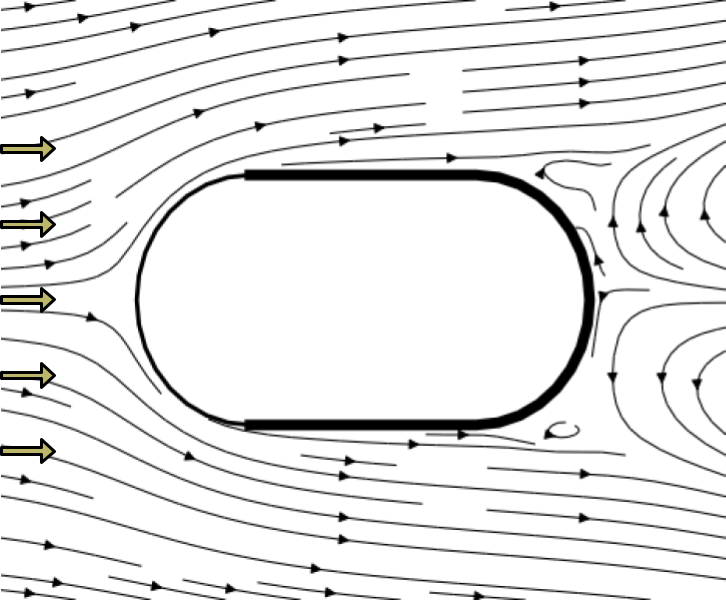}\put(65,90){\shortstack[c]{$\alpha = 0.0\text{rad}$ \\ $t=4.0\text{s}$}}\end{overpic}}
    \hspace{0.01cm}
    \subfloat{\includegraphics[height=0.345\linewidth]{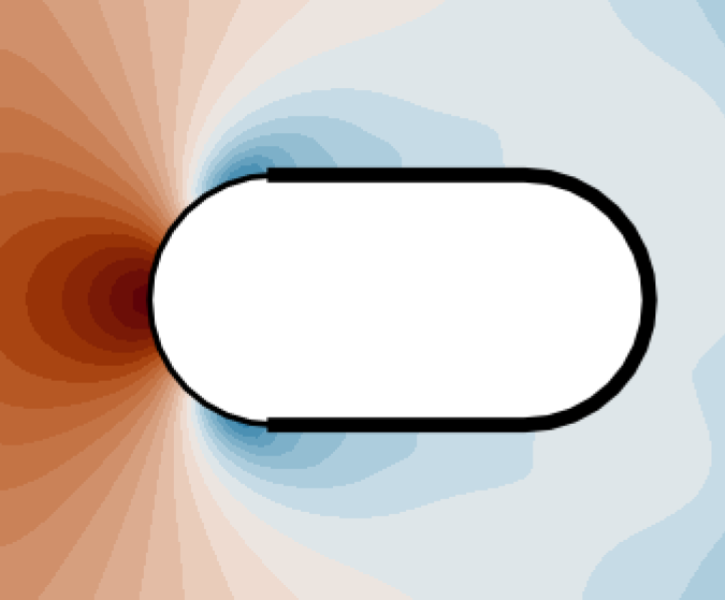}}
    \subfloat{\includegraphics[height=0.345\linewidth]{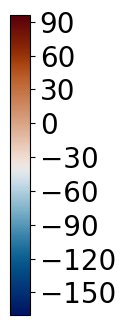}}
    \vspace{-0.3cm}
    
    \subfloat{\includegraphics[width=0.9\linewidth]{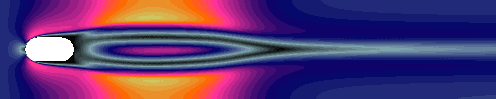}}
    \vspace{-0.4cm}

    \subfloat{\includegraphics[width=0.9\linewidth]{Figures/NS_Morphing/Colorbar_Velocity.png}}
    \end{minipage}
    \hspace{-0.25cm}
    \begin{minipage}{0.33\linewidth}
    \hspace{-0.17cm}
    \subfloat{\begin{overpic}[height=0.345\linewidth]{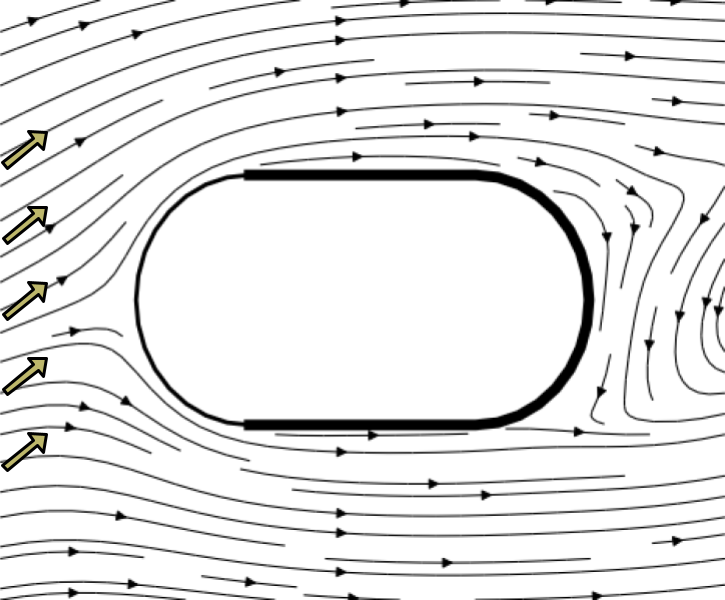}\put(65,90){\shortstack[c]{$\alpha = 0.5\text{rad}$ \\ $t=4.0\text{s}$}}\end{overpic}}
    \hspace{0.01cm}
    \subfloat{\includegraphics[height=0.345\linewidth]{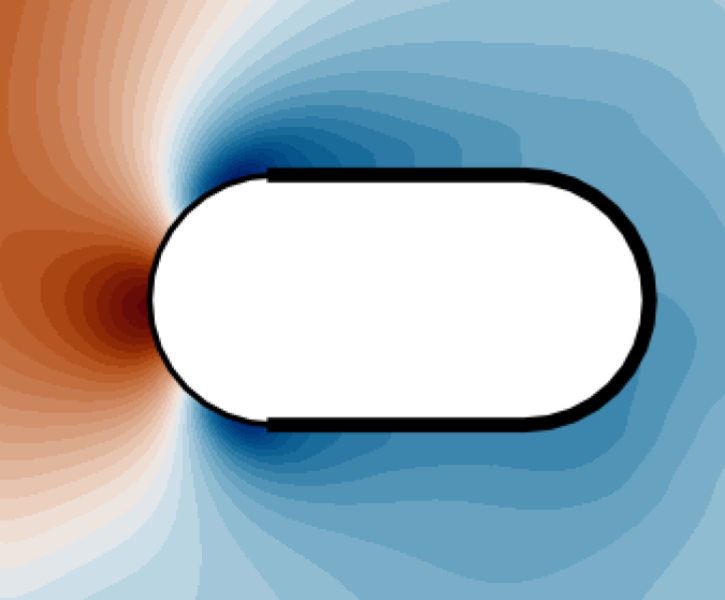}}
    \subfloat{\includegraphics[height=0.345\linewidth]{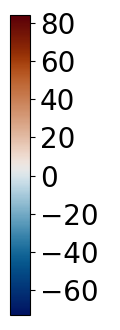}}
    \vspace{-0.3cm}
    
    \subfloat{\includegraphics[width=0.9\linewidth]{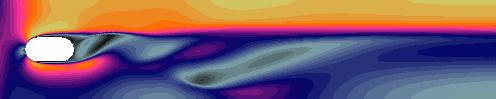}}
    \vspace{-0.4cm}

    \subfloat{\includegraphics[width=0.9\linewidth]{Figures/NS_Morphing/Colorbar_Velocity.png}}
    \end{minipage}
    \vspace{-0.3cm}

    \begin{sideways} \makebox[0pt][l]{\hspace{-0.7cm} {\bf \footnotesize HypeMARL}} \end{sideways}
    \begin{minipage}{0.33\linewidth}
    \subfloat{\includegraphics[height=0.345\linewidth]{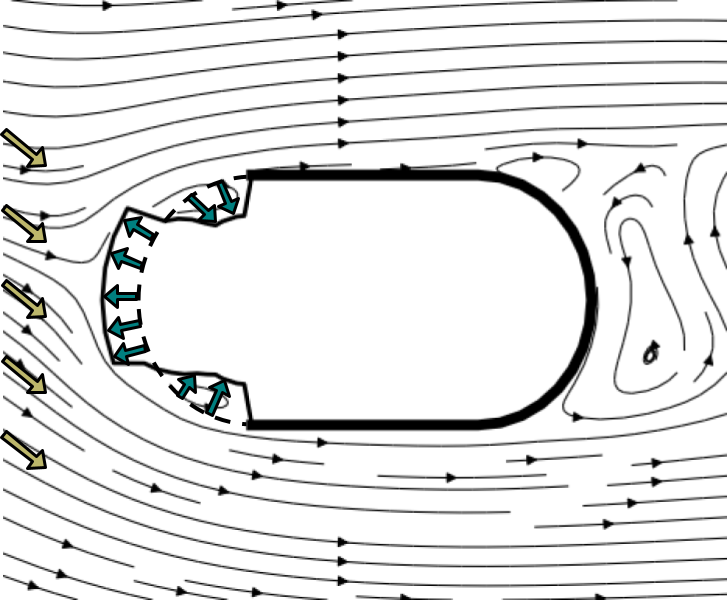}}
    \hspace{0.01cm}
    \subfloat{\includegraphics[height=0.345\linewidth]{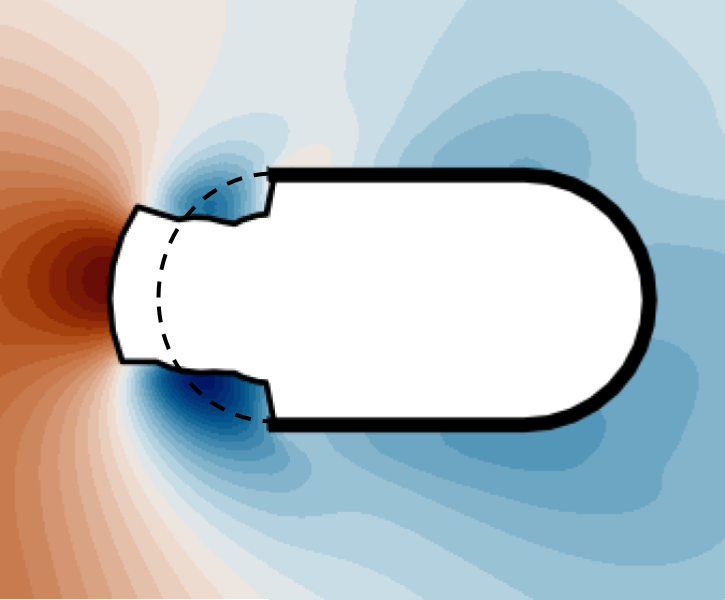}}
    \subfloat{\includegraphics[height=0.345\linewidth]{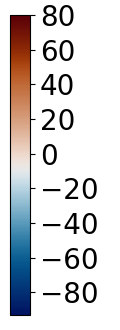}}
    \vspace{-0.3cm}
    
    \subfloat{\includegraphics[width=0.9\linewidth]{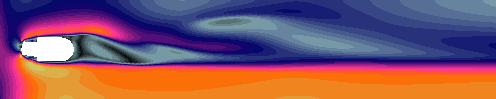}}
    \vspace{-0.4cm}

    \subfloat{\includegraphics[width=0.9\linewidth]{Figures/NS_Morphing/Colorbar_Velocity.png}}
    \end{minipage}
    \hspace{-0.25cm}
    \begin{minipage}{0.33\linewidth}
    \subfloat{\includegraphics[height=0.345\linewidth]{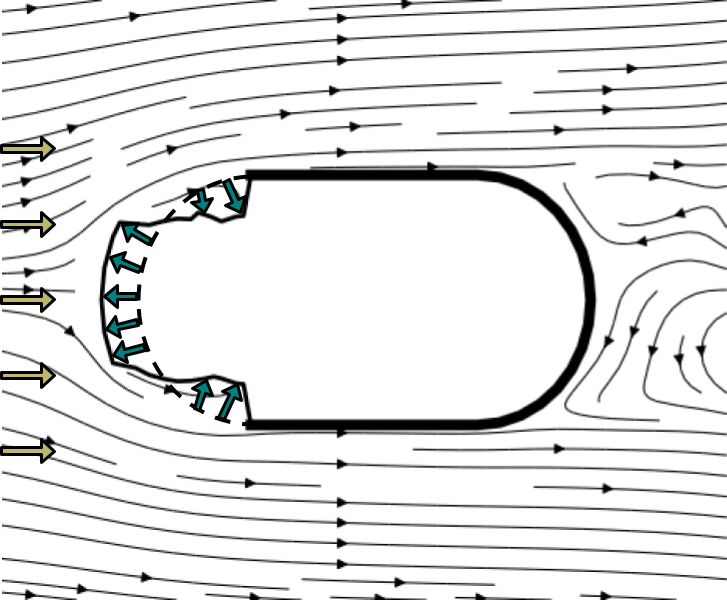}}
    \hspace{0.01cm}
    \subfloat{\includegraphics[height=0.345\linewidth]{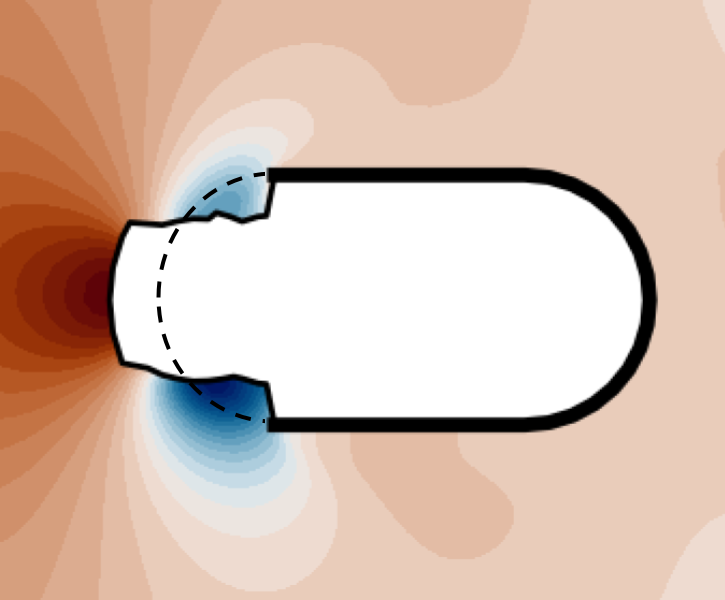}}
    \subfloat{\includegraphics[height=0.345\linewidth]{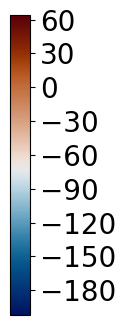}}
    \vspace{-0.3cm}
    
    \subfloat{\includegraphics[width=0.9\linewidth]{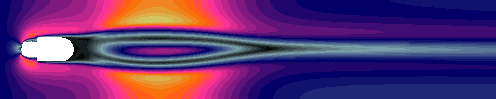}}
    \vspace{-0.4cm}

    \subfloat{\includegraphics[width=0.9\linewidth]{Figures/NS_Morphing/Colorbar_Velocity.png}}
    \end{minipage}
    \hspace{-0.25cm}
    \begin{minipage}{0.33\linewidth}
    \subfloat{\includegraphics[height=0.345\linewidth]{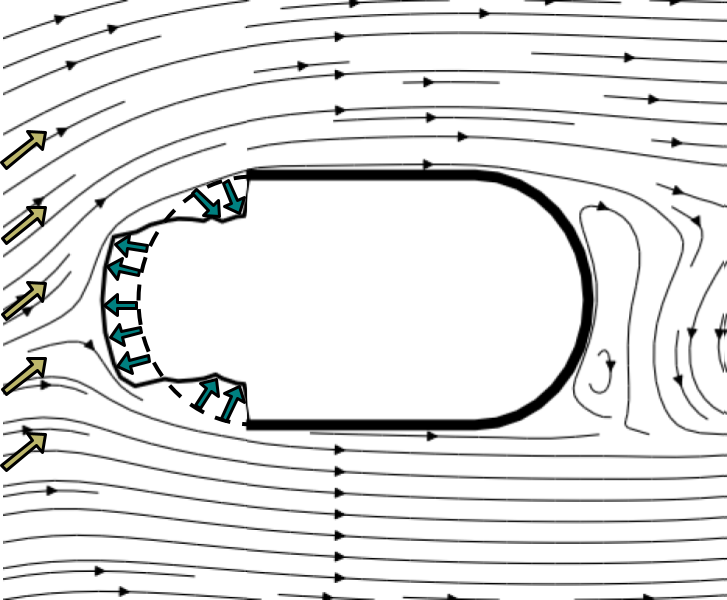}}
    \hspace{0.01cm}
    \subfloat{\includegraphics[height=0.345\linewidth]{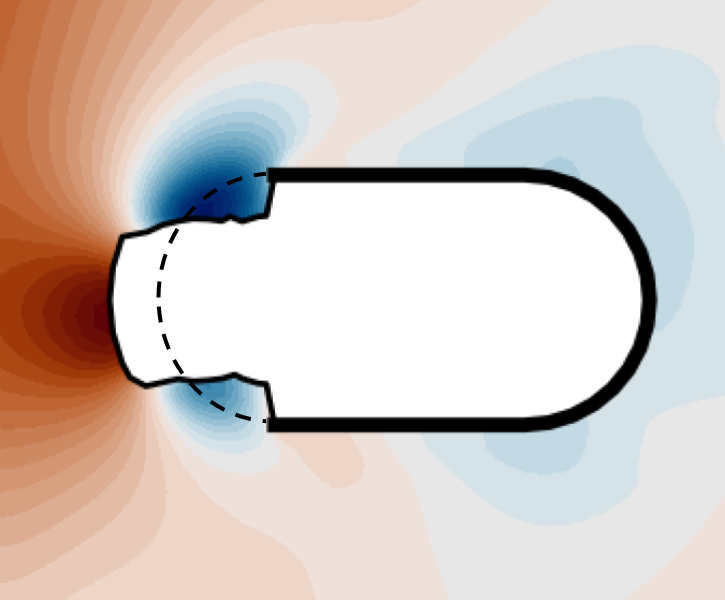}}
    \subfloat{\includegraphics[height=0.345\linewidth]{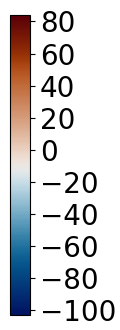}}
    \vspace{-0.3cm}
    
    \subfloat{\includegraphics[width=0.9\linewidth]{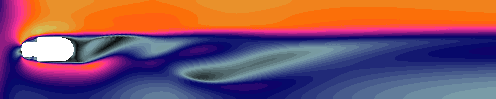}}
    \vspace{-0.4cm}

    \subfloat{\includegraphics[width=0.9\linewidth]{Figures/NS_Morphing/Colorbar_Velocity.png}}
    \end{minipage}
    \vspace{-0.3cm}

    \begin{sideways} \makebox[0pt][l]{\hspace{-1.1cm} {\bf \footnotesize MB-HypeMARL}} \end{sideways}
    \begin{minipage}{0.33\linewidth}
    \subfloat{\includegraphics[height=0.345\linewidth]{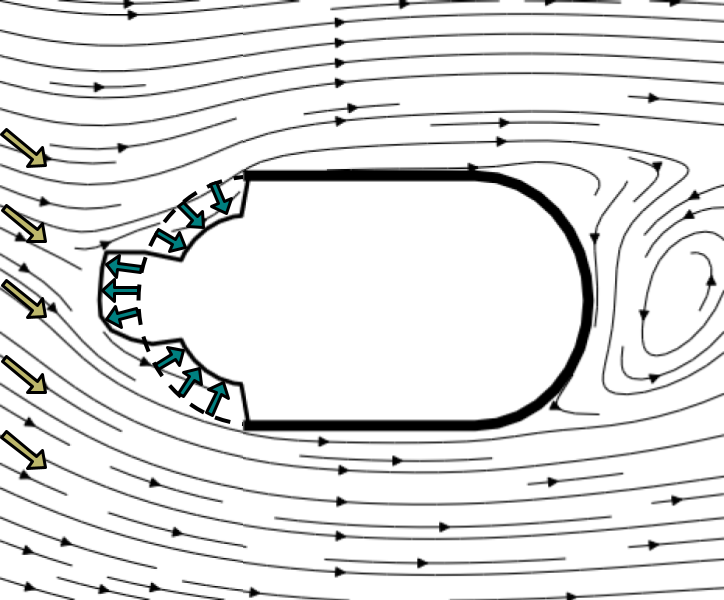}}
    \hspace{0.01cm}
    \subfloat{\includegraphics[height=0.345\linewidth]{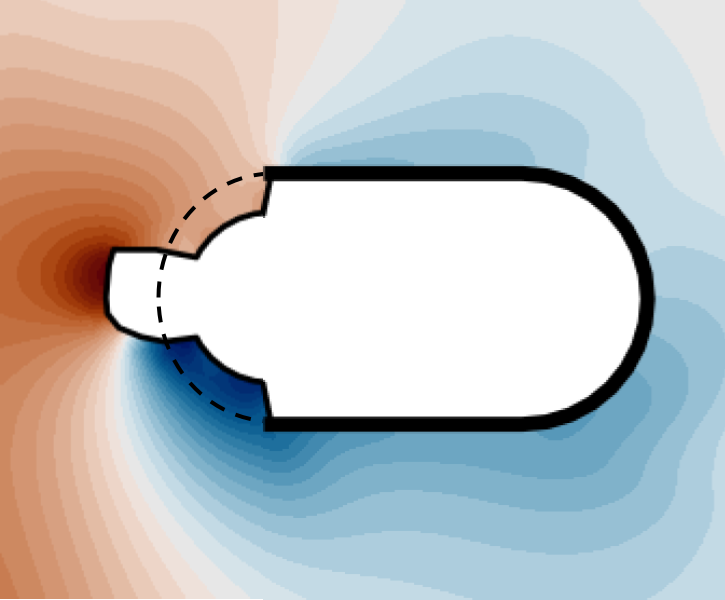}}
    \subfloat{\includegraphics[height=0.345\linewidth]{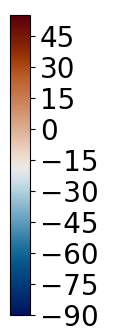}}
    \vspace{-0.3cm}
    
    \subfloat{\includegraphics[width=0.9\linewidth]{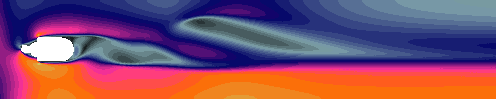}}
    \vspace{-0.4cm}

    \subfloat{\includegraphics[width=0.9\linewidth]{Figures/NS_Morphing/Colorbar_Velocity.png}}
    \end{minipage}
    \hspace{-0.25cm}
    \begin{minipage}{0.33\linewidth}
    \subfloat{\includegraphics[height=0.345\linewidth]{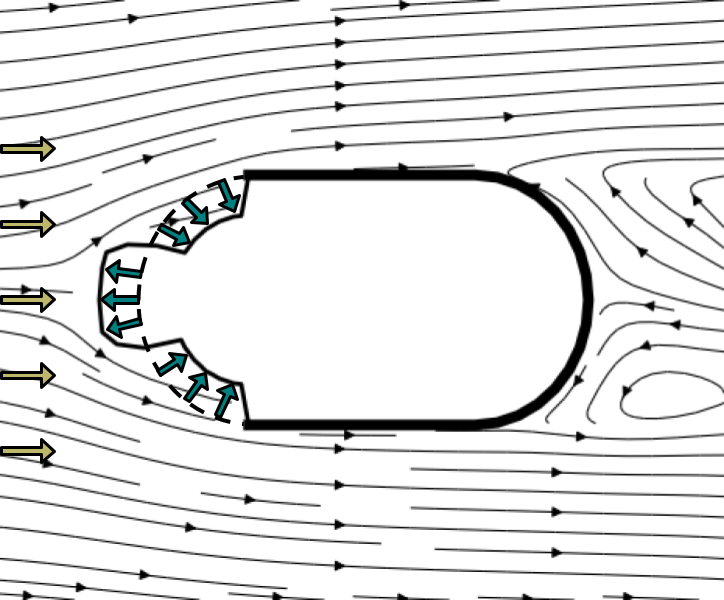}}
    \hspace{0.01cm}
    \subfloat{\includegraphics[height=0.345\linewidth]{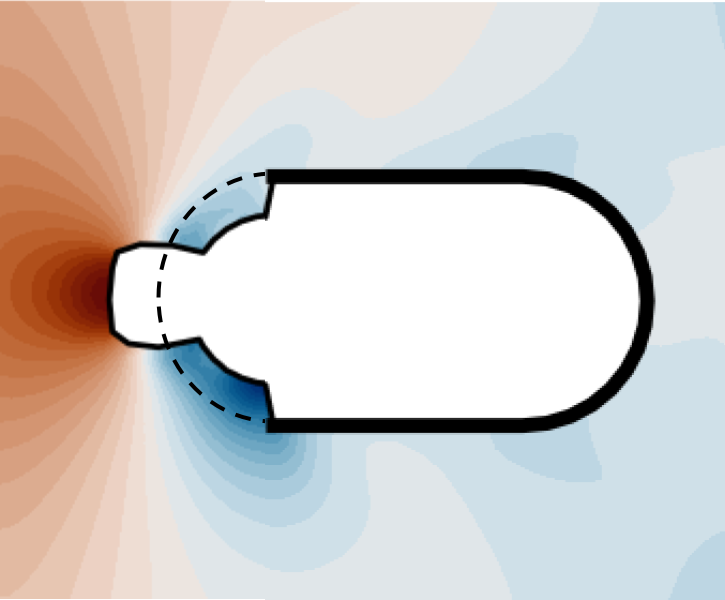}}
    \subfloat{\includegraphics[height=0.345\linewidth]{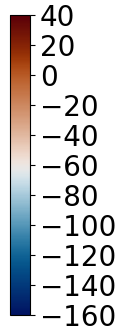}}
    \vspace{-0.3cm}
    
    \subfloat{\includegraphics[width=0.9\linewidth]{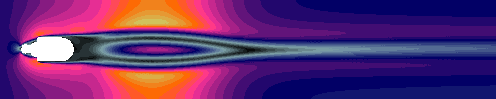}}
    \vspace{-0.4cm}

    \subfloat{\includegraphics[width=0.9\linewidth]{Figures/NS_Morphing/Colorbar_Velocity.png}}
    \end{minipage}
    \hspace{-0.25cm}
    \begin{minipage}{0.33\linewidth}
    \subfloat{\includegraphics[height=0.345\linewidth]{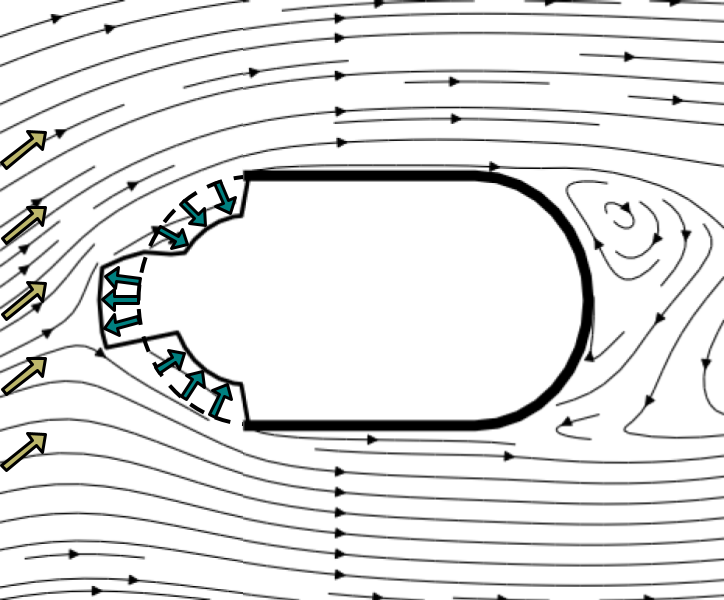}}
    \hspace{0.01cm}
    \subfloat{\includegraphics[height=0.345\linewidth]{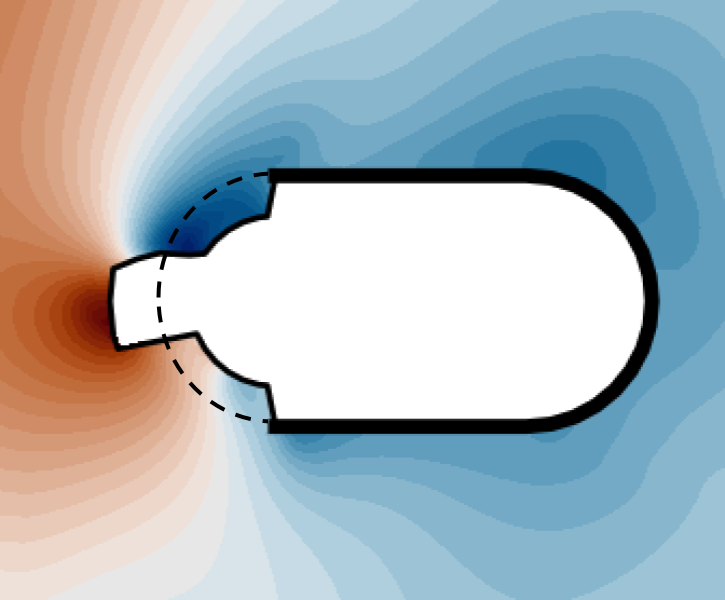}}
    \subfloat{\includegraphics[height=0.345\linewidth]{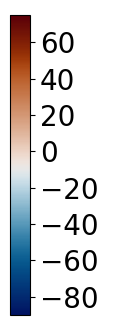}}
    \vspace{-0.3cm}
    
    \subfloat{\includegraphics[width=0.9\linewidth]{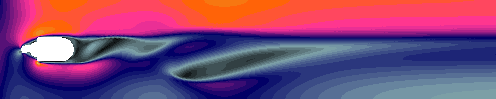}}
    \vspace{-0.4cm}

    \subfloat{\includegraphics[width=0.9\linewidth]{Figures/NS_Morphing/Colorbar_Velocity.png}}
    \end{minipage}
    \vspace{-0.3cm}

    \begin{sideways} \makebox[0pt][l]{\hspace{-1.2cm} {\bf \footnotesize \phantom{Hype}MARL\phantom{Hype}}} \end{sideways}
    \begin{minipage}{0.33\linewidth}
    \subfloat{\includegraphics[height=0.345\linewidth]{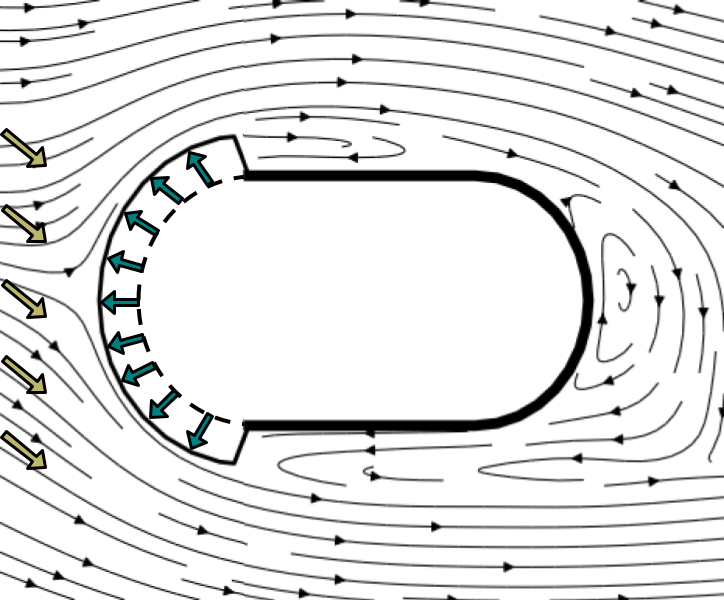}}
    \hspace{0.01cm}
    \subfloat{\includegraphics[height=0.345\linewidth]{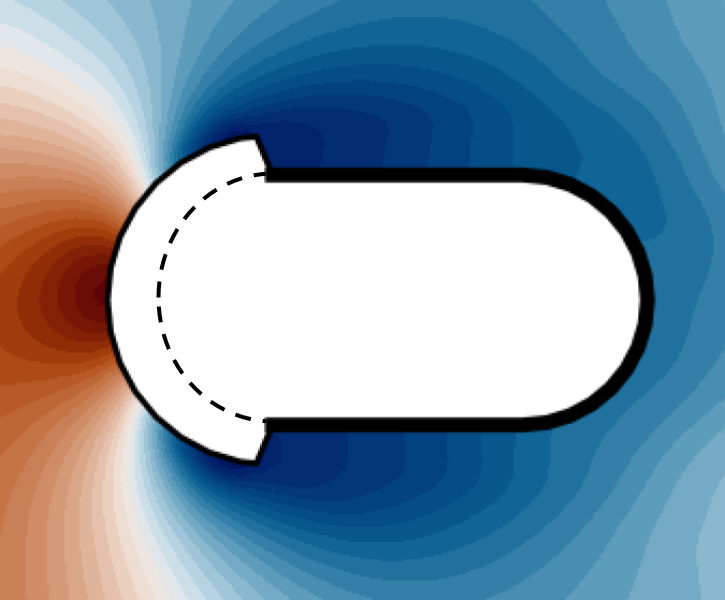}}
    \subfloat{\includegraphics[height=0.345\linewidth]{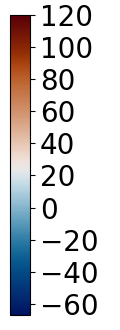}}
    \vspace{-0.3cm}
    
    \subfloat{\includegraphics[width=0.9\linewidth]{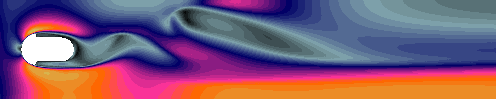}}
    \vspace{-0.4cm}

    \subfloat{\includegraphics[width=0.9\linewidth]{Figures/NS_Morphing/Colorbar_Velocity.png}}
    \end{minipage}
    \hspace{-0.25cm}
    \begin{minipage}{0.33\linewidth}
    \subfloat{\includegraphics[height=0.345\linewidth]{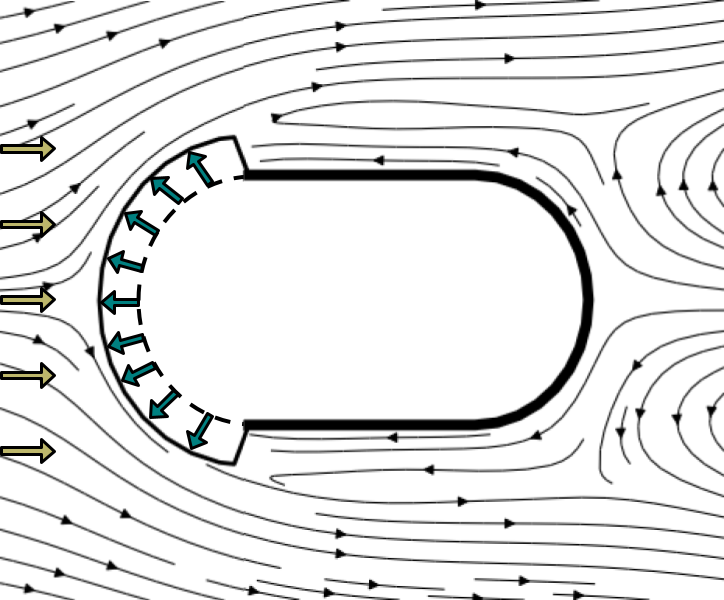}}
    \hspace{0.01cm}
    \subfloat{\includegraphics[height=0.345\linewidth]{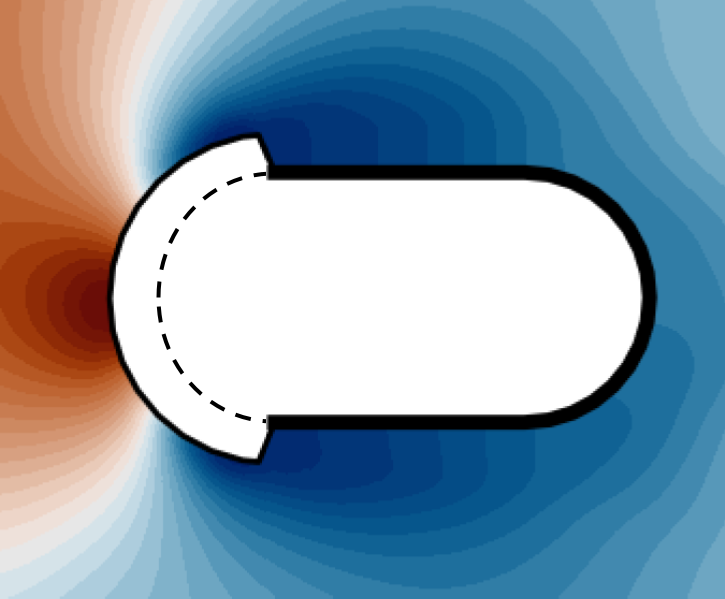}}
    \subfloat{\includegraphics[height=0.345\linewidth]{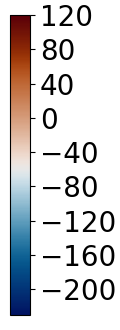}}
    \vspace{-0.3cm}
    
    \subfloat{\includegraphics[width=0.9\linewidth]{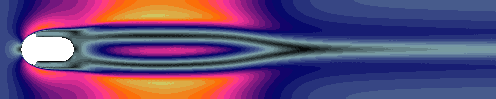}}
    \vspace{-0.4cm}

    \subfloat{\includegraphics[width=0.9\linewidth]{Figures/NS_Morphing/Colorbar_Velocity.png}}
    \end{minipage}
    \hspace{-0.25cm}
    \begin{minipage}{0.33\linewidth}
    \subfloat{\includegraphics[height=0.345\linewidth]{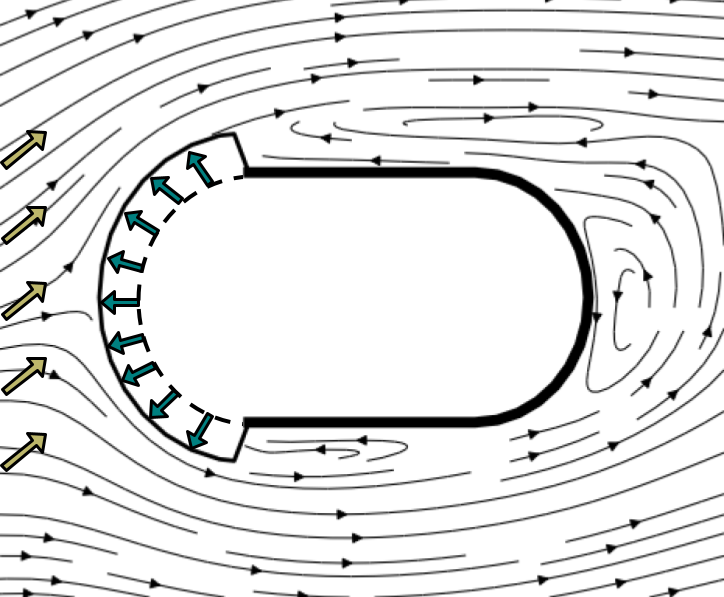}}
    \hspace{0.01cm}
    \subfloat{\includegraphics[height=0.345\linewidth]{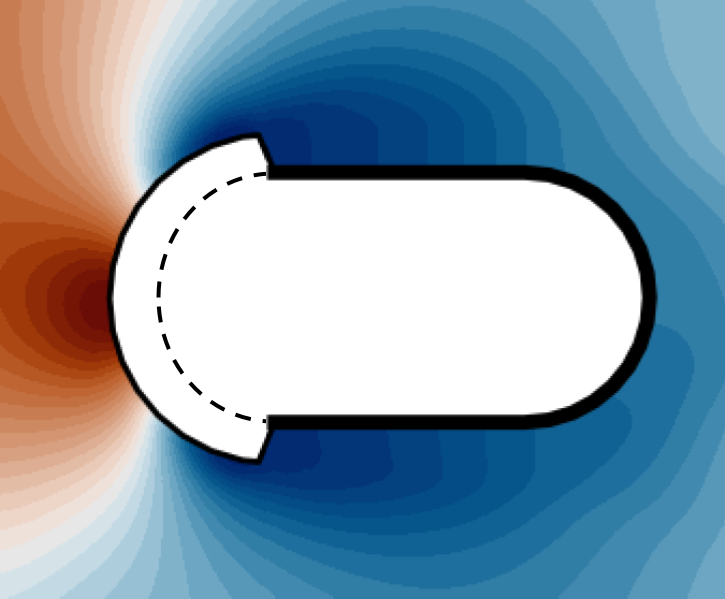}}
    \subfloat{\includegraphics[height=0.345\linewidth]{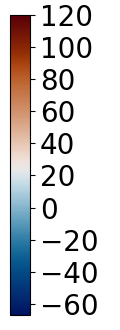}}
    \vspace{-0.3cm}
    
    \subfloat{\includegraphics[width=0.9\linewidth]{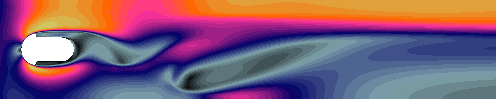}}
    \vspace{-0.4cm}

    \subfloat{\includegraphics[width=0.9\linewidth]{Figures/NS_Morphing/Colorbar_Velocity.png}}
    \end{minipage}
    
    \caption{{\em Flow control with shape morphing}. Uncontrolled dynamics (first row), controlled dynamics obtained with HypeMARL (second row), MB-HypeMARL (third row) and MARL (fourth row) in three different parametric settings. Every panel presents the deformed obstacle shape with the velocity streamlines around the obstacle (top left), the pressure around the obstacle (top right), and the flow velocity norm (bottom).}
    \label{fig:NS_Morphing2}
\end{figure*}

Figure~\ref{fig:NS_Morphing2} shows three examples of shape morphing when facing as many values of the angle of attack, i.e. $\alpha = \{-0.5, 0.0, 0.5\}\text{rad}$. Specifically, it is possible to assess that HypeMARL and MB-HypeMARL are able to learn non-trivial narrow front shapes, allowing for local pressure minimization. Moreover, the resulting shapes clearly exploit the parametric knowledge to adapt the front orientation with respect to the different angles of attack $\alpha$. Conversely, MARL is only capable of finding a trivial shape corresponding to the maximum displacement allowed, without any adaptation corresponding to the different parameter values. Similarly to the previous test cases, MB-HypeMARL is again capable of reducing the number of real environment interactions, with a minimal performance deterioration with respect to HypeMARL. 

While the results obtained in this very challenging example by HypeMARL and MB-HypeMARL are interesting and promising, as they show a collective and coordinated behavior of the agents, it is worth highlighting that the shapes found are not the optimal shapes usually obtained in the context of shape optimization in fluid flows. Improving the shape smoothness requires higher degrees of freedom in the boundary displacements. In our experiments, we only consider radial displacements of the front part of the object, limiting the shapes that can be obtained. Moreover, feasibility constraints should be taken into account to guarantee physical and geometrical admissibility of the resulting shape. In addition, the obstacle aerodynamics could be enhanced by maximizing global reward metrics, such as promoting drag reduction or penalizing non-smooth shapes. However, the maximization of global reward functions in decentralized settings is extremely challenging, and it will be the subject of future works, as highlighted in Section \ref{sec:discussion}.

\section{Conclusions}\label{sec:discussion}
In this work, we propose a novel decentralized MARL approach, named HypeMARL, that allows for solving challenging high-dimensional, parametric, and distributed control problems. HypeMARL relies on {\em (i)} sinusoidal positional encoding of the agents' relative position to enhance spatial awareness and coordination among the agents, and {\em (ii)} hypernetworks to effectively parametrize the control policies as functions of system parameters and positional information. Moreover, we propose a model-based version of the algorithm, named MB-HypeMARL, allowing for a significant reduction of interactions with the real environment (by a factor $\sim 10$) thanks to a cheap deep learning-based surrogate model approximating local dynamics.

In numerical experiments dealing with two challenging classes of high-dimensional control problems, namely density and flow control, we show that both HypeMARL and MB-HypeMARL can effectively solve problems that the state-of-the-art decentralized MARL approach cannot. Specifically, our results demonstrate that hypernetworks combined with the sinusoidal embedding of the agents' (relative) position allow for strategic coordination strategies among the agents by only relying on local state and reward information. The model-based algorithm MB-HypeMARL is especially appealing for applications where only a limited amount of data is available, as the agents rely on surrogate models approximating the local dynamics to reduce the need for real systems interactions. In this work, local surrogate models are modeled through shallow neural networks and show impressive accuracy even if trained with limited data, opening new doors towards scaling reinforcement learning to complex real-world engineering systems.

In particular, the control of density functions is extremely relevant in several scientific fields. For example, a density function can represent a robot swarm and its optimal control can have impact in security, survey, monitoring, and surveillance applications \cite{tahir2019swarms}. Density functions can also model wildlife, or traffic flows. In all these cases, developing (decentralized) algorithms that can achieve coordination among the agents and collective behavior while scaling to large number of agents, such as the ones learned by HypeMARL and MB-HypeMARL, is essential for solving complex problems. While the high-dimensionality of the states is not a limiting factor for state-of-the-art deep reinforcement learning algorithms, dealing with large actions spaces still remains an open challenge.

The ability to manipulate a flow field actively is one the most import technological challenges of the present and past decades. Flow control involves transition delay, separation postponement, lift enhancement, turbulence augmentation, noise suppression and drag reduction \cite{gad2001flow}. Drag reduction is especially critical for reducing fuel consumption of land, air, and sea vehicles and emissions. However, actuation strategies such as jets, or morphing structures typically require coordination among a large number of distributed actuators and, consequently, control algorithms capable of dealing with high-dimensional control problems. In addition, collecting data from real-world flow experiments, for example in a wind tunnel, is extremely costly and tedious. At the same time, full-order models capable of simulating fluid flows or fluid-structure interactions are computationally expensive. Data-efficient approaches as HypeMARL and MB-HypeMARL can lay the foundations towards the next generation of energy-efficient vehicles through active flow control.     

The underlying assumption of decentralized MARL approaches is that the maximization of the cumulative local rewards corresponds to the maximization of a global task-specific metric. Although it is feasible to design such reward functions for various problems, it is not always straightforward. Indeed, local rewards and global task-specific metrics may be conflicting incentives due to the effect of the agents' interactions. Moreover, the credit assignment problem may arise whenever local rewards are uninformative about global performance. While HypeMARL and MB-HypeMARL prioritize collective behavior of the agents over immediate local payoff, the direct maximization of global reward functions will be the subject of future works. Another interesting future direction is the learning of local reward functions corresponding to global reward metrics from given expert demonstrations, in the direction of Inverse RL \cite{ng2000algorithms} and imitation learning \cite{hussein2017imitation}.

\section*{Acknowledgements}

NB and AM acknowledge the Project “Reduced Order Modeling and Deep Learning for the real- time approximation of PDEs (DREAM)” (Starting Grant No. FIS00003154), funded by the Italian Science Fund (FIS) - Ministero dell'Università e della Ricerca.  AM also acknowledges the project “Dipartimento di Eccellenza” 2023-2027 funded by MUR and the project FAIR (Future Artificial Intelligence Research), funded by the NextGenerationEU program within the PNRR-PE-AI scheme (M4C2, Investment 1.3, Line on Artificial Intelligence).

\bibliographystyle{unsrt}  
\bibliography{references}  

\begin{thebibliography}{10}

\bibitem{van2011control}
Jan~H van Schuppen, Olivier Boutin, Pia~L Kempker, Jan Komenda, Tom{\'a}{\v{s}} Masopust, Nicola Pambakian, and Andr{\'e}~CM Ran.
\newblock Control of distributed systems: Tutorial and overview.
\newblock {\em European Journal of Control}, 17(5-6):579--602, 2011.

\bibitem{bucsoniu2010multi}
Lucian Bu{\c{s}}oniu, Robert Babu{\v{s}}ka, and Bart De~Schutter.
\newblock Multi-agent reinforcement learning: An overview.
\newblock {\em Innovations in multi-agent systems and applications-1}, pages 183--221, 2010.

\bibitem{sutton2018reinforcement}
Richard~S Sutton and Andrew~G Barto.
\newblock {\em Reinforcement learning: An introduction}.
\newblock MIT press, 2018.

\bibitem{sutton1999policy}
Richard~S Sutton, David McAllester, Satinder Singh, and Yishay Mansour.
\newblock Policy gradient methods for reinforcement learning with function approximation.
\newblock {\em Advances in neural information processing systems}, 12, 1999.

\bibitem{orr2023multi}
James Orr and Ayan Dutta.
\newblock Multi-agent deep reinforcement learning for multi-robot applications: A survey.
\newblock {\em Sensors}, 23(7):3625, 2023.

\bibitem{huttenrauch2019deep}
Maximilian H{\"u}ttenrauch, Adrian {\v{S}}o{\v{s}}i{\'c}, and Gerhard Neumann.
\newblock Deep reinforcement learning for swarm systems.
\newblock {\em Journal of Machine Learning Research}, 20(54):1--31, 2019.

\bibitem{peng2020multi}
Haixia Peng and Xuemin Shen.
\newblock Multi-agent reinforcement learning based resource management in mec-and uav-assisted vehicular networks.
\newblock {\em IEEE Journal on Selected Areas in Communications}, 39(1):131--141, 2020.

\bibitem{littman1994markov}
Michael~L Littman.
\newblock Markov games as a framework for multi-agent reinforcement learning.
\newblock In {\em Machine learning proceedings 1994}, pages 157--163. Elsevier, 1994.

\bibitem{tampuu2017multiagent}
Ardi Tampuu, Tambet Matiisen, Dorian Kodelja, Ilya Kuzovkin, Kristjan Korjus, Juhan Aru, Jaan Aru, and Raul Vicente.
\newblock Multiagent cooperation and competition with deep reinforcement learning.
\newblock {\em PloS one}, 12(4):e0172395, 2017.

\bibitem{bazzan2009opportunities}
Ana~LC Bazzan.
\newblock Opportunities for multiagent systems and multiagent reinforcement learning in traffic control.
\newblock {\em Autonomous Agents and Multi-Agent Systems}, 18:342--375, 2009.

\bibitem{wiering2000multi}
Marco~A Wiering et~al.
\newblock Multi-agent reinforcement learning for traffic light control.
\newblock In {\em Machine Learning: Proceedings of the Seventeenth International Conference (ICML'2000)}, pages 1151--1158, 2000.

\bibitem{van2016deep}
Hado Van~Hasselt, Arthur Guez, and David Silver.
\newblock Deep reinforcement learning with double q-learning.
\newblock In {\em Proceedings of the AAAI conference on artificial intelligence}, volume~30, 2016.

\bibitem{vignon2023effective}
Colin Vignon, Jean Rabault, Joel Vasanth, Francisco Alc{\'a}ntara-{\'A}vila, Mikael Mortensen, and Ricardo Vinuesa.
\newblock Effective control of two-dimensional rayleigh--b{\'e}nard convection: Invariant multi-agent reinforcement learning is all you need.
\newblock {\em Physics of Fluids}, 35(6), 2023.

\bibitem{suarez2025active}
Pol Su{\'a}rez, Francisco Alc{\'a}ntara-{\'A}vila, Arnau Mir{\'o}, Jean Rabault, Bernat Font, Oriol Lehmkuhl, and Ricardo Vinuesa.
\newblock Active flow control for drag reduction through multi-agent reinforcement learning on a turbulent cylinder at r e d= 3900.
\newblock {\em Flow, Turbulence and Combustion}, pages 1--25, 2025.

\bibitem{lowe2017multi}
Ryan Lowe, Yi~I Wu, Aviv Tamar, Jean Harb, OpenAI Pieter~Abbeel, and Igor Mordatch.
\newblock Multi-agent actor-critic for mixed cooperative-competitive environments.
\newblock {\em Advances in neural information processing systems}, 30, 2017.

\bibitem{yu2022surprising}
Chao Yu, Akash Velu, Eugene Vinitsky, Jiaxuan Gao, Yu~Wang, Alexandre Bayen, and Yi~Wu.
\newblock The surprising effectiveness of ppo in cooperative multi-agent games.
\newblock {\em Advances in neural information processing systems}, 35:24611--24624, 2022.

\bibitem{kuba2021trust}
Jakub~Grudzien Kuba, Ruiqing Chen, Muning Wen, Ying Wen, Fanglei Sun, Jun Wang, and Yaodong Yang.
\newblock Trust region policy optimisation in multi-agent reinforcement learning.
\newblock {\em arXiv preprint arXiv:2109.11251}, 2021.

\bibitem{iqbal2019actor}
Shariq Iqbal and Fei Sha.
\newblock Actor-attention-critic for multi-agent reinforcement learning.
\newblock In {\em International conference on machine learning}, pages 2961--2970. PMLR, 2019.

\bibitem{sunehag2017value}
Peter Sunehag, Guy Lever, Audrunas Gruslys, Wojciech~Marian Czarnecki, Vinicius Zambaldi, Max Jaderberg, Marc Lanctot, Nicolas Sonnerat, Joel~Z Leibo, Karl Tuyls, et~al.
\newblock Value-decomposition networks for cooperative multi-agent learning.
\newblock {\em arXiv preprint arXiv:1706.05296}, 2017.

\bibitem{rashid2020monotonic}
Tabish Rashid, Mikayel Samvelyan, Christian~Schroeder De~Witt, Gregory Farquhar, Jakob Foerster, and Shimon Whiteson.
\newblock Monotonic value function factorisation for deep multi-agent reinforcement learning.
\newblock {\em Journal of Machine Learning Research}, 21(178):1--51, 2020.

\bibitem{de2020deep}
Christian~Schroeder de~Witt, Bei Peng, Pierre-Alexandre Kamienny, Philip Torr, Wendelin B{\"o}hmer, and Shimon Whiteson.
\newblock Deep multi-agent reinforcement learning for decentralized continuous cooperative control.
\newblock {\em arXiv preprint arXiv:2003.06709}, 19, 2020.

\bibitem{son2019qtran}
Kyunghwan Son, Daewoo Kim, Wan~Ju Kang, David~Earl Hostallero, and Yung Yi.
\newblock Qtran: Learning to factorize with transformation for cooperative multi-agent reinforcement learning.
\newblock In {\em International conference on machine learning}, pages 5887--5896. PMLR, 2019.

\bibitem{yang2018mean}
Yaodong Yang, Rui Luo, Minne Li, Ming Zhou, Weinan Zhang, and Jun Wang.
\newblock Mean field multi-agent reinforcement learning.
\newblock In {\em International conference on machine learning}, pages 5571--5580. PMLR, 2018.

\bibitem{wei2016lenient}
Ermo Wei and Sean Luke.
\newblock Lenient learning in independent-learner stochastic cooperative games.
\newblock {\em Journal of Machine Learning Research}, 17(84):1--42, 2016.

\bibitem{de2020independent}
Christian~Schroeder De~Witt, Tarun Gupta, Denys Makoviichuk, Viktor Makoviychuk, Philip~HS Torr, Mingfei Sun, and Shimon Whiteson.
\newblock Is independent learning all you need in the starcraft multi-agent challenge?
\newblock {\em arXiv preprint arXiv:2011.09533}, 2020.

\bibitem{tan1993multi}
Ming Tan.
\newblock Multi-agent reinforcement learning: Independent vs. cooperative agents.
\newblock In {\em Proceedings of the tenth international conference on machine learning}, pages 330--337, 1993.

\bibitem{foerster2018counterfactual}
Jakob Foerster, Gregory Farquhar, Triantafyllos Afouras, Nantas Nardelli, and Shimon Whiteson.
\newblock Counterfactual multi-agent policy gradients.
\newblock In {\em Proceedings of the AAAI conference on artificial intelligence}, volume~32, 2018.

\bibitem{matignon2012independent}
Laetitia Matignon, Guillaume~J Laurent, and Nadine Le~Fort-Piat.
\newblock Independent reinforcement learners in cooperative markov games: a survey regarding coordination problems.
\newblock {\em The Knowledge Engineering Review}, 27(1):1--31, 2012.

\bibitem{gupta2017cooperative}
Jayesh~K Gupta, Maxim Egorov, and Mykel Kochenderfer.
\newblock Cooperative multi-agent control using deep reinforcement learning.
\newblock In {\em Autonomous Agents and Multiagent Systems: AAMAS 2017 Workshops, Best Papers, S{\~a}o Paulo, Brazil, May 8-12, 2017, Revised Selected Papers 16}, pages 66--83. Springer, 2017.

\bibitem{albrecht2024multi}
Stefano~V Albrecht, Filippos Christianos, and Lukas Sch{\"a}fer.
\newblock {\em Multi-agent reinforcement learning: Foundations and modern approaches}.
\newblock MIT Press, 2024.

\bibitem{ha2016hypernetworks}
David Ha, Andrew Dai, and Quoc~V. Le.
\newblock Hypernetworks, 2016.

\bibitem{chauhan2023brief}
Vinod~Kumar Chauhan, Jiandong Zhou, Ping Lu, Soheila Molaei, and David~A Clifton.
\newblock A brief review of hypernetworks in deep learning.
\newblock {\em arXiv preprint arXiv:2306.06955}, 2023.

\bibitem{sarafian2021recomposing}
Elad Sarafian, Shai Keynan, and Sarit Kraus.
\newblock Recomposing the reinforcement learning building blocks with hypernetworks.
\newblock In {\em International Conference on Machine Learning}, pages 9301--9312. PMLR, 2021.

\bibitem{beck2023hypernetworks}
Jacob Beck, Matthew~Thomas Jackson, Risto Vuorio, and Shimon Whiteson.
\newblock Hypernetworks in meta-reinforcement learning.
\newblock In {\em Conference on Robot Learning}, pages 1478--1487. PMLR, 2023.

\bibitem{rezaei2023hypernetworks}
Sahand Rezaei-Shoshtari, Charlotte Morissette, Francois~R Hogan, Gregory Dudek, and David Meger.
\newblock Hypernetworks for zero-shot transfer in reinforcement learning.
\newblock In {\em Proceedings of the AAAI Conference on Artificial Intelligence}, volume~37, pages 9579--9587, 2023.

\bibitem{huang2021continual}
Yizhou Huang, Kevin Xie, Homanga Bharadhwaj, and Florian Shkurti.
\newblock Continual model-based reinforcement learning with hypernetworks.
\newblock In {\em 2021 IEEE International Conference on Robotics and Automation (ICRA)}, pages 799--805. IEEE, 2021.

\bibitem{botteghi2025hyperlparameterinformedreinforcementlearning}
Nicolò Botteghi, Stefania Fresca, Mengwu Guo, and Andrea Manzoni.
\newblock Hyperl: Parameter-informed reinforcement learning for parametric pdes, 2025.

\bibitem{christianos2021scaling}
Filippos Christianos, Georgios Papoudakis, Muhammad~A Rahman, and Stefano~V Albrecht.
\newblock Scaling multi-agent reinforcement learning with selective parameter sharing.
\newblock In {\em International Conference on Machine Learning}, pages 1989--1998. PMLR, 2021.

\bibitem{terry2020revisiting}
Justin~K Terry, Nathaniel Grammel, Sanghyun Son, Benjamin Black, and Aakriti Agrawal.
\newblock Revisiting parameter sharing in multi-agent deep reinforcement learning.
\newblock {\em arXiv preprint arXiv:2005.13625}, 2020.

\bibitem{vaswani2017attention}
Ashish Vaswani, Noam Shazeer, Niki Parmar, Jakob Uszkoreit, Llion Jones, Aidan~N Gomez, {\L}ukasz Kaiser, and Illia Polosukhin.
\newblock Attention is all you need.
\newblock {\em Advances in neural information processing systems}, 30, 2017.

\bibitem{botteghi2024parametric}
Nicol{\`o} Botteghi and Urban Fasel.
\newblock Parametric pde control with deep reinforcement learning and differentiable l0-sparse polynomial policies.
\newblock {\em arXiv preprint arXiv:2403.15267}, 2024.

\bibitem{gehring2017convolutional}
Jonas Gehring, Michael Auli, David Grangier, Denis Yarats, and Yann~N Dauphin.
\newblock Convolutional sequence to sequence learning.
\newblock In {\em International conference on machine learning}, pages 1243--1252. PMLR, 2017.

\bibitem{fujimoto2018addressing}
Scott Fujimoto, Herke Hoof, and David Meger.
\newblock Addressing function approximation error in actor-critic methods.
\newblock In {\em International conference on machine learning}, pages 1587--1596. PMLR, 2018.

\bibitem{peitz2023distributed}
Sebastian Peitz, Jan Stenner, Vikas Chidananda, Oliver Wallscheid, Steven~L Brunton, and Kunihiko Taira.
\newblock Distributed control of partial differential equations using convolutional reinforcement learning.
\newblock {\em arXiv preprint arXiv:2301.10737}, 2023.

\bibitem{tomasetto2024latent}
Matteo Tomasetto, Francesco Braghin, and Andrea Manzoni.
\newblock Latent feedback control of distributed systems in multiple scenarios through deep learning-based reduced order models.
\newblock {\em Computer Methods in Applied Mechanics and Engineering}, 442:118030, 2025.

\bibitem{tahir2019swarms}
Anam Tahir, Jari B{\"o}ling, Mohammad-Hashem Haghbayan, Hannu~T Toivonen, and Juha Plosila.
\newblock Swarms of unmanned aerial vehicles—a survey.
\newblock {\em Journal of Industrial Information Integration}, 16:100106, 2019.

\bibitem{gad2001flow}
Mohamed Gad-el Hak.
\newblock Flow control: The future.
\newblock {\em Journal of aircraft}, 38(3):402--418, 2001.

\bibitem{ng2000algorithms}
Andrew~Y Ng, Stuart Russell, et~al.
\newblock Algorithms for inverse reinforcement learning.
\newblock In {\em Icml}, volume~1, page~2, 2000.

\bibitem{hussein2017imitation}
Ahmed Hussein, Mohamed~Medhat Gaber, Eyad Elyan, and Chrisina Jayne.
\newblock Imitation learning: A survey of learning methods.
\newblock {\em ACM Computing Surveys (CSUR)}, 50(2):1--35, 2017.

\bibitem{watkins1992q}
Christopher~JCH Watkins and Peter Dayan.
\newblock Q-learning.
\newblock {\em Machine learning}, 8:279--292, 1992.

\bibitem{singh2000convergence}
Satinder Singh, Tommi Jaakkola, Michael~L Littman, and Csaba Szepesv{\'a}ri.
\newblock Convergence results for single-step on-policy reinforcement-learning algorithms.
\newblock {\em Machine learning}, 38:287--308, 2000.

\bibitem{arulkumaran2017deep}
Kai Arulkumaran, Marc~Peter Deisenroth, Miles Brundage, and Anil~Anthony Bharath.
\newblock Deep reinforcement learning: A brief survey.
\newblock {\em IEEE Signal Processing Magazine}, 34(6):26--38, 2017.

\bibitem{li2017deep}
Yuxi Li.
\newblock Deep reinforcement learning: An overview.
\newblock {\em arXiv preprint arXiv:1701.07274}, 2017.

\bibitem{franccois2018introduction}
Vincent Fran{\c{c}}ois-Lavet, Peter Henderson, Riashat Islam, Marc~G Bellemare, Joelle Pineau, et~al.
\newblock An introduction to deep reinforcement learning.
\newblock {\em Foundations and Trends{\textregistered} in Machine Learning}, 11(3-4):219--354, 2018.

\bibitem{williams1992simple}
Ronald~J Williams.
\newblock Simple statistical gradient-following algorithms for connectionist reinforcement learning.
\newblock {\em Machine learning}, 8:229--256, 1992.

\bibitem{silver2014deterministic}
David Silver, Guy Lever, Nicolas Heess, Thomas Degris, Daan Wierstra, and Martin Riedmiller.
\newblock Deterministic policy gradient algorithms.
\newblock In {\em International conference on machine learning}, pages 387--395. Pmlr, 2014.

\bibitem{schulman2017proximal}
John Schulman, Filip Wolski, Prafulla Dhariwal, Alec Radford, and Oleg Klimov.
\newblock Proximal policy optimization algorithms.
\newblock {\em arXiv preprint arXiv:1707.06347}, 2017.

\bibitem{lillicrap2015continuous}
Timothy~P Lillicrap, Jonathan~J Hunt, Alexander Pritzel, Nicolas Heess, Tom Erez, Yuval Tassa, David Silver, and Daan Wierstra.
\newblock Continuous control with deep reinforcement learning.
\newblock {\em arXiv preprint arXiv:1509.02971}, 2015.

\bibitem{fujimoto2023sale}
Scott Fujimoto, Wei-Di Chang, Edward Smith, Shixiang~Shane Gu, Doina Precup, and David Meger.
\newblock For sale: State-action representation learning for deep reinforcement learning.
\newblock {\em Advances in neural information processing systems}, 36:61573--61624, 2023.

\end{thebibliography}

\appendix
\section{Reinforcement learning}

\emph{Reinforcement learning} (RL) is a branch of machine learning concerned with how \emph{agents}, namely the controllers of the distributed system, should take \emph{actions}, namely select control inputs, in an environment in order to maximize their cumulative reward \cite{sutton2018reinforcement}. 
Unlike supervised learning, where algorithms learn from a fixed dataset of labeled input-output pairs, RL agents learn from trial and error. An agent improves its behavior over time thanks to the experience gained while exploring the environment with different actions and receiving feedback signals in the form of rewards. Differently from unsupervised learning where no feedback is provided, the reward feedback is crucial to discriminate actions and to guide agent learning. 
Since the agent does not initially know which actions will yield the highest reward, it must explore different strategies to improve is behavior, balancing the trade-off between exploiting known high reward actions and exploring new, potentially even higher reward actions. This framework is suitable for a wide range of sequential decision-making problems, such as game playing, robotic control, autonomous driving, and resource management. 

Formally, RL problems are often modeled using Markov Decision Processes (MDPs), which provide a mathematical framework for modeling decision-making in situations where outcomes are partly random and partly under the control of the agent. 
\begin{definition}
    An MDP is defined by a tuple $(\mathcal{Y}, \mathcal{U}, F, R)$, where $\mathcal{Y}$ and $\mathcal{U}$ denote the state and action spaces, respectively, $F:\mathcal{Y}\times \mathcal{Y}\times \mathcal{U}\mapsto[0,1]$ indicates the transition probability for any state $y, y' \in \mathcal{Y}$ and $u \in \mathcal{U}$, and $R:\mathcal{Y}\times \mathcal{U}\mapsto \mathbb{R}$ indicates the reward function that determines the instantaneous reward the agent receives for each action taken in a given state.  
\end{definition}
For a given time step $k$, the agent observes the state $y_k$ and takes action $u_k$. As a result, the environment changes its state to some $y_{k+1}$ according to the transition probability $F$. Additionally, the agent receives the reward $r_k$ according to the reward function $R$. It is worth mentioning that for deterministic systems, the transition probability function is replaced by the transition function $\bar{F}:\mathcal{Y}\times \mathcal{U}\mapsto\mathcal{Y}$.

The agent's behavior is defined by its \emph{policy} $\pi$, namely the set of rules that the agent exploits to select a specific action in a given state. The policy may be either stochastic $\pi:\mathbb{R}^{N_y} \times \mathbb{R}^{N_u}\to [0,1]$ or deterministic $\pi:\mathbb{R}^{N_y}\to \mathbb{R}^{N_u}$, where $N_y$ and $N_u$ indicate the dimensions of the state and the action space, respectively. The agent's goal is to find the optimal policy, i.e. the policy that maximizes the total expected cumulative reward from every state $\mathbf{y}$:
\begin{equation}
    J(y) = \mathbb{E}_{\pi}\left[\sum_{k=0}^H \gamma^kr_k|Y_0=y\right]\, ,
\end{equation}
where $\gamma \in [0, 1)$ is the discount factor and $H$ is the horizon over which actions are taken that may be either finite or infinite. However, maximizing the long-term behavior while only receiving immediate reward is extremely challenging. 

RL algorithms can be classified based on the strategy employed to find the optimal policy: {\em (i)} \emph{value-based methods}, {\em (ii)} \emph{policy-based methods}, and {\em (iii)} \emph{actor-critic methods}. Value-based methods look for an estimate of the state-action value function $Q(\mathbf{y},\mathbf{u}):\mathbb{R}^{N_y}\times \mathbb{R}^{N_u} \to \mathbb{R}$ associated with a policy $\pi$. The state-action value function gives the expected return obtained when following the policy $\pi$ over the whole horizon $H$, starting from a state $\mathbf{y}$. The optimal policy is derived by taking the greedy action with respect to the value function estimate. Examples of value-based methods are Q-learning \cite{watkins1992q} and SARSA \cite{singh2000convergence}. Policy-based methods, instead, consider a parametrization of the policy $\pi(\cdot;\boldsymbol{\theta})=\pi_{\boldsymbol{\theta}}(\cdot)$ with respect to a set of parameters $\boldsymbol{\theta}$ -- which may be, for instance, neural network weights and biases in the context of deep reinforcement learning (DRL) \cite{arulkumaran2017deep, li2017deep, franccois2018introduction} -- and aim at optimizing those parameters through the policy gradient \cite{sutton1999policy, sutton2018reinforcement}, i.e. the gradient of the return $J$ with respect to the policy parameters $\boldsymbol{\theta}$
\begin{equation}
    \nabla J(\boldsymbol{\theta})=\mathbb{E}_{\mathbf{u}\sim\pi_{\boldsymbol{\theta}}(\cdot),\mathbf{y}\sim\eta_{\pi_{\boldsymbol{\theta}}}}[Q_{\pi_{\boldsymbol{\theta}}}(\mathbf{y},\mathbf{u})\nabla \log\pi_{\boldsymbol{\theta}}(\mathbf{u}|\mathbf{y})]\, ,
\end{equation}
where $\nabla \log\pi_{\boldsymbol{\theta}}(\mathbf{u}|\mathbf{y})$ is the score function of the policy and $\eta_{\pi_{\boldsymbol{\theta}}}$ is the state occupancy measure under the policy $\pi_{\boldsymbol{\theta}}$. Examples of policy-based methods are REINFORCE \cite{williams1992simple} and deterministic policy gradient \cite{silver2014deterministic}. Eventually, actor-critic strategies combine the estimation of the value function with the direct optimization of the policy parameters. Examples are proximal policy optimization (PPO) \cite{schulman2017proximal} and deep deterministic policy gradient (DDPG) \cite{lillicrap2015continuous}. Another important distinction regarding RL algorithms is that between {\em (i)} \emph{model-free} and {\em (ii)} \emph{model-based} algorithms. Model-free algorithms utilize the data collected from the interaction with the environment to directly learn the optimal policy and value function. Conversely, model-based RL exploits the data stream to build a surrogate model of the environment dynamics that serves as a computationally efficient proxy of the real environment to optimize the control strategy. While model-free RL is typically easier to implement and effective in high-dimensional and complex environments, it suffers from data inefficiency. Model-based RL addresses the problem of the sample inefficiency of model-free RL, but it is often subjected to model biases that degrade the policy performance when deployed on the real system.

\subsection{Multi-agent reinforcement learning}

When multiple decision-making agents interact with the environment to achieve certain goals, we can talk about a multi-agent system \cite{albrecht2024multi}. Multi-agent systems find applications in a large variety of domains such as robotics, distributed control, collaborative decision-support systems. Multi-agent reinforcement learning (MARL) algorithms learn optimal policies for a set of agents in a multi-agent system. 

MARL is formally modeled using the multi-agent extension of the MDP, namely the stochastic game \cite{bucsoniu2010multi}.
\begin{definition}
    A stochastic game is defined by a tuple $(\mathcal{Y}, \mathcal{U}_1, \dots,\mathcal{U}_N, F, R_1, \dots,R_N)$, where $\mathcal{Y}$ denotes the state space, $\mathcal{U}_i$ with $i=1,\dots,N$  are the actions spaces of the individual agents, $N$ is the number of agents, $\mathcal{U}=\mathcal{U}_1 \times \mathcal{U}_2 \times \dots \times \mathcal{U}_N$ is the joint action space, $F:\mathcal{Y}\times \mathcal{Y}\times \mathcal{U}\mapsto[0,1]$ indicates the transition probability, and $R_i:\mathcal{Y}\times \mathcal{U}\mapsto \mathbb{R}$ with $i=1,\dots,N$ indicates the reward functions of the agents.  
\end{definition}

An important distinction among MARL algorithms  lies in the assumptions made when training and after training the agents. In particular, we can identify three classes of methods, namely (i) centralized training and execution, (ii) decentralized training and execution, and (iii) centralized training and decentralized execution. In the centralized training and execution approach each agent share all observations with the others during the learning and deployment phase. Conversely, decentralized training and executions approaches assume no shared knowledge among the agents and each agent only relies on local information. Eventually, centralized training and decentralized execution methods combine the two aforementioned approaches, where the information are shared among the agents only during training and during execution only local information is used. Many other factors can be used to classify MARL algorithms, such as the number of agents, the assumptions about the rewards, or the type of solution to achieve. For more information, we refer the reader to \cite{albrecht2024multi}.



\section{Twin-delayed deep deterministic policy gradient}\label{sec:TD3}

In our numerical experiments, we utilize a model-free, online, off-policy, and actor-critic approach, namely twin-delayed deep deterministic policy gradient (TD3) \cite{fujimoto2018addressing}. TD3 learns a deterministic policy $\pi(\cdot)$, i.e., the actor, and the action-value function $Q(\cdot)$, i.e., the critic. The actor and the critic are parametrized by means of two DNNs of parameters $\boldsymbol{\theta}_Q$ and $\boldsymbol{\theta}_{\pi}$, respectively.  We indicate the parametrized policy with $\pi(\mathbf{y}_t; \boldsymbol{\theta}_{\pi})$ and the action-value function with $Q(\mathbf{y}_t, \mathbf{u}_t;\boldsymbol{\theta}_Q)$. TD3 can handle continuous state and action spaces, making it a suitable candidate for controlling parametric PDEs using smooth control strategies.

TD3 relies on a memory buffer $\mathcal{D}$ to store the interaction data $(\mathbf{y}_t, \mathbf{u}_t, r_t, \mathbf{y}_{t+1})$ for all time steps $t$. Given a randomly-sampled batch of interaction tuples, we can update the parameters $\boldsymbol{\theta}_Q$ of the action-value function $Q(\mathbf{y}_t, \mathbf{u}_t;\boldsymbol{\theta}_Q)$ as:
\begin{equation}
\begin{split}
        \mathcal{L}(\boldsymbol{\theta}_Q) &= \mathbb{E}_{\mathbf{y}_t, \mathbf{u}_t, \mathbf{y}_{t+1}, r_t \sim \mathcal{D}}\Big[f\Big(r_k + \gamma \bar{Q}(\mathbf{y}_{t+1}, \mathbf{u}_{t+1}; \boldsymbol{\theta}_{\bar{Q}}) - Q(\mathbf{y}_t, \mathbf{u}_t; \boldsymbol{\theta}_Q)\Big)\Big] \\
        &= \mathbb{E}_{\mathbf{y}_t, \mathbf{u}_t, \mathbf{y}_{t+1}, r_t \sim \mathcal{D}}\Big[f\Big(\underbrace{r_t + \gamma \bar{Q}(\mathbf{y}_{t+1}, \bar{\pi}(\mathbf{y}_{t+1};\boldsymbol{\theta}_{\bar{\pi}})+\boldsymbol{\epsilon}; \boldsymbol{\theta}_{\bar{Q}})}_{\text{target value}} - Q(\mathbf{y}_t, \mathbf{u}_t; \boldsymbol{\theta}_Q)\Big)\Big]\,,
\end{split}
\label{eq:valuefunctionloss}
\end{equation}
where $f$ is a arbitrary chosen loss function, e.g. mean-squared error, the so-called target networks $\bar{Q}(\mathbf{y}_t, \mathbf{u}_t; \boldsymbol{\theta}_{\bar{Q}})$ and $\bar{\pi}(\mathbf{y}_t;\boldsymbol{\theta}_{\bar{\pi}})$ are copies of $Q(\mathbf{y}_t, \mathbf{u}_t;\boldsymbol{\theta}_Q)$ and $\pi(\mathbf{y}_t; \boldsymbol{\theta}_{\pi})$, respectively, with frozen parameters, i.e., they are not updated in the backpropagation step to improve the stability of the training. We indicate with $\boldsymbol{\epsilon} \sim \text{clip}(\mathcal{N}(\bm{0}, \bar{\boldsymbol{\sigma}}), -c, c)$ the noise added to estimate the action value in the interval $[-c, c]$ around the target action. Similarly to \cite{fujimoto2023sale}, we choose $f$ to be the Huber loss.

Instead of relying on the standard target-value update that TD3 uses, we opt for the modified objective proposed in \cite{fujimoto2023sale} as shown to be improving the value function estimation. We still estimate to two independent action-value functions, namely $Q_1(\mathbf{y}_t,\mathbf{u}_t;\boldsymbol{\theta}_{Q_1})$ and  $Q_2(\mathbf{y}_t,\mathbf{u}_t;\boldsymbol{\theta}_{Q_2})$, and two target action-value functions $\bar{Q}_1(\mathbf{y}_t,\mathbf{u}_t;\boldsymbol{\theta}_{\bar{Q}_1})$ and  $\bar{Q}_2(\mathbf{y}_t,\mathbf{u}_t;\boldsymbol{\theta}_{\bar{Q}_2})$, and computes the target value for regression as:
\begin{equation}
    \underbrace{    r_t + \gamma \text{clip} \Big(\min(Q_1(\cdot), Q_2(\cdot), Q_{\text{min}},Q_{\text{max}}\Big)\,,}_{\text{target value}}
\end{equation}
where $Q_{\text{min}}$ and $,Q_{\text{max}}$ are the minimum and maximum q-values estimates over a batch of samples.

The action-value functions $Q_1(\mathbf{y}_t, \mathbf{u}_t;\boldsymbol{\theta}_{Q_1})$ and $Q_2(\mathbf{y}_t, \mathbf{u}_t;\boldsymbol{\theta}_{Q_2})$ are used to update the parameters of the deterministic policy $\pi(\mathbf{y}_t;\boldsymbol{\theta}_{\pi})$ according to the deterministic policy gradient theorem \cite{silver2014deterministic}. In particular, the gradient of the critic guides the improvements of the actor and the policy parameters are updated to ascend the action-value function:
\begin{equation}
    \mathcal{L}(\boldsymbol{\theta}_{\pi}) = \mathbb{E}_{\mathbf{y}_t \sim \mathcal{D}}\Big[-\frac{1}{2}\nabla_{\mathbf{u}_t} Q_1(\mathbf{y}_t, \pi(\mathbf{y}_t; \boldsymbol{\theta}_{\pi}); \boldsymbol{\theta}_{Q_1})-\frac{1}{2}\nabla_{\mathbf{u}_t} Q_2(\mathbf{y}_t, \pi(\mathbf{y}_t; \boldsymbol{\theta}_{\pi}); \boldsymbol{\theta}_{Q_2})\Big].
\label{eq:policytd3}
\end{equation}

The target networks, parametrized by $\boldsymbol{\theta}_{\bar{Q}_1}$, $\boldsymbol{\theta}_{\bar{Q}_2}$, and $\boldsymbol{\theta}_{\bar{\pi}}$, respectively, are updated with a slower frequency than the actor and the critic according to:
\begin{equation}
\begin{split}
\boldsymbol{\theta}_{\bar{Q}_1} &= \rho \boldsymbol{\theta}_{Q_1} + (1-\rho)\boldsymbol{\theta}_{\bar{Q}_1}\, , \\
\boldsymbol{\theta}_{\bar{Q}_2} &= \rho \boldsymbol{\theta}_{Q_2} + (1-\rho)\boldsymbol{\theta}_{\bar{Q}_1}\, , \\
\boldsymbol{\theta}_{\bar{\pi}} &= \rho \boldsymbol{\theta}_{\pi} + (1-\rho)\boldsymbol{\theta}_{\bar{\pi}}\, , \\
\end{split}
\label{target_nns_update}
\end{equation}
where $\rho$ is a constant factor determining the speed of the updates of the target parameters.

\subsection{HypeRL TD3}
In \cite{botteghi2025hyperlparameterinformedreinforcementlearning}, the TD3 algorithm was enhanced by the use of hypernetworks. Similarly to TD3, HypeRL TD3 estimates two action-value functions  $Q_1(\mathbf{y}_t, \mathbf{u}_t;\boldsymbol{\theta}_{Q_1})$ and $Q_2(\mathbf{y}_t, \mathbf{u}_t;\boldsymbol{\theta}_{Q_2})$ and a policy  $\pi(\mathbf{y}_t;\boldsymbol{\theta}_{\pi})$ by means of neural networks. However, the parameters of the main networks are now learned using three hypernetworks $H_{Q_1}(\bm{z}_t; \boldsymbol{\theta}_{H_{Q_1}})$, $H_{Q_2}(\bm{z}_t; \boldsymbol{\theta}_{H_{Q_2}})$, and $H_{\pi}(\bm{z}_t; \boldsymbol{\theta}_{H_{\pi}})$:
\begin{equation}
    \begin{split}
        \boldsymbol{\theta}_{Q_1} &= H_{Q_1}(\bm{z}_t; \boldsymbol{\theta}_{h_{Q_1}})\, , \\
        \boldsymbol{\theta}_{Q_2} &= H_{Q_2}(\bm{z}_t; \boldsymbol{\theta}_{h_{Q_2}})\, , \\
        \boldsymbol{\theta}_{\pi} &= H_{\pi}(\bm{z}_t; \boldsymbol{\theta}_{h_{\pi}})\, ,\\
    \end{split}
\end{equation}
where $\bm{z}_t$ indicates the hypernetwork input. HypeRL TD3 replaces the policy with:
\begin{equation}
\begin{split}
    \boldsymbol{\theta}_{\pi} &= H_{\pi}(\bm{z}_t;\boldsymbol{\theta}_{H_{\pi}})\, , \\
    \mathbf{u}_t &= \pi(\mathbf{y}_t; \boldsymbol{\theta}_{\pi})\, , \\  
\end{split}
\end{equation}
and analogously, the value function with:
\begin{equation}
\begin{split}
    \boldsymbol{\theta}_{Q_k} &= H_{Q_k}(\bm{z}_t;\boldsymbol{\theta}_{H_{Q_k}})\, , \\
    q_{k,t} &= Q_k(\mathbf{y}_t, \mathbf{u}_t; \boldsymbol{\theta}_{Q_k})\, , \\   
\end{split}
\end{equation}
where $q_{k,t}$ is the predicted Q-value by the action-value function $Q_k(\cdot, \cdot)$ and $k=\{1,2\}$. The hypernetwork parameters are jointly optimized with the main network parameters by simply allowing the gradient of the TD3 training objectives (see Equations \eqref{eq:valuefunctionloss} and \eqref{eq:policytd3}).

\section{Comparison with RL}
\label{sec:RLcomparison}
\subsection{Density control in a vacuum}

\begin{figure*}[h!]
    \centering
\subfloat{\includegraphics[height=0.26\textwidth]{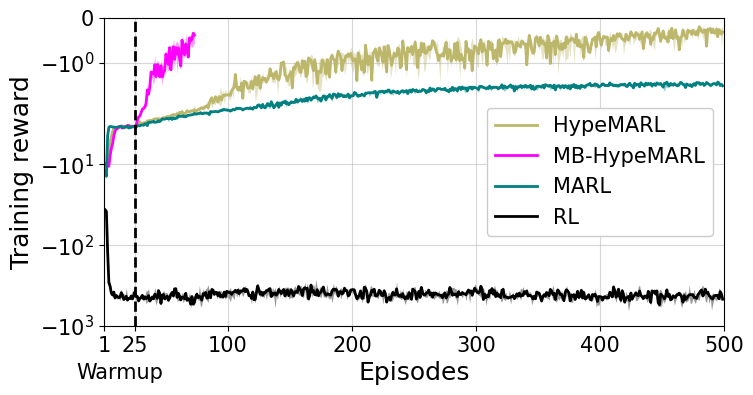}}
    \subfloat{\includegraphics[height=0.26\textwidth]{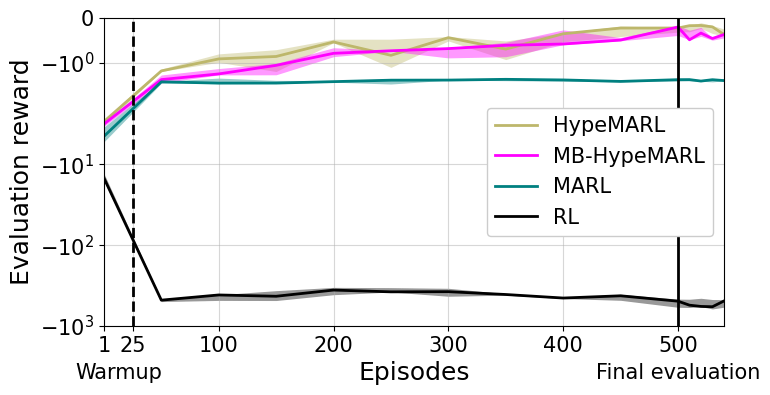}}
    \caption{{\em Density control in a vacuum}. Median and interquartile range of HypeMARL, MB-HypeMARL, MARL and single-agent RL training end evaluation rewards.}
    \label{fig:FP_Vacuum3}
\end{figure*}

In Figure \ref{fig:FP_Vacuum3}, we show a comparison of HypeMARL, MB-HypeMARL, MARL with a single-agent RL learning a single global policy mapping high-dimensional states to high-dimensional actions. Similarly to MARL-based methods, the single-agent RL relies on TD3. However, due to the high-dimensionality of the problem, the single-agent RL cannot learn a good policy and all the MARL agents drastically outperform single RL.

\section{Hyperparameters of the numerical experiments}
\label{sec:hyperparam}
In this section, we report the value of the hyperparameters of our experiments. In particular, we list
in Table \ref{table:app3} the hyperparameters of HypeMARL, MB-HypeMARL, MARL, and RL.


\begin{table}
\caption{Agents}
\centering
\begin{tabular}{||c c c c c||} 
 \hline
  & HypeMARL & MB-HypeMARL & MARL & RL \\ 
 \hline\hline
 RL algorithm & TD3 & TD3 & TD3 & TD3 \\
 \hline
 batch size & $32$ & $32$ & $32$ & $32$\\ 
 \hline
 positional encoding $d$ & $2048$ & $2048$ & $-$ & $-$\\
  \hline
 positional encoding $n$ & $1000$ & $1000$ & $-$ & $-$\\
 \hline
 discount $\gamma$ & $0.99$ & $0.99$ & $0.99$ & $0.99$\\
 \hline
 learning rate actor $\pi$ & $1e-6$ & $1e-6$ & $3e-4$ & $3e-4$ \\
 \hline
 learning rate critic $Q$ & $5e-5$ & $5e-5$ & $3e-4$ & $3e-4$ \\ 
\hline
 num. hidden layer actor $\pi$ & $1$ & $1$ & $2$ & $2$\\
\hline
 num. hidden layer critic $Q$ & $1$ & $1$ & $2$ & $2$ \\
 \hline
 hidden dim. actor $\pi$ & $256$ & $256$ & $256$ & $256$\\
\hline
 hidden dim. critic $Q$ & $256$ & $256$ & $256$ & $256$\\
\hline
 learning rate forward dynamics $\bar{F}$ & $-$ & $1e-4$ & $-$ & $-$ \\
\hline
 num. layer forward dynamics $\bar{F}$ & $-$ & $1$ & $-$ & $-$ \\
\hline
 hidden dim. forward dynamics $\bar{F}$ & $-$ & $256$ & $-$ & $-$ \\ 
 \hline
\end{tabular}
\label{table:app3}
\end{table}

\end{document}